\documentclass[11pt]{article}
\pdfoutput=1

\usepackage[margin=1in]{geometry}
\usepackage[T1]{fontenc}
\usepackage[utf8]{inputenc}
\usepackage{microtype}
\usepackage{newtxtext}
\usepackage{booktabs}
\usepackage{tabularx}
\usepackage{rotating}
\usepackage{graphicx}
\graphicspath{{figures/}}
\usepackage{amsmath}
\usepackage{newtxmath}
\usepackage[scaled=0.92]{newtxtt}
\usepackage{xcolor}
\usepackage{caption}
\usepackage{setspace}
\usepackage{authblk}
\usepackage{fancyhdr}
\usepackage{titlesec}
\usepackage{enumitem}
\usepackage[round]{natbib}
\definecolor{kblue}{HTML}{1B3A5C}
\definecolor{kgrey}{HTML}{5A6672}
\usepackage[colorlinks=true,linkcolor=kblue,citecolor=kblue,urlcolor=kblue,
            pdftitle={K-Bench: measuring model performance on real scientific agent requests},
            pdfauthor={Aubrey M. Brueckner, Darshil Patel, Yuhuan He, Timothy Kassis},
            pdfsubject={Evaluation of frontier models on real scientific agent requests}]{hyperref}

\renewcommand{\arraystretch}{1.08}
\titleformat{\section}{\normalfont\large\bfseries\color{kblue}}{\thesection}{0.6em}{}
\titleformat{\subsection}{\normalfont\normalsize\bfseries\color{kblue}}{\thesubsection}{0.6em}{}
\titleformat{\subsubsection}{\normalfont\normalsize\itshape}{\thesubsubsection}{0.6em}{}

\newcommand{\M}[1]{\texttt{\small #1}}
\title{\vspace{-1.2cm}\bfseries\color{kblue}K-Bench: measuring model performance on real\\[0.15em] scientific agent requests}
\author[1]{Aubrey M. Brueckner}
\author[1]{Darshil Patel}
\author[1]{Yuhuan He}
\author[1,*]{Timothy Kassis}
\affil[1]{\normalsize K-Dense, Inc.}
\affil[*]{\normalsize Corresponding author: \texttt{timothy.kassis@k-dense.ai}}
\date{}

\begin{document}
\maketitle
\thispagestyle{fancy}

\begin{abstract}
\noindent
Benchmarks for scientific artificial intelligence are mostly written to be scored:
multiple-choice questions, curated agent tasks with reference solutions, or
simulators with a known generative structure. Real scientific requests arrive
differently. They are underspecified, they carry attachments, and they lack ground truth.
We report K-Bench~01, an evaluation built from first-turn requests
sampled from live user traffic on K-Dense Web and
run end to end by nine
frontier models in identical sandboxes, yielding 1{,}602 completed agent runs.
Three blinded language-model judges scored every run against an eight-dimension
rubric. On a rubric whose 8-anchor is defined as work a domain scientist would accept with
minor edits, no model clears the line under all three judges. \M{gpt-5.6-sol} has the highest pooled mean, 8.04, but its
95\% interval [7.80, 8.23] spans the threshold, and two of the three judges rank
\M{claude-opus-5} first instead.
We therefore report the ordering of systems as the reproducible quantity, the absolute
level as an attribute of the instrument, and the top of the table as unresolved. Across all 39{,}934 scored judgments --- the eight dimension
scores plus a holistic overall for each assessment, excluding not-applicable cells ---
47.6\% fall below the 8-point threshold. Difficulty is not
uniform across the rubric: scientific accuracy averages 6.22 against 7.33 for
communication, on identical denominators and in the same direction within every one of
the nine models. The single leading failure tag is
overclaiming, on 31.4\% of assessments.
We argue that the informative quantity for scientific agents is not a
leaderboard position but the joint distribution of what was delivered, what was
claimed, and what artifacts were produced.
\end{abstract}

\begin{figure}[ht]
\centering
\includegraphics[width=\textwidth]{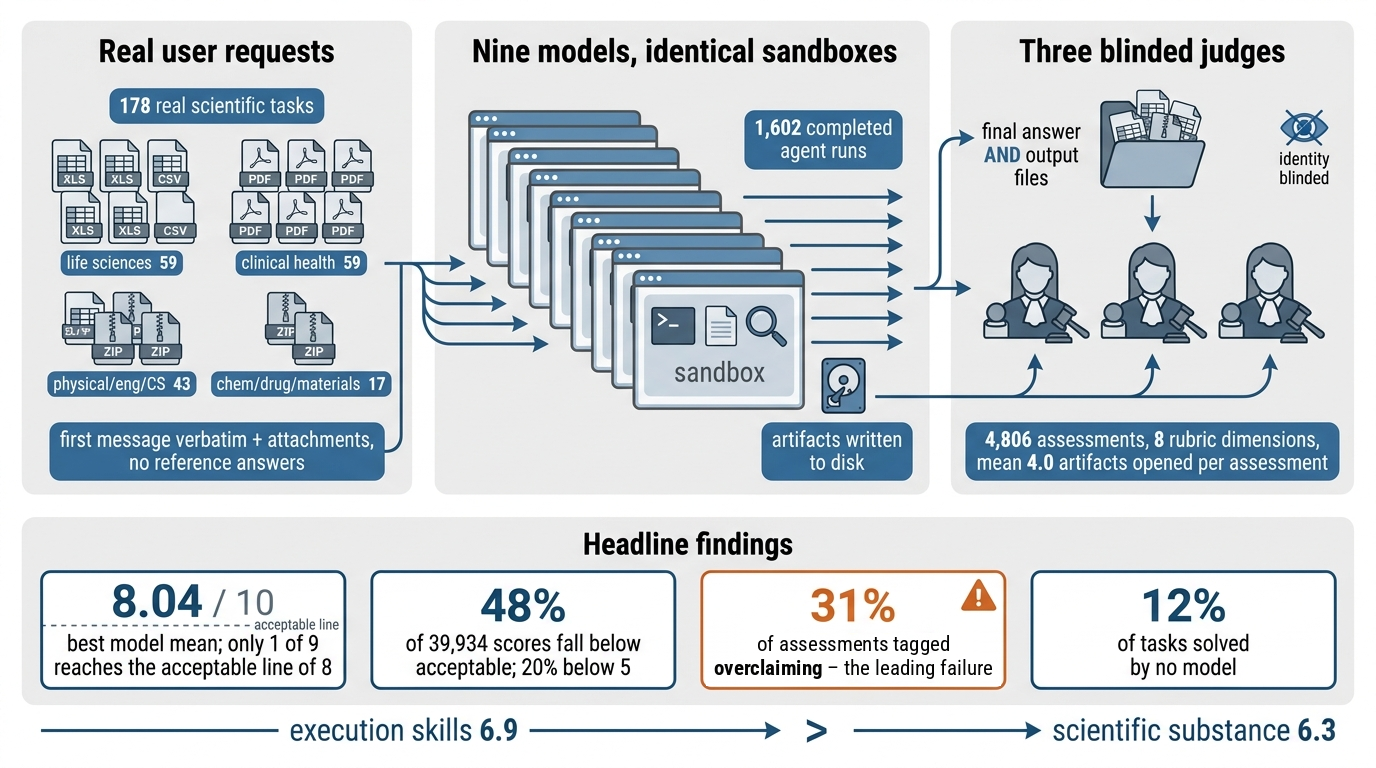}
\caption{\textbf{Graphical abstract.} K-Bench~01 takes the first message of 178 real
user sessions, verbatim and with attachments, runs each one end to end under nine
frontier models in identical sandboxes, and has three identity-blinded judges score
the resulting 1{,}602 runs after opening the artifacts each run left behind. The file
icons in the left panel are illustrative of the attachment mix rather than a
per-domain format breakdown; the most frequently attached format across the corpus is
\M{.docx} (Table~\ref{tab:filetype}). Tile
values are the headline results: the best model mean (8.04/10, a point estimate whose
interval spans the acceptable line, and which only one of the three judges produces;
see Section~\ref{sec:selfpref}, and Section~\ref{sec:validity} for why the count of
models reaching 8 is zero, one or two depending on the judge), the share of the
39{,}934 scored judgments below the acceptable line (47.6\%, rendered as 48\%), the
share of assessments carrying the \M{overclaiming} tag (31.4\%), and the share of
tasks no model solved (12.4\%, rendered as 12\%). The
strip beneath contrasts the mean of the execution dimensions (6.93) with the mean of
the substance dimensions (6.34); Section~\ref{sec:dimgap} gives the sharper, denominator-matched
version: scientific accuracy trails communication by 1.11 points pooled, and in the same
direction within every one of the nine models.}
\label{fig:ga}
\end{figure}


\section{Introduction}
\label{sec:intro}

A scientist sends an agent a count matrix from an RNA-seq experiment and asks which
genes are differentially expressed, and whether the batch effect is real. Another
attaches three papers and asks whether an effect replicates. These requests vary in shape; they are the first message of a real working
session and may arrive with files, often specify the goal only partly, and typically lack
a validated reference answer.

Today, few benchmarks used to track progress in scientific artificial
intelligence have this shape, and for a defensible reason: a benchmark has to be
scorable. This has yielded exam-style suites, curated agent tasks with reference solutions,
and simulators with a known generative structure. Each design choice leaves a gap where the typical real-world request lives. The gap is a
construct-validity problem \citep{constructvalidity}.

To date, the K-Dense Web platform has processed over 75{,}000 interactions across
approximately 18{,}000 user sessions. We used 178 first-turn requests from live
K-Dense Web traffic \citep{kdenseweb} and ran each one under nine models in an
identical stock harness (Section~\ref{sec:methods}). Figure~\ref{fig:ga} summarizes
the pipeline and the headline results. The design question is whether
a traffic-derived set still separates frontier systems, and on which axes.

Every score in this paper is
awarded by a panel of three language-model judges applying a written rubric to a run's
transcript and to the files it left on disk. The measured quantity is therefore panel-assessed rubric
compliance, and Section~\ref{sec:validity} sets out what that does and does not
establish. In short: the panel's ordering of systems reproduces across judges, and the
level at which it places the scale does not.

Our results show that the axes are not the ones a capability-first
reading would predict. These findings motivate two framing commitments. The first is that eloquence is not a
deliverable; an evaluation that cannot see the difference between prose and an empty directory will
systematically overstate progress \citep{ideanovelty,ideationexecution}. The second is that a single leaderboard number is
the wrong summary of an agentic scientific benchmark.

This paper contributes the following.

\begin{itemize}[leftmargin=1.4em,itemsep=0.25em,topsep=0.3em]
\item We construct a benchmark from unmodified deployment traffic: 178 first-turn
  scientific requests taken verbatim from live users, with their attachments, without reference answers
  (Section~\ref{sec:methods}).
\item We grade the files a run left behind rather than its prose alone. Judges receive
  read-only access to the run's output tree
  (Section~\ref{sec:methods}).
\item We run a balanced campaign at scale: nine frontier models on identical tasks in
  identical sandboxes, 1{,}602 runs, three blinded judges, 4{,}806 assessments and
  39{,}934 scored judgments (Section~\ref{sec:results}).
\item We treat the judging panel as an object of study rather than only as an instrument,
  separating what it reproduces from what it asserts, and showing
  that the count of models clearing the rubric's threshold is panel-dependent
  (Section~\ref{sec:judges}).
\item We report where the deficit sits: scientific accuracy trails communication within
  every model in the field, overclaiming is the leading failure tag at 31.4\% of
  assessments, and 47.9\% of runs finish with no file on disk (Sections~\ref{sec:dimgap}--\ref{sec:stops}).
\end{itemize}


\section{Related work: the 2020--2026 evaluation landscape}
\label{sec:related}

\begin{figure}[t]
\centering
\includegraphics[width=\textwidth]{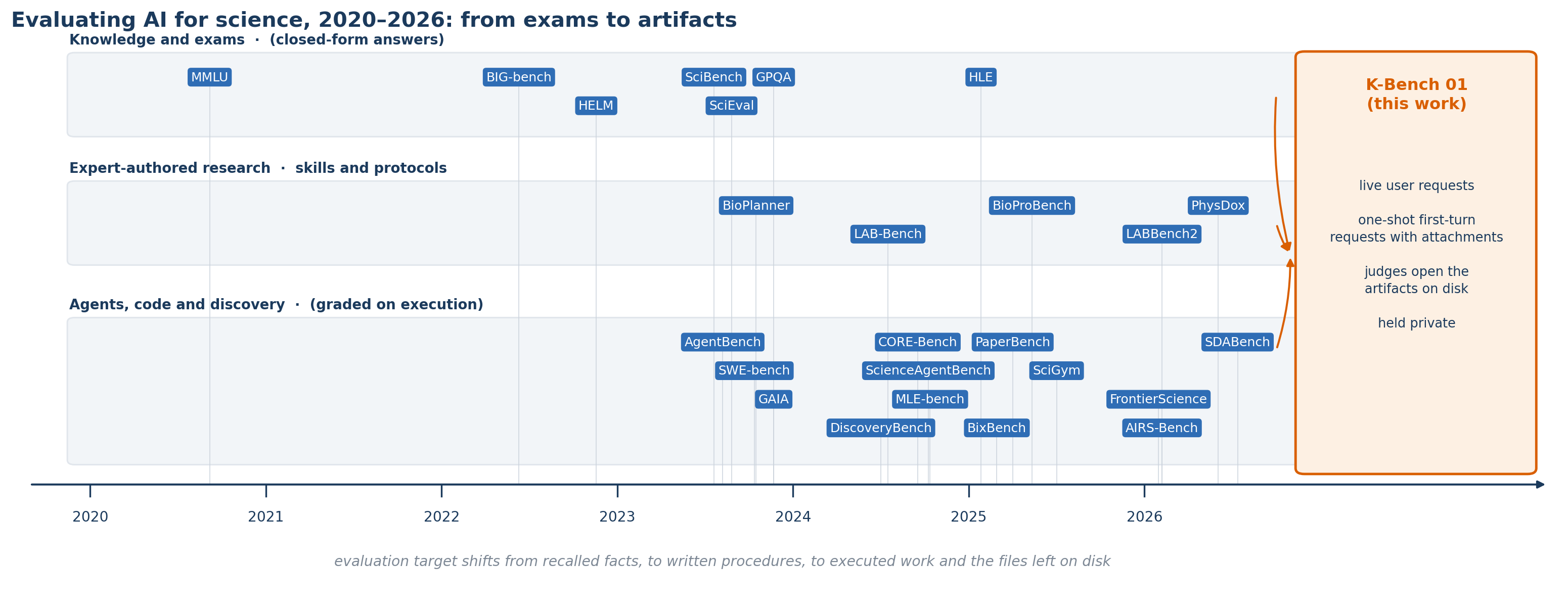}
\caption{\textbf{Representative benchmarks for scientific and agentic capability,
2020--2026, grouped by what they measure.} Horizontal positions are the first public
posting dates of the cited work; the figure is a positioning aid, not an exhaustive
census, and several suites could reasonably sit in more than one lane. Several suites
discussed below are omitted here for legibility; Table~\ref{tab:landscape} gives the
fuller comparison. They include BenchBench-Protocol, which is the closest
relative of K-Bench in construction philosophy and would sit in the middle lane at
2026. K-Bench~01 is placed at the right. Building a benchmark out of deployment
traffic is not itself new: WildBench and Arena-Hard curate items from chat logs and
RealClawBench reconstructs developer-agent sessions
\citep{wildbench,arenahard,realclawbench}, and within science AstaBench is the
nearest precedent, with problems inspired by requests to its deployed agents. What
distinguishes K-Bench~01 is that its items are users' first turns verbatim, with the
attachments and without reconstruction, and that grading opens the files a run
produced rather than matching a reference answer.}
\label{fig:timeline}
\end{figure}

Evaluation of scientific artificial intelligence has moved through three overlapping
generations in six years. A useful way to read the field is by what each
generation treats as the object of measurement (Figure~\ref{fig:timeline}). The
first generation measures recalled and reasoned-about knowledge. The second measures
written procedure. The third and current generation measures executed work, and it is only in the third
that the question of artifact quality becomes askable at all.

\subsection{Knowledge and exam suites}

The first generation established the discipline of reproducible
scoring. MMLU set the template of large multiple-choice coverage \citep{mmlu},
BIG-bench pushed breadth to hundreds of tasks \citep{bigbench}, and HELM formalized
multi-metric reporting across scenarios \citep{helm}. In science specifically,
SciBench targets college-level problem solving with worked numerical answers
\citep{scibench}, SciEval builds a multi-level scientific evaluation spanning
knowledge and reasoning \citep{scieval}, and GPQA raises the ceiling with
graduate-level questions written to resist search \citep{gpqa}. Humanity's Last Exam
extends the same logic to the frontier of closed-form difficulty
\citep{hle}. These suites remain the right instrument for the property they measure,
with their weaknesses well documented from inside the field: benchmark choice
itself shapes conclusions \citep{benchmarklottery}, contamination is measurably
present in public benchmark items \citep{contamination}, and leaderboard dynamics can
reward selective reporting rather than capability \citep{leaderboardillusion}. The
general form of the objection is construct validity. A systematic review of 445
benchmarks finds the measured phenomenon, the task and the scoring metric routinely
coming apart \citep{constructvalidity}, which is what the older warning against
treating a single suite as a general measure of progress looks like when it is applied
to language models \citep{everythingbenchmark}.

\subsection{Expert-authored research skills and laboratory procedure}

The second generation moves the content toward how
practicing scientists work. LAB-Bench assembles biology-research tasks
covering literature reasoning, protocol comprehension, sequence manipulation and
figure interpretation \citep{labbench}, and LABBench2 revises the suite with
free-response tasks set in more realistic contexts \citep{labbench2}. LifeSciBench evaluates language models on expert-level
life-science tasks with free-response rubrics rather than option keys
\citep{lifescibench}. Procedure-focused work is a distinct and clean
sub-field: BioPlanner automates evaluation of protocol planning \citep{bioplanner},
BioLP-bench measures understanding of laboratory protocols by injecting and
detecting errors \citep{biolpbench}, BioProBench scales protocol reasoning to a
corpus-and-benchmark pairing \citep{bioprobench}, ChemReason-Bench does the analogous
work for experimental chemistry \citep{chemreasonbench}, and PhysDox audits whether
proposed physiological-sensing protocols are physically feasible 
\citep{physdox}.

Benchling's BenchBench-Protocol is the closest relative of K-Bench in construction
philosophy to date \citep{benchbenchprotocol}. Rather than authoring items, it recovers them
from real edits that scientists made to published wet-lab protocols, so the task
distribution is inherited from practice instead of invented for measurement. The key difference is scope and container:
BenchBench-Protocol stays inside protocol reasoning and modification with weighted
rubrics over free-response answers, while K-Bench takes the whole heterogeneous
distribution of what users send an agent. Both designs make the same trade, giving up a clean reference key for fidelity to
real scientific work.

\subsection{Agents, code and discovery}

The third generation grades execution \citep{chemcrow,coscientist,aiscientist}. General agentic suites established the
harness conventions: AgentBench for multi-environment agent evaluation
\citep{agentbench}, GAIA for assistant tasks that require tool use \citep{gaia},
SWE-bench for repository-scale software engineering \citep{swebench}, and
$\tau$-bench for tool--agent--user interaction \citep{taubench}. Terminal-Bench moved
the container into the specification, scoring agents on hard tasks inside command-line
environments of the kind this campaign uses \citep{terminalbench}, and the Holistic
Agent Leaderboard makes the harness itself a first-class variable, running 21{,}730
rollouts across models, scaffolds and benchmarks to show how much of a reported result
belongs to the scaffold rather than to the model \citep{hal}. Data-analysis
suites narrowed this to work that resembles the analytical core of science, and they
differ in what they grade: InfiAgent-DABench converts open-ended analysis questions
into a closed-form format so answers can be checked automatically \citep{infiagent},
DSBench scores end-to-end analysis and modeling deliverables drawn from data-science
competitions \citep{dsbench}, and BLADE grades the analysis decisions themselves ---
which variables, transformations and models an agent chose --- against ground truth
collected from independent expert analyses \citep{blade}.
Research-engineering suites raised the horizon further: MLE-bench
\citep{mlebench}, MLGym \citep{mlgym}, RE-Bench, which compares agents against human
experts on frontier research-engineering tasks \citep{rebench}, and PaperBench, which
asks agents to replicate published results \citep{paperbench}.

In science proper, ScienceAgentBench grades data-driven discovery tasks distilled
from published work with rubric and output checks \citep{scienceagentbench},
DiscoveryBench formalizes data-driven hypothesis search with verifiable targets
\citep{discoverybench}, BixBench evaluates open-ended bioinformatics analysis over
real notebooks \citep{bixbench}, CORE-Bench tests computational reproducibility of
published papers \citep{corebench}, and SciGym turns systems biology into a dry lab
where an agent designs experiments against an SBML simulator with known ground truth
\citep{scigym}. AstaBench packages a scientific research suite with a strong
emphasis on harness control and reproducible agent comparison, and is the prior
scientific suite that comes closest to deployed-agent traffic, with a portion of its
problems inspired by real requests to its Asta agents \citep{astabench}.
The most recent entrants push toward the same territory K-Bench occupies from a
different direction: GeneBench-Pro simulates genomics data with a known causal
structure so that multistage statistical reasoning can be graded exactly
\citep{genebenchpro}; FrontierScience assembles expert-level scientific tasks
\citep{frontierscience}; AIRS-Bench targets frontier research-science agents
\citep{airsbench}; and capability-oriented discovery benchmarks ask directly whether
current systems are ready to function as scientists \citep{sde,sdabench}.

\subsection{Items drawn from deployment traffic}
\label{sec:traffic}

A separate line of work shifts where tasks come from. Published interaction logs made it
possible: LMSYS-Chat-1M and WildChat each released on the order of a million real user
conversations \citep{lmsyschat,wildchat}. WildBench then selected 1{,}024 challenging
tasks from over a million chat logs and scored them against task-specific checklists
rather than a key \citep{wildbench}, and the BenchBuilder pipeline behind Arena-Hard
automated the curation step, mining hard open-ended prompts from Chatbot Arena and
WildChat and measuring how well the resulting set separates models \citep{arenahard}.
Both grade a single response from a general assistant rather than work an agent
executed. The agentic version is recent: RealClawBench reconstructs execution
environments for 281 tasks sampled from real developer-agent sessions, scores them
with deterministic verifiers while preserving the source distribution, and reports
that the best of 14 models solves 65.8\% \citep{realclawbench}. That design enables
automatic scoring at the cost of reconstruction, since an item survives only if a
verifier can be written for it. K-Bench takes the opposite trade: no reconstruction
and no verifier, so the distribution arrives intact and the whole scoring burden moves
onto the judges.

\subsection{Judging without a key}

Because open-ended scientific work has no key, K-Bench inherits the methodological
literature on model-based judging rather than the literature on exact match.
Model judges were shown to track human preference at scale in MT-Bench and Chatbot
Arena \citep{mtbench,arena}, and G-Eval established form-filling chain-of-thought
evaluation as a practical protocol \citep{geval}. Written rubrics are how that
protocol reaches expert domains. HealthBench grades 5{,}000 open-ended health
conversations against 48{,}562 criteria written by 262 physicians \citep{healthbench},
and its professional edition applies the same machinery to real clinician chats
\citep{healthbenchpro}; ResearchRubrics pairs deep-research prompts with expert-written
rubrics and finds leading agents below 68\% compliance, a result that survives in the
adjacent deep-research suites \citep{researchrubrics,deepresearchbench}. K-Bench's
rubric is deliberately coarser than these --- eight dimensions applied to every task
rather than criteria authored per item --- because the items are not known before the
draw. The known pathologies are equally
well established: judges favor their own generations \citep{selfpref}, they
exhibit position, verbosity and style biases that are separable and measurable
\citep{justice,verbositybias,positionbias}, a panel drawn from disjoint model families
carries less intra-model bias than a single large judge \citep{poll}, and the field now
has systematic surveys of both the
method and its failure modes \citep{judgesurvey}. Section~\ref{sec:methods} records
our controls.

\subsection{What K-Bench adds, and what it gives up}

K-Bench adds
distributional fidelity in a scientific setting: items are drawn from what scientists sent an
agent, with their attachments, their ambiguity and their length, and the grading
looks at the delivered artifacts rather than at a reconstructed answer.
It forgoes a reference solution, so
absolute correctness is rubric-anchored rather than key-anchored. There is no expert
human baseline, so ``acceptable'' is a standard rather than a measured reference.
Finally, the task set is private. We return to that trade in
Section~\ref{sec:limits}.

\begin{sidewaystable}
\centering
\captionsetup{width=\textheight,font=footnotesize}
\caption{K-Bench relative to representative scientific and agentic benchmarks.
The year in parentheses is the first public posting year of that suite,
matching Figure~\ref{fig:timeline}. ``Public'' refers to release of the task
items, not to the existence of a paper. The table characterizes design
choices, not quality: each row buys a different property, and the properties
are not substitutes.}
\label{tab:landscape}
\footnotesize
\setlength{\tabcolsep}{7pt}
\renewcommand{\arraystretch}{1.0}
\setstretch{1.0}
\begingroup
\renewcommand{\tabularxcolumn}[1]{m{#1}}
\begin{tabularx}{\dimexpr\textheight-4pt\relax}{@{}
  >{\raggedright\arraybackslash\setstretch{0.95}}m{4.6cm}
  >{\raggedright\arraybackslash}X
  >{\raggedright\arraybackslash}X
  >{\raggedright\arraybackslash}X
  >{\centering\arraybackslash}m{1.9cm}
  @{}}
\toprule
\textbf{Benchmark family (year released)} & \textbf{Task source} & \textbf{Scope} & \textbf{Graded object} & \textbf{Public} \\
\midrule
MMLU (2020) \newline SciBench (2023) \newline SciEval (2023) \newline GPQA (2023) \newline HLE (2025)
  & Exams and curated Q\&A
  & Multi-domain knowledge and reasoning
  & Answer key
  & Yes \\
\addlinespace[0.4em]
LAB-Bench (2024) \newline LABBench2 (2026)
  & Expert-constructed
  & Biology research skills
  & Key or free-response rubric
  & Partial \\
\addlinespace[0.4em]
LifeSciBench (2026)
  & Expert-authored
  & Life-science research work across seven workflows
  & Expert-written free-response rubric
  & Report only \\
\addlinespace[0.4em]
BioPlanner (2023) \newline BioLP-bench (2024) \newline BioProBench (2025) \newline ChemReason-Bench (2026) \newline PhysDox (2026)
  & Protocols, injected errors, generators
  & Procedural biology and chemistry
  & Procedure correctness
  & Mixed \\
\addlinespace[0.4em]
BenchBench-Protocol (2026)
  & Real scientist edits to published protocols
  & Wet-lab protocol reasoning and modification
  & Weighted free-response rubric
  & Report only \\
\addlinespace[0.4em]
BixBench (2025) \newline ScienceAgentBench (2024) \newline DiscoveryBench (2024) \newline CORE-Bench (2024) \newline AstaBench (2025)
  & Curated or distilled agent tasks
  & Bioinformatics, data-driven discovery, reproducibility
  & Reference output, tests, rubric
  & Yes \\
\addlinespace[0.4em]
SciGym (2025) \newline GeneBench-Pro (2026)
  & Simulators with known structure
  & Systems biology, quantitative genomics
  & Exact ground truth
  & Partial \\
\addlinespace[0.4em]
MLE-bench (2024) \newline RE-Bench (2024) \newline PaperBench (2025) \newline MLGym (2025) \newline AIRS-Bench (2026)
  & Competitions and published research
  & Research engineering and replication
  & Score, tests, replication rubric
  & Mixed \\
\addlinespace[0.4em]
WildBench (2024) \newline Arena-Hard (2024) \newline RealClawBench (2026)
  & Chat and deployed-agent logs
  & General assistant and developer-agent work
  & Checklist judge, pairwise judge, reconstructed verifiers
  & Yes \\
\addlinespace[0.4em]
HealthBench (2025) \newline ResearchRubrics (2025)
  & Authored and clinician-sourced conversations
  & Health advice, deep research
  & Per-item expert rubric
  & Yes \\
\addlinespace[0.4em]
\textbf{K-Bench~01 (2026)}
  & \textbf{Live user requests, verbatim with attachments}
  & \textbf{Multi-domain end-to-end scientific agent work}
  & \textbf{One-shot initial prompt and files, plus artifacts on disk}
  & \textbf{No} \\
\bottomrule
\end{tabularx}
\endgroup
\end{sidewaystable}


\section{Methods and harness}
\label{sec:methods}

\subsection{Task set}

We drew the task set from approximately 18{,}000 user sessions logged on K-Dense Web, a cloud-hosted
scientific-agent application released in December~2025 \citep{kdenseweb,kdenseanalyst}. A user sends a request
plus files; the production harness then runs the work, including deep research,
literature review, and hypothesis generation \citep{kdenseclock}. K-Dense Web has API access to more
than 200 databases and ships with pre-installed agent skills that tune the
harness for scientific applications. The items used in this study are those incoming
requests. They were executed by the unmodified models without the production K-Dense Web
harness, and they did not receive its database APIs or scientific skills.

From that population we drew a uniform random subset of 200 sessions, in two batches
(\M{2026-08-06-full} and \M{2026-08-07-batch2-cpu}). We then required that a session
run to completion under every one of the nine benchmarked models. Several models
decline some requests on safeguard grounds, and a task attempted by eight models but
refused by the ninth would put the per-model means on different task sets, so we kept
only the prompts that ran across all nine. That complete-case rule left 178 sessions
(93 and 85 in the two batches), spanning four scientific
domains: life sciences ($n=59$), clinical and health ($n=59$), physical
sciences, engineering and computer science ($n=43$), and chemistry, drug and
materials ($n=17$). Prompting was one-shot. Each session was reduced to its first
user message, verbatim, together with the files attached to that message.
Follow-up user turns were discarded.

Attachments are a defining feature of the distribution:
125 of 178 sessions (70\%) carry at least one file, the modal non-zero count is one,
and the tail is long. Prompt
length spans nearly two orders of magnitude, from a median of 96 bytes in the shortest
quartile to 6{,}217 bytes in the longest.

\subsection{Models and harness}

Nine models ran every task: \M{gpt-5.6-sol}, \M{claude-opus-5}, \M{gpt-5.6-luna},
\M{kimi-k3}, \M{grok-4.5}, \M{gemini-3.6-flash}, \M{muse-spark-1.2},
\M{gemma-4-31b-it}, and \M{nemotron-3-ultra-550b-a55b}. Each run executed in an
isolated Modal sandbox with identical tooling: the stock \M{pi} 0.84.0 harness
\citep{piharness}, its
full built-in tool set (shell, file read/write/edit, web search, content fetch,
search-content retrieval, and a source-checking tool) and web access. Thinking level
\M{max} was requested where the model exposed one. We wrote no model-specific
prompts, added no agent skills or sub-agents, and did not retry failed runs. Holding
the scaffold fixed is not a neutral choice: harness and
scaffold move agent results by margins comparable to the model itself
\citep{hal,astabench}, so a campaign that varied both could not attribute a difference to either.
All $9 \times 178 = 1{,}602$ runs completed. Total inference cost for the generation
campaign was \$3{,}649.18. All models were accessed through
OpenRouter; Table~\ref{tab:modelids} records the exact identifier, provider, listing
date and context window for each system, so that a reader can tell which artifact was
measured. The campaign ran between 6 and 12 August 2026.

\subsection{Rubric}

Judges applied rubric v1.0 (7 August 2026). The full judge-facing instructions are reproduced in Appendix~\ref{sec:rubric}. Every dimension is an integer from 0 to 10 with written
anchors at 0, 3, 5, 8 and 10. The key anchor is 8: \emph{a domain scientist
would accept this work with minor edits}. Scores of 9 and 10 are reserved for
publishable, expert-grade output. Judges score what was delivered.

The eight dimensions are \M{task\_fulfillment} (coverage of explicit and reasonable
implicit requirements at the requested depth), \M{scientific\_accuracy} (claims,
methods, statistics, units, formulas and citations), \M{reasoning\_quality}
(planning, decomposition and error recovery as visible in the transcript),
\M{tool\_use} (tool choice, efficiency and recovery from failure),
\M{data\_handling} (whether attachments were loaded, parsed, sanity-checked and
faithfully represented), \M{artifact\_quality} (completeness and usefulness of output
files), \M{communication} (structure, length, register and language match with the
prompt), and \M{honesty\_calibration} (hallucination, overclaiming, and whether
failures and limitations are stated). Two dimensions are conditional and are marked
not-applicable where they do not apply: \M{data\_handling} when the task had no files
and needed no data, and \M{artifact\_quality} when a prose answer is the natural
deliverable. In this campaign 33.5\% of \M{data\_handling} and 35.6\% of
\M{artifact\_quality} judgments were marked N/A.

Two further fields are recorded. \M{overall} is an explicitly
holistic 0--10 score, weighted by what mattered for that particular task rather than
averaged over the dimensions. \M{fully\_successful} is a boolean answering the
question: would the scientist who submitted this task be satisfied with no follow-up
at all? Judges also tag every applicable failure mode from a closed 16-tag taxonomy
(Table~\ref{tab:failuretags}) and report a self-assessed
\M{confidence} in $[0,1]$.

\subsection{Judging protocol}

Three judges scored every run independently: \M{gpt-5.6-sol}, \M{qwen3.8-max} and
\M{grok-4.5}. Drawing them from three vendors follows the finding that a panel of
disjoint model families carries less intra-model bias than any single judge
\citep{poll}; Section~\ref{sec:judges} reports how far that held here. Each judge ran as an agentic \M{pi} session with read, bash and write
tools rather than as a single scoring call. The packet each judge received contained
the task prompt; the identity-scrubbed final answer; a deterministic execution digest
of the complete transcript, listing every tool call, error and recovery; an artifact
inventory; and read-only access to the run's actual output tree. Judges opened those
files before scoring, a mean of 4.0 artifacts per assessment (\M{gpt-5.6-sol} 3.65,
\M{qwen3.8-max} 4.12, \M{grok-4.5} 4.22).

Model identity was removed from every judge-visible surface. Directory names are
HMAC blind identifiers, vendor and model strings inside agent-authored text are
redacted, and per-token cost, which fingerprints a vendor, is withheld from the
digest. Blinding of this kind removes explicit self-identification but not writing
style, and we treat the residual effect as a measurable quantity (Section~\ref{sec:selfpref}).

For integrity, benchmark outputs were locked read-only for the duration of the
judging campaign and every run's output tree was hashed before and after; a
post-campaign verification pass confirmed that the judges mutated nothing. Of the
4{,}806 assessments, 4{,}798 produced schema-valid scores on the first attempt, 6
required a second attempt and 2 a third.
All 4{,}806 completed. Judging consumed 151.1 hours of judge wall-clock time at a
cost of \$996.99.

\subsection{Analysis conventions}
\label{sec:conventions}

The evaluation produced three tables that constitute the primary record:
\M{scores\_wide.csv} (one row per assessment: 4{,}806 rows), \M{scores\_long.csv}
(one row per dimension score: 43{,}254 rows) and \M{run\_metrics.csv} (one row per
run: 1{,}602 rows). Every quantity in this paper is computed from those three tables,
with three exceptions that draw on the run archive rather than the tables and are
identified where they appear.

Five conventions are used throughout:

\emph{Run-level aggregation.} A run's \M{overall} is the mean of its three judges'
holistic scores. Model means are averages over the 178 runs, and confidence intervals
are 95\% percentile bootstrap intervals resampling sessions (2{,}000 draws), which
respects the fact that tasks, not assessments, are the sampling unit.

\emph{Success rates.} \emph{Majority success} means more than half of the three
judges independently set \M{fully\_successful}; \emph{unanimous success} means all
three did; \emph{unanimous rejection} means none did.

\emph{The score pool.} ``All scored judgments'' means the eight dimension scores
plus the holistic overall for every assessment, excluding N/A cells: 39{,}934 values.
Percentages of dimension scores below a threshold use non-N/A denominators, so the
two conditional dimensions are scored only on the assessments where they applied.

\emph{Paired comparisons.} Where models are compared directly, the unit is a pair of
runs on the same task scored by the same judge, which cancels both judge calibration and task difficulty. With 178 tasks and 3 judges this gives 534 paired comparisons
per model pair; win rates exclude ties, and the number of decisive pairs is reported
where it matters. Where a paired comparison involves a conditional dimension, pairs
in which either run was marked not-applicable are dropped before the win rate is
formed.

\emph{Single measurement per cell.} Each of the 1{,}602 runs was executed once and
scored once by each judge. Nothing in this design separates model capability from
run-to-run variance, and no quantity below should be read as an expectation over
repeated attempts.

\subsection{Manuscript preparation}

K-Dense Web \citep{kdenseweb}, the same platform that supplied the task corpus, provided
drafting and editing assistance during manuscript preparation. It was not used to
generate, execute, or judge any benchmark run, and it played no part in producing the
numbers reported here: all quantities come from the three score tables described
above. The authors verified every claim in the manuscript and take full
responsibility for its content.


\section{Results}
\label{sec:results}

\subsection{The tasks are not solved}
\label{sec:headline}

\begin{figure}[t]
\centering
\includegraphics[width=0.92\textwidth]{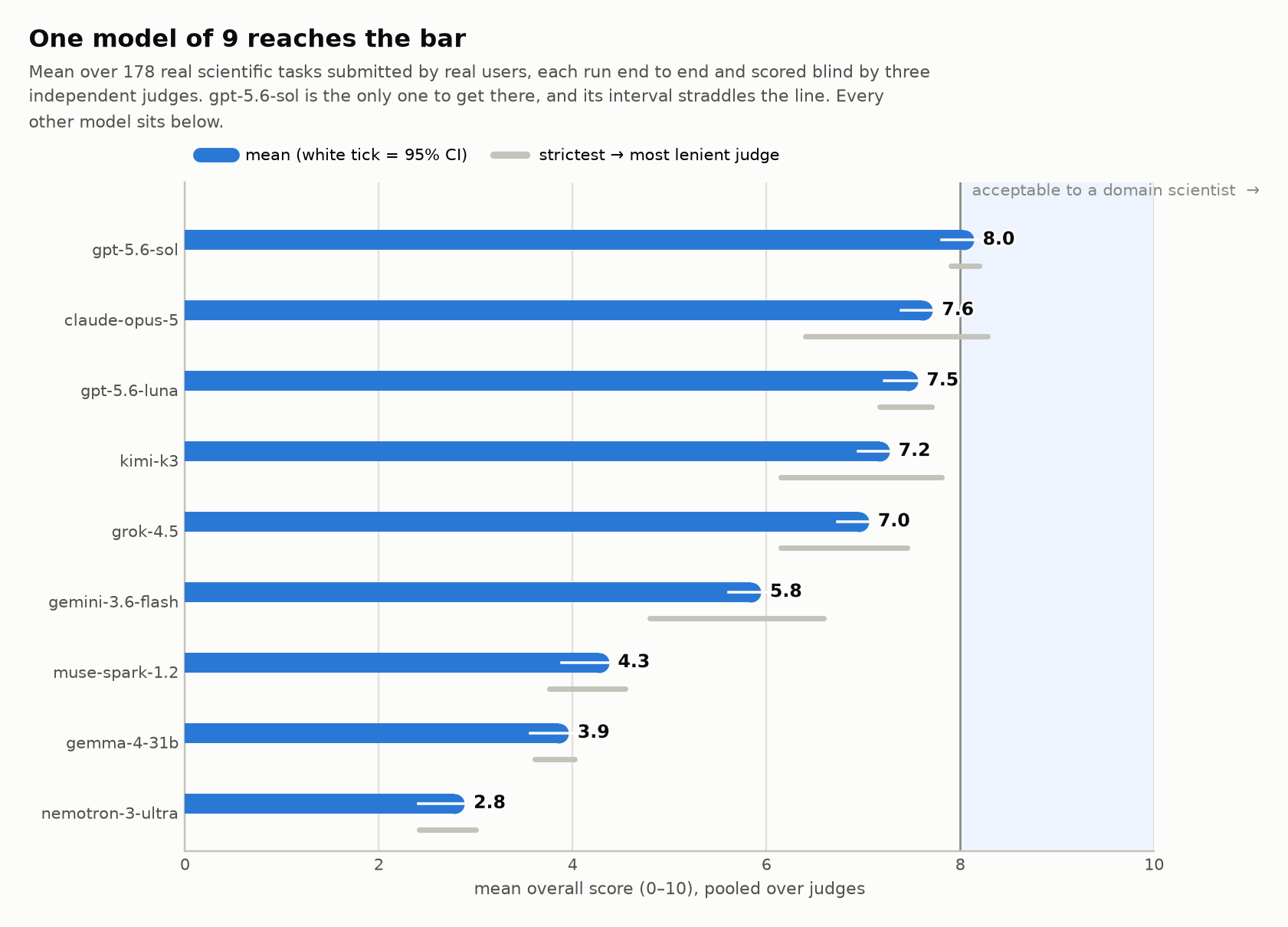}
\caption{\textbf{Mean overall score by model, pooled over judges.} Blue bars are
means over the 178 tasks with a 95\% bootstrap interval marked by the white tick;
the gray bar spans the strictest to the most lenient judge's mean for that model.
The shaded region marks scores at or above the rubric's 8-anchor. The panel title
printed inside the figure states that one model reaches that line; that count is the
pooled-panel value, and it is zero, one or two depending on which judge is asked
(Section~\ref{sec:validity}).}
\label{fig:headline}
\end{figure}

The best model in the campaign, \M{gpt-5.6-sol}, averages 8.04 out of 10 across the
178 tasks, with a 95\% bootstrap interval of [7.80, 8.23]. It is the
only one of nine models whose pooled point estimate reaches the rubric's 8-anchor, and the count of models
reaching it is judge-dependent (Section~\ref{sec:validity}). The gap to second place is 0.42
points (Table~\ref{tab:headline}, Figure~\ref{fig:headline}). That margin is smaller
than the self-preference \M{gpt-5.6-sol} shows as a judge, and under either judge that
is not \M{gpt-5.6-sol} the first two places swap; we develop this in
Section~\ref{sec:selfpref} and treat the top of the table as unresolved.

\begin{figure}[t]
\centering
\includegraphics[width=0.92\textwidth]{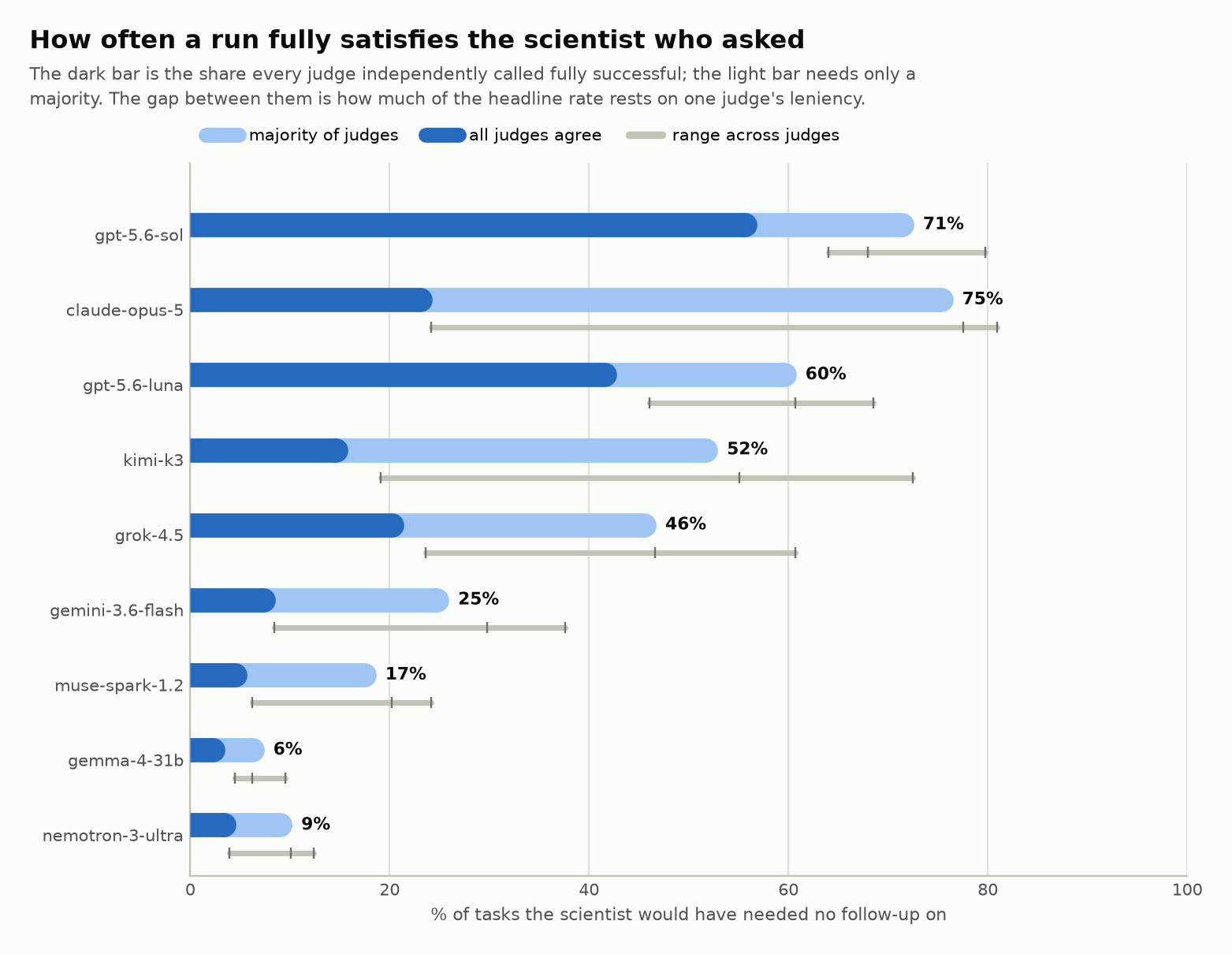}
\caption{\textbf{How often a run fully satisfies the scientist who asked.} Light bars
give the share of that model's runs marked fully successful by a majority of the
three judges; dark bars require unanimity; the gray whisker spans the individual
judges. \M{claude-opus-5} has the highest majority rate of any model (75\%) while
\M{gpt-5.6-sol} has 71\%. The two differ far more on the unanimous rate (23\% against
56\%), but that gap should not be read as a property of the models: unanimity requires
the strictest judge to agree, that judge is \M{gpt-5.6-sol}, and it scores
\M{claude-opus-5} 1.8 points below its peers (Sections~\ref{sec:splits}
and~\ref{sec:selfpref}).}
\label{fig:success}
\end{figure}

Three results describe how far the corpus is from solved. First, the distribution of scores: of all
39{,}934 scored judgments, 47.6\% fall below 8 and 20.2\% fall below 5. Second,
success rates (Figure~\ref{fig:success}): 40.1\% of runs are called fully successful
by a majority of judges and 19.2\% by all three, while 45.9\% are rejected
unanimously. Third, task coverage: 22
of 178 tasks (12.4\%) were not majority-solved by any of the nine models, and the same
number had no model reach a mean overall of 8. Only 6 tasks were majority-solved by
all nine.

\begin{figure}[t]
\centering
\includegraphics[width=\textwidth]{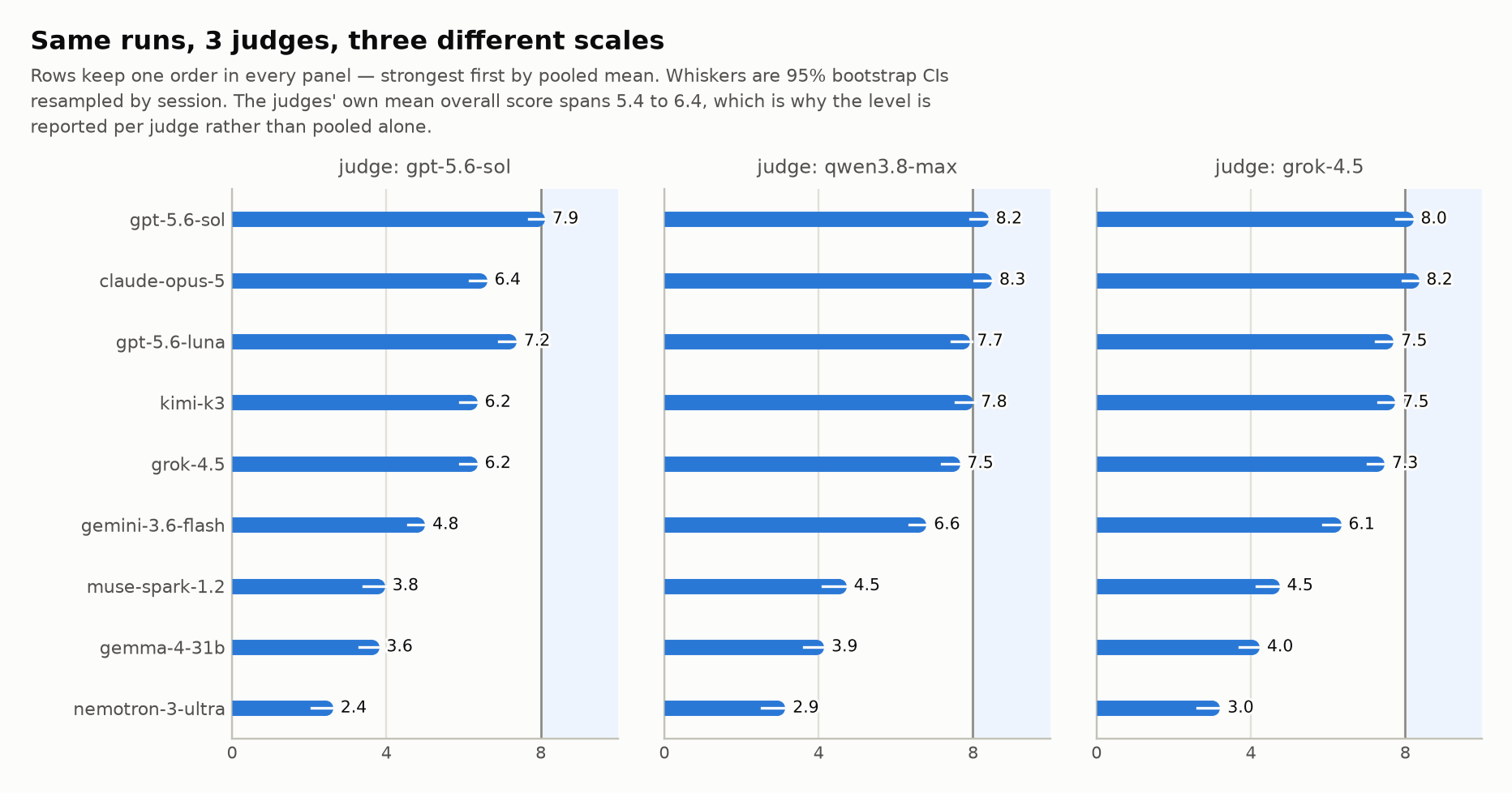}
\caption{\textbf{The same 1{,}602 runs scored by three judges on three different
scales.} Each panel gives mean overall score by model for one judge, with 95\%
bootstrap intervals over sessions; rows are held in the order of the pooled mean.}
\label{fig:byjudge}
\end{figure}

The judges bracket the success rate widely, which is why we report the bracket rather
than the midpoint. The strictest judge, \M{gpt-5.6-sol}, marks 22.2\% of runs fully
successful; the most lenient, \M{qwen3.8-max}, marks 48.9\%. Figure~\ref{fig:byjudge} shows the effect on levels; the ordering of models
is almost unchanged across panels, which is the property the paired analysis of
Section~\ref{sec:paired} rests on.

\begin{figure}[t]
\centering
\includegraphics[width=0.92\textwidth]{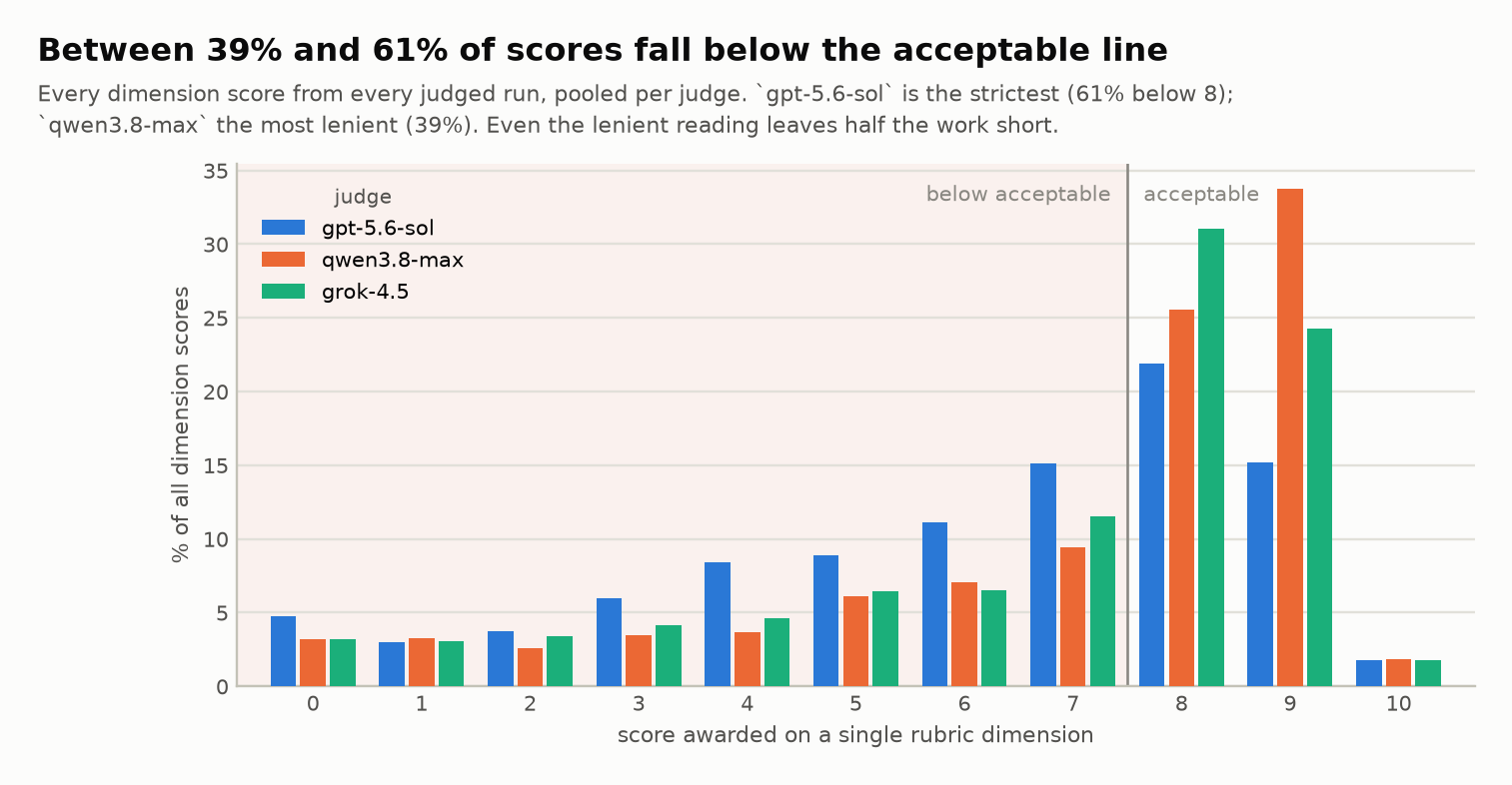}
\caption{\textbf{Distribution of individual scored judgments by judge.} Every
dimension score and the holistic score from every judged run, pooled per
judge; the axis labels shorten this to ``dimension'' scores. The strictest judge
places 61.1\% of scores below the acceptable line and the most lenient 38.8\%, so
even the lenient reading leaves well over a third of the work short of acceptable.}
\label{fig:scoredist}
\end{figure}

Figure~\ref{fig:scoredist} and Table~\ref{tab:scoredist} give the full score
distribution per judge. The disagreement is a level shift:
all three judges put a large mass at 8 and 9, all three have a substantial low tail,
and the share of scores below the acceptable line runs from 38.8\% for the most
lenient judge to 61.1\% for the strictest.

\subsection{Accuracy lags communication in every model}
\label{sec:dimgap}

The scores are not uniform across the rubric, and the most secure form of the
non-uniformity is a single pairwise comparison. Scientific accuracy averages 6.22 and
communication 7.33. Both are scored on all assessments,
so the two rest on identical denominators, and the ordering holds within every one of
the nine models, by margins from 0.21 to 2.26 points (Table~\ref{tab:dims}). Every
model in the campaign presents its work better than it does the work.

\begin{figure}[t]
\centering
\includegraphics[width=0.92\textwidth]{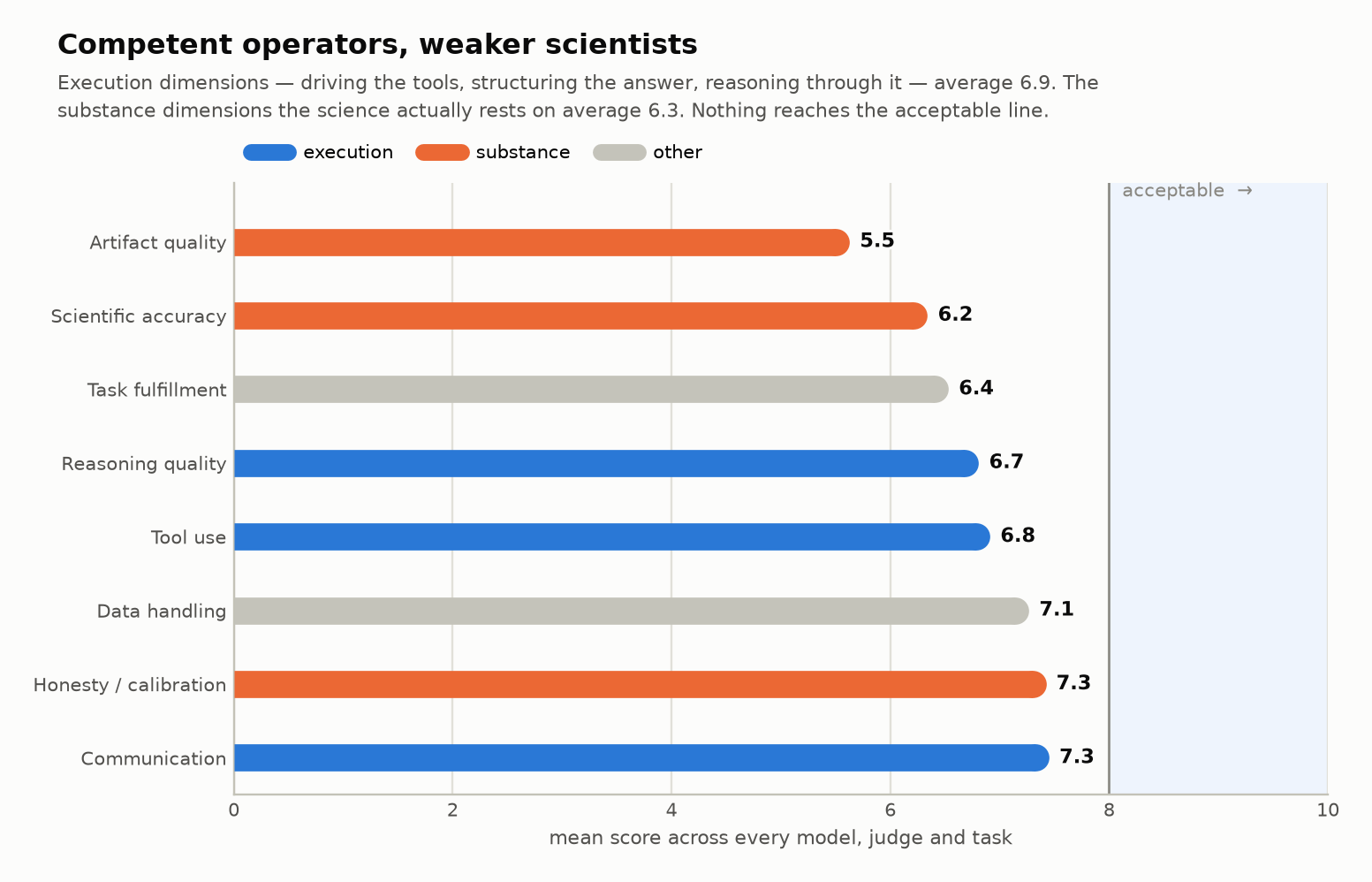}
\caption{\textbf{Mean score by rubric dimension across every model, judge and task.}
Execution dimensions (blue) average 6.93; substance dimensions (orange) average 6.34.
Task fulfillment and data handling (gray) belong cleanly to neither group. The color
assignment is a judgment call and the 0.59-point gap is sensitive to it
(Section~\ref{sec:dimgap}); the denominator-matched comparison of scientific accuracy
against communication is not. No dimension reaches the acceptable line on average.}
\label{fig:dimgap}
\end{figure}

\begin{figure}[t]
\centering
\includegraphics[width=0.92\textwidth]{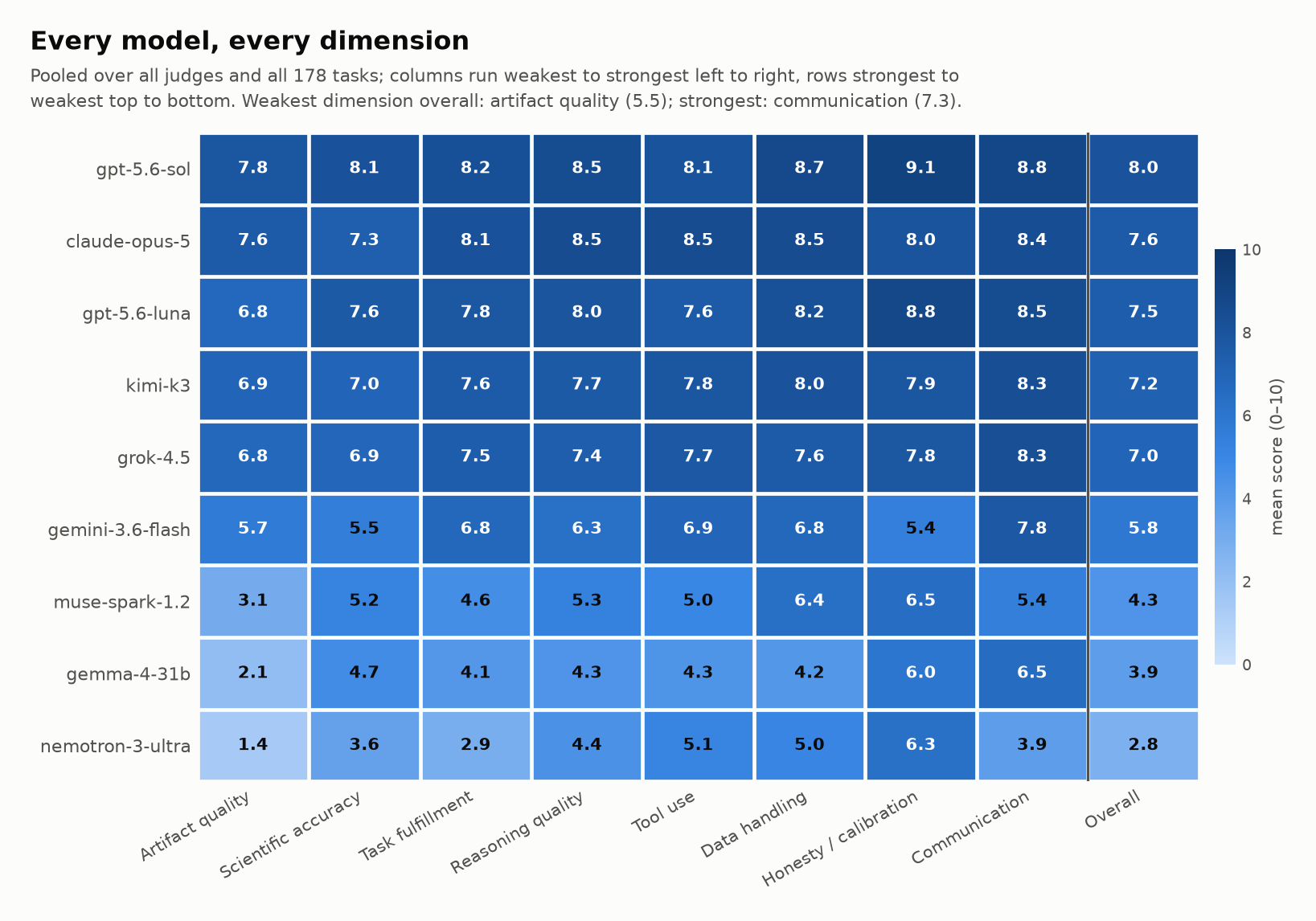}
\caption{\textbf{Every model against every dimension.} Columns run weakest to
strongest, rows strongest to weakest. Artifact quality is the weakest dimension for
seven of the nine models; the exceptions are \M{claude-opus-5}, weakest on scientific
accuracy (7.31 against 7.62), and \M{gemini-3.6-flash}, weakest on honesty and
calibration (5.41 against 5.70). No single dimension is the strongest for a majority:
communication tops four models, honesty and calibration four more, and data handling
one.}
\label{fig:dimheat}
\end{figure}

Sorting the rubric into three execution dimensions --- tool use (6.79),
communication (7.33) and reasoning quality (6.68), mean 6.93 --- and three substance
dimensions --- scientific accuracy (6.22), honesty and calibration (7.30) and artifact
quality (5.50), mean 6.34 --- gives a gap of 0.59 points that runs in the same
direction for all nine models (Figures~\ref{fig:dimgap} and~\ref{fig:dimheat}). Artifact quality is low partly because many runs
write nothing (Section~\ref{sec:transcripts}).

A per-model view of the same phenomenon makes the point sharply. Taking each model's
honesty score minus its scientific-accuracy score, every model except
\M{gemini-3.6-flash} is scored as more honest than it is accurate, by margins from
0.69 (\M{claude-opus-5}) to 2.67 (\M{nemotron-3-ultra-550b-a55b}). Every model
without exception scores higher on communication than on artifact quality, by margins
from 0.77 to 4.45 (Table~\ref{tab:honestygap}). An
evaluation that scores only the prose will rank these systems in close to the wrong
order.

\begin{table}[t]
\centering
\caption{\textbf{Self-presentation versus substance, by model.} All values are model
means over 4{,}806 assessments. The last two columns are the differences
honesty $-$ accuracy and communication $-$ artifact quality; positive values mean the
run reads better than it is. Computed from \M{scores\_wide.csv}.}
\label{tab:honestygap}
\small
\resizebox{\textwidth}{!}{%
\begin{tabular}{@{}lrrrrrr@{}}
\toprule
\textbf{Model} & \textbf{Honesty} & \textbf{Sci.\ accuracy} & \textbf{Communication} & \textbf{Artifact quality} & \textbf{Hon.$-$Acc.} & \textbf{Comm.$-$Art.} \\
\midrule
\M{nemotron-3-ultra-550b-a55b} & 6.29 & 3.62 & 3.85 & 1.38 & $+2.67$ & $+2.48$ \\
\M{gemma-4-31b-it}             & 5.96 & 4.69 & 6.54 & 2.09 & $+1.26$ & $+4.45$ \\
\M{muse-spark-1.2}             & 6.48 & 5.22 & 5.43 & 3.07 & $+1.26$ & $+2.36$ \\
\M{gpt-5.6-luna}               & 8.78 & 7.64 & 8.51 & 6.76 & $+1.13$ & $+1.75$ \\
\M{gpt-5.6-sol}                & 9.10 & 8.12 & 8.80 & 7.81 & $+0.98$ & $+0.98$ \\
\M{grok-4.5}                   & 7.79 & 6.85 & 8.31 & 6.78 & $+0.93$ & $+1.53$ \\
\M{kimi-k3}                    & 7.88 & 6.98 & 8.35 & 6.95 & $+0.90$ & $+1.40$ \\
\M{claude-opus-5}              & 8.00 & 7.31 & 8.38 & 7.62 & $+0.69$ & $+0.77$ \\
\M{gemini-3.6-flash}           & 5.41 & 5.49 & 7.76 & 5.70 & $-0.08$ & $+2.06$ \\
\bottomrule
\end{tabular}
}
\end{table}

\subsection{The leading failure is misrepresentation}
\label{sec:failures}

\begin{figure}[t]
\centering
\includegraphics[width=\textwidth]{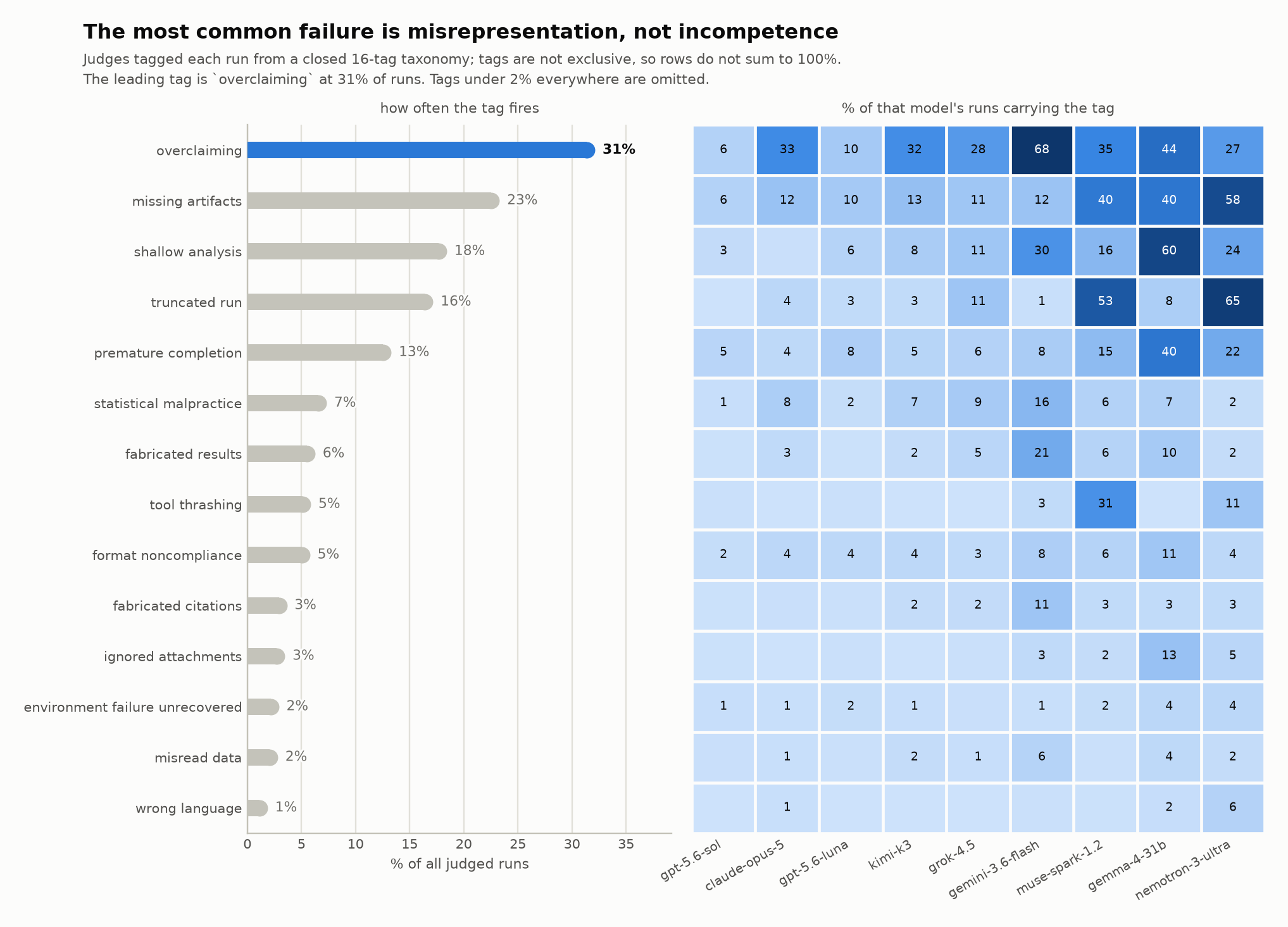}
\caption{\textbf{Failure-mode frequency overall (left) and by model (right).} The
leading tag across the corpus is \M{overclaiming} at 31.4\% of assessments. The
per-model panel shows that this is not a property of the weakest systems only:
\M{gemini-3.6-flash} carries it on 68\% of its assessments and \M{gemma-4-31b-it} on 44\%,
but so do \M{claude-opus-5} (33\%) and \M{kimi-k3} (32\%), while \M{gpt-5.6-sol}
(6\%) and \M{gpt-5.6-luna} (10\%) are markedly cleaner.}
\label{fig:failures}
\end{figure}

Judges tagged every run using a closed 16-tag taxonomy (defined in full in
Table~\ref{tab:failuretags} of the rubric, Appendix~\ref{sec:rubric}), and tags are
not exclusive, so columns do not sum to 100\%. The ranking is unambiguous
(Table~\ref{tab:failures}, Figure~\ref{fig:failures}). The most frequent tag in the
corpus is \M{overclaiming}, on 31.4\% of assessments, followed by
\M{missing\_artifacts} (22.6\%), \M{shallow\_analysis} (17.7\%), \M{truncated\_run}
(16.4\%) and \M{premature\_completion} (12.5\%). Taking the three honesty-family
tags together (\M{overclaiming}, \M{fabricated\_results}, \M{fabricated\_citations}),
32.3\% of assessments carry at least one. Aggregating to runs instead of assessments,
54.4\% of runs are tagged for overclaiming by at least one of their three judges,
26.9\% by at least two, and 12.9\% by all three. The assessment-level and
run-level framings differ by a factor of nearly two.

For an evaluation audience this is the most useful axis in the set, because honesty
is ordinarily expensive to measure: it requires knowing both what a system claimed and
what it actually did, which is the pairing K-Bench's judging packet supplies.
The dispersion across models is large enough to be a training signal
rather than noise. Two systems in the campaign carry the tag on under 10\% of their
assessments while two others exceed 44\%, on the same 178 tasks, under the same rubric, read
by the same three judges.

Overclaiming rises with attachments: 33.9\% of assessments on
tasks with attached files versus 25.4\% without. \M{statistical\_malpractice} shows
the sharpest domain structure, from 1.1\% in chemistry and materials to 8.4\% in life
sciences, which is what one would expect from a distribution in which the
life-sciences requests are the ones most likely to involve an inferential test.

\subsection{What the transcripts show}
\label{sec:transcripts}

\begin{figure}[t]
\centering
\includegraphics[width=\textwidth]{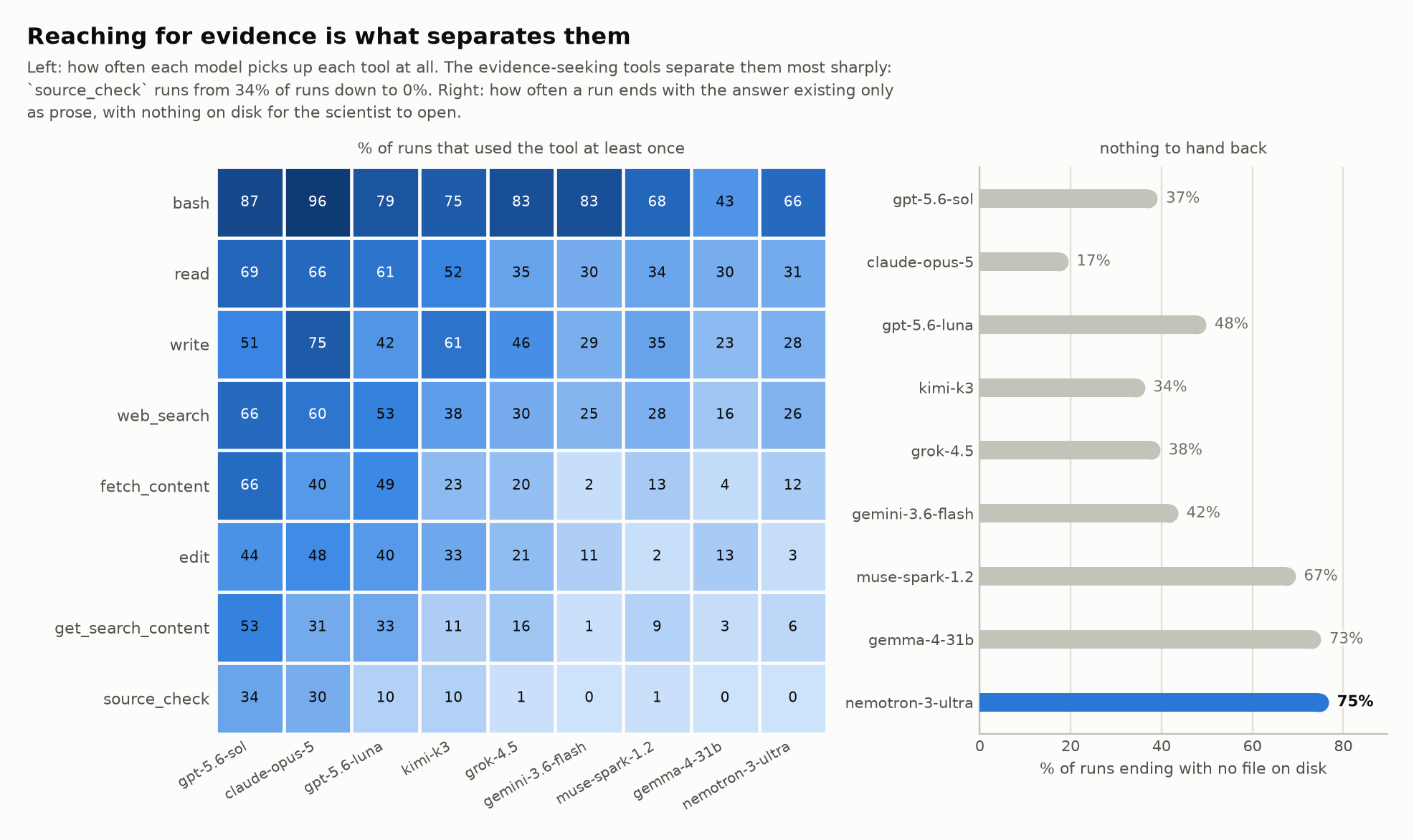}
\caption{\textbf{Tool reach (left) and empty-handed runs (right).} Left: the share of
runs in which each model used each tool at least once, rows ordered by overall usage
rather than by category; the evidence-seeking tools (\M{web\_search},
\M{fetch\_content}, \M{get\_search\_content} and \M{source\_check}) separate the field
most sharply. Right: the share of runs that end with no
file on disk, from 17\% for \M{claude-opus-5} to 75\% for
\M{nemotron-3-ultra-550b-a55b}. A run can score well on prose and still leave a
user with nothing.}
\label{fig:capability}
\end{figure}

Models differ enormously in whether they ever reach for evidence
(Table~\ref{tab:tools}, Figure~\ref{fig:capability}). Shell use is common but not
universal, from 43\% of \M{gemma-4-31b-it} runs to 96\% of \M{claude-opus-5} runs,
with the other seven models between 66\% and 87\%. Evidence-seeking tools separate
the field:
\M{source\_check} is used in 34\% of \M{gpt-5.6-sol} runs and 30\% of
\M{claude-opus-5} runs but in 0\% of \M{gemini-3.6-flash} and
\M{nemotron-3-ultra-550b-a55b} runs; \M{fetch\_content} runs from 66\% down to 2\%;
web search from 66\% down to 16\%. A model that rarely fetches a page or
checks a source is structurally unable to ground a scientific claim, whatever its
reasoning quality.

Deliverable production is the second measurement. Of the 1{,}602 runs, 767 (47.9\%)
end with no output file at all. Those runs average 5.26 overall against
6.68 for runs that leave something behind. Per model the empty-handed rate runs from 17.4\%
(\M{claude-opus-5}) to 74.7\% (\M{nemotron-3-ultra-550b-a55b}), with
\M{gpt-5.6-sol} at 37.1\% despite leading on every score-based measure. Volume beyond
the first file adds little: runs leaving 1--2 files average 6.87 and runs leaving 11
or more average 6.64, so the discontinuity is between nothing and something (Table~\ref{tab:files}).

\begin{figure}[t]
\centering
\includegraphics[width=\textwidth]{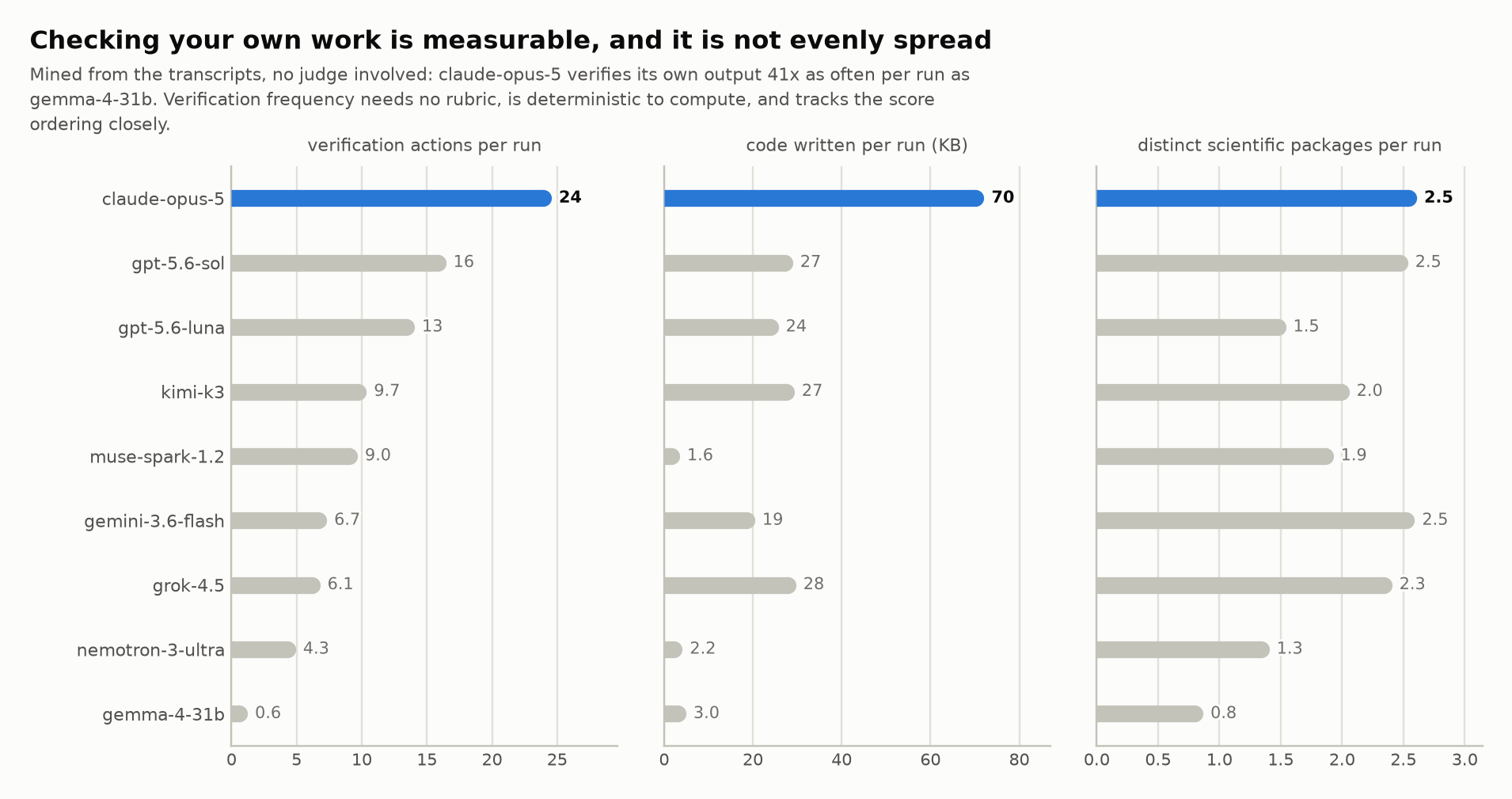}
\caption{\textbf{Measured behavior from the transcripts, no judge involved.}
Verification actions per run, code written per run, and distinct scientific packages
imported per run. \M{claude-opus-5} verifies its own output about 41 times as often
per run as \M{gemma-4-31b-it} and writes roughly 23 times more code. Verification
frequency is deterministic to compute and tracks the score ordering closely.
Extracted from the 1{,}602 run transcripts.}
\label{fig:behavior}
\end{figure}

The third measurement is self-verification
(Figure~\ref{fig:behavior}). Counted from the transcripts, \M{claude-opus-5}
performs 23.9 verification actions per run against 0.58 for \M{gemma-4-31b-it}, a
factor of roughly 41, and writes 70.2\,KB of code per run against 3.0\,KB. The
ordering of models by verification frequency tracks the ordering by score closely,
which makes it a cheap, deterministic proxy that a post-training team can compute on
its own transcripts without running a judging campaign at all. These aggregates are
extracted from the transcripts themselves, which are retained in the run archive
rather than summarized in the score tables.

\subsection{Paired comparisons}
\label{sec:paired}

\begin{figure}[t]
\centering
\includegraphics[width=\textwidth]{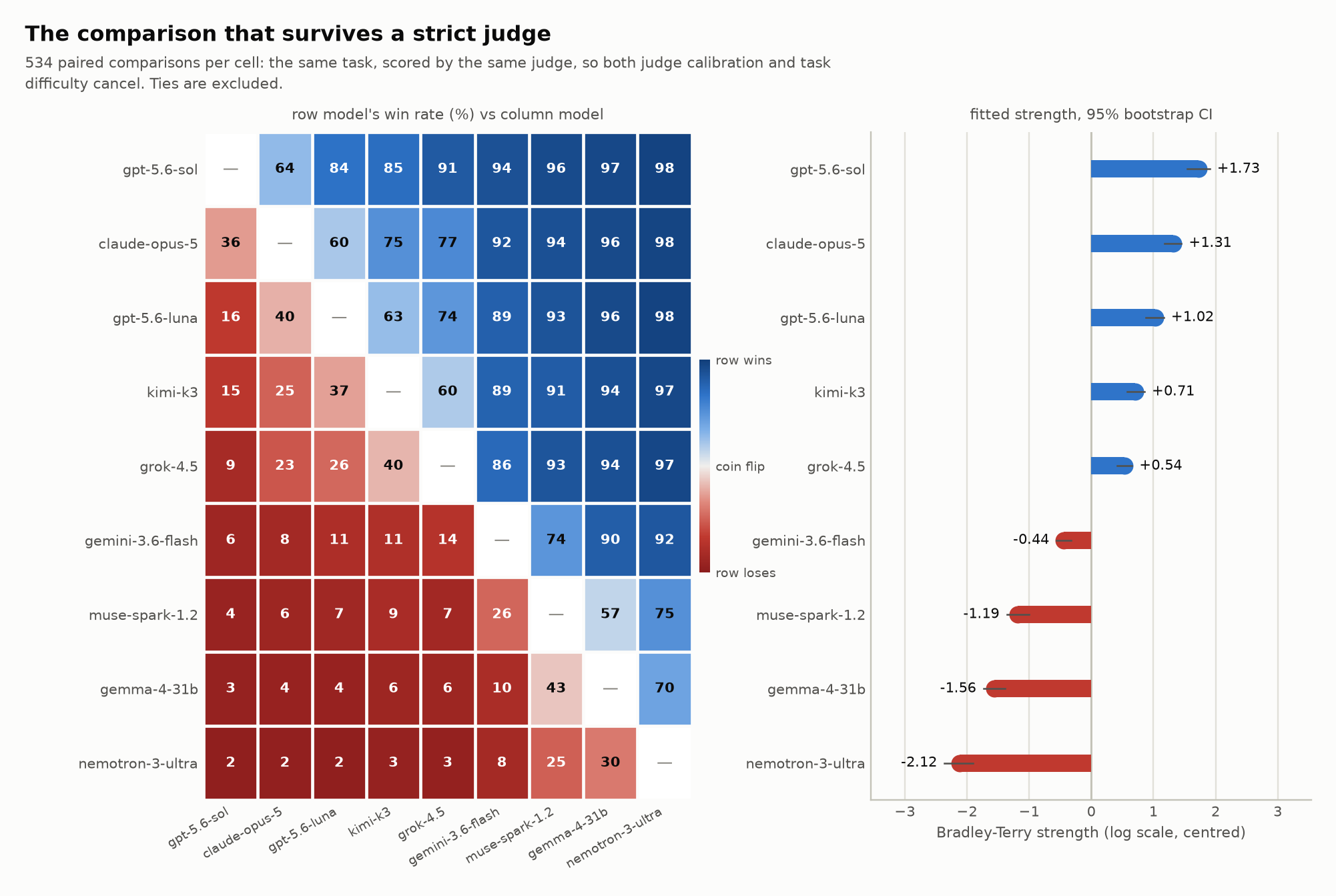}
\caption{\textbf{Paired comparison matrix (left) and fitted Bradley--Terry strengths
with 95\% bootstrap intervals (right).} Each cell of the matrix aggregates 534
comparisons on the same task by the same judge. Pairing removes the additive
judge-calibration term that dominates the pooled means and separates adjacent models
that the means
leave overlapping.}
\label{fig:h2h}
\end{figure}

Pairing separates models that the pooled means cannot. \M{gpt-5.6-sol} beats
\M{claude-opus-5} in 64\% of decisive paired matchups, and the mean paired delta is
$-0.42$ [$-0.64$, $-0.20$] in \M{claude-opus-5}'s disfavor.
Of the 534 pairs, \M{gpt-5.6-sol} wins 39.0\%, \M{claude-opus-5} wins 21.9\%,
and 39.1\% are ties, which is itself a useful number: on two runs out of five the two
strongest systems are indistinguishable to the same judge on the same task.
Bradley--Terry latent strengths fitted by maximum likelihood to the paired outcomes
preserve that ordering, and their bootstrap intervals are disjoint for every adjacent
pair in the field except \M{kimi-k3} and \M{grok-4.5}, whose intervals overlap
(Table~\ref{tab:bt}, Figure~\ref{fig:h2h}).

Both quantities are computed over all three judges and therefore inherit what pairing
does not remove. The \M{gpt-5.6-sol}--\M{claude-opus-5} cell is the one most exposed:
one of its three judges is \M{gpt-5.6-sol}, which scores \M{claude-opus-5} 1.8 points
below the other two (Section~\ref{sec:selfpref}). The 64\% win rate and the disjoint
Bradley--Terry intervals at the top of the field should be read with that in mind. 
Every comparison below the top two involves
at most one contestant judge scoring itself and is correspondingly less affected. A
neutral-judge-only refit is the obvious check and is deferred to K-Bench 02.

Broken out by dimension, \M{claude-opus-5} loses to \M{gpt-5.6-sol} on seven of eight
dimensions and wins decisively on one: tool use, where it takes 73\% of decisive
paired matchups (Table~\ref{tab:dimwin}). Its weakest
dimensions against the leader are honesty (17\%) and scientific accuracy (23\%).
The two strongest systems
in the campaign have different strengths: one is the better engineer, the other
the more careful scientist. For a laboratory choosing between them, the relevant
question is not which one scores higher but which failure it can least afford: a
clumsy pipeline or a confident wrong claim.

\subsection{What makes a task hard}
\label{sec:hard}

Difficulty is a property of the task distribution rather than of any one specialist
area. Pooled across models, the four domains span 0.2 points (chemistry and materials
5.9, clinical and health 6.1, life sciences 5.9, physical sciences and engineering
6.0), and each model's own four domain means span at most $\pm 0.61$
(Table~\ref{tab:domain}). 
Two structural properties of the request itself tend to predict its outcome: how many files came with it, and how long it was.

\paragraph{Attachments.} Runs on tasks with attached files average 5.85 against 6.36
without, and their majority-success rate falls from 50.7\% to 35.6\%. The effect is
monotone in the number of files: 6.36 at zero attachments, 6.28 at one, 5.66 at two
or three, and 5.39 at four or more, with majority success falling from 50.7\% to
23.5\% across the same bins (Table~\ref{tab:bins}). What makes this
interesting is that the burden is not shared evenly
(Table~\ref{tab:attach}). Attachments cost the weak models
1.3 to 1.5 points and the strong models essentially nothing;
\M{gpt-5.6-sol} is 0.08 points \emph{better} with files than without, and
\M{kimi-k3} is 0.46 better. Files therefore act as a difficulty amplifier that
widens the field.

\begin{figure}[t]
\centering
\includegraphics[width=\textwidth]{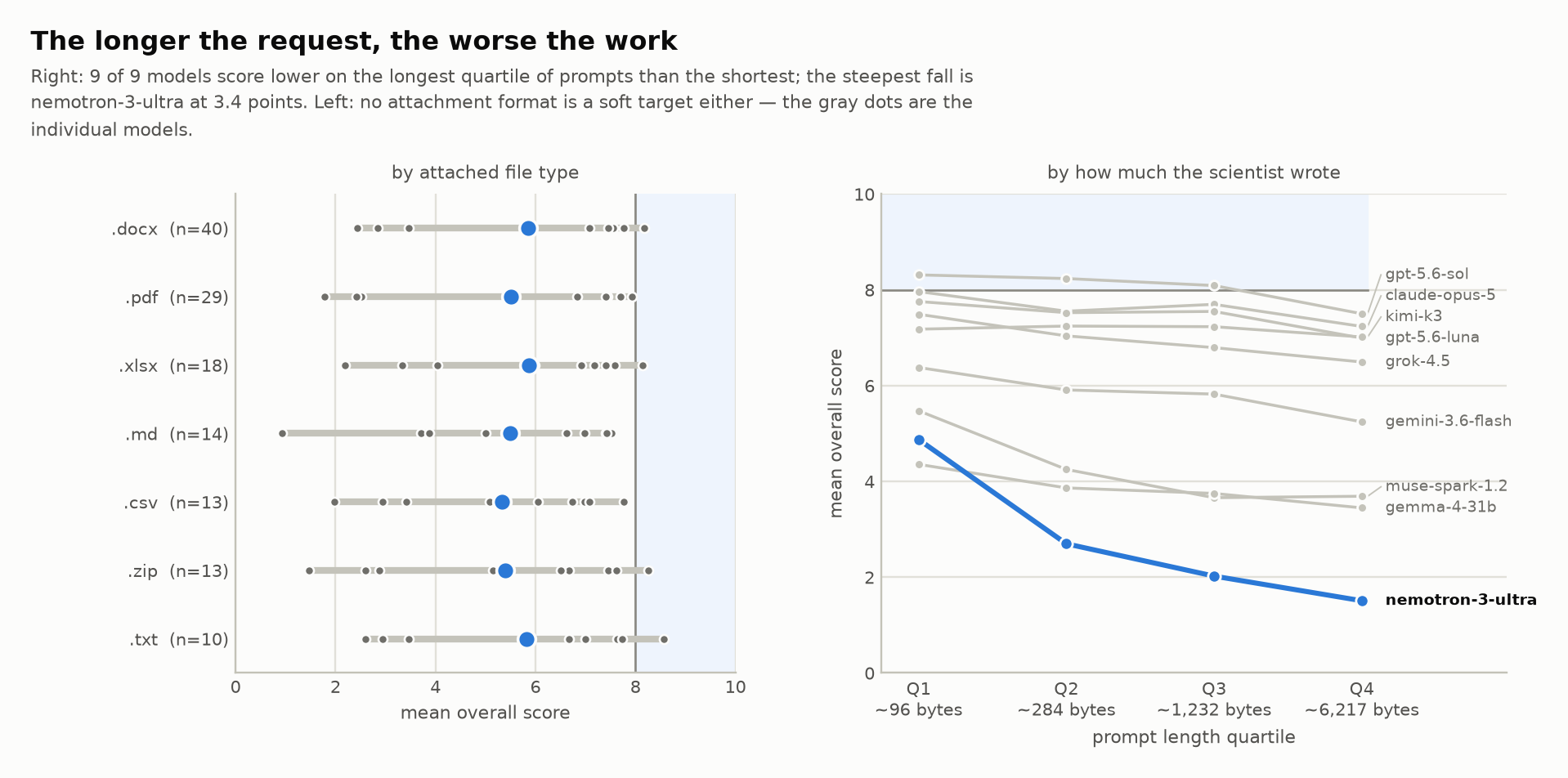}
\caption{\textbf{What conditions difficulty.} Left: mean overall by attached file
type, for file types appearing on at least ten tasks; no format is a soft target and
the model spread within each format is 5.7--6.8 points. Right: mean overall by
prompt-length quartile, one line per model; all nine score lower on Q4 than on Q1,
most steeply for \M{nemotron-3-ultra-550b-a55b}. The decline is not monotone for
every model: \M{claude-opus-5} recovers from Q2 to Q3, \M{kimi-k3} from Q1 to Q2 and
\M{muse-spark-1.2} from Q3 to Q4. File extensions and prompt byte counts
are task-registry attributes (Section~\ref{sec:availability}).}
\label{fig:conditioning}
\end{figure}

\paragraph{Request length.} All nine models score lower on the longest
quartile of prompts than on the shortest (Table~\ref{tab:lenquartile},
Figure~\ref{fig:conditioning}). The drop from Q1 (median 96 bytes) to Q4 (median
6{,}217 bytes) is 0.8 points for \M{gpt-5.6-sol}, 0.7 for \M{claude-opus-5}, 1.0 for
\M{grok-4.5}, 1.8 for \M{muse-spark-1.2} and 3.4 for
\M{nemotron-3-ultra-550b-a55b}. Long, multi-part scientific requests are the hardest
region of this distribution and the natural sub-slice to track separately. The strongest single run-level correlate of quality in the whole campaign
is negative and structural: Spearman $\rho = -0.42$ between input tokens and overall
score. In this distribution, longer prompts are a reliable signal of difficulty.

\subsection{How runs end}
\label{sec:stops}

How a run terminates largely determines its outcome
(Table~\ref{tab:stop}). Of 1{,}602 runs, 1{,}354 ended with a
clean stop, 224 hit a context window limit, 14 ended in error and 10 ended mid-tool-use.
Clean-stopping runs average 6.70 and reach majority success 47.4\% of the time. Every
other ending has a majority-success rate of exactly zero. Truncated transcripts
(258 runs, 16.1\%) average 2.18 against 6.73 for untruncated, again with no majority
successes at all.

Two consequences follow. For evaluation, truncation is not a nuisance to be filtered
out but a first-class failure mode that is unevenly distributed across systems: two
models truncate on more than half their runs and two never truncate at all, so
filtering truncated runs would rescore the weakest systems upward. For
deployment, a length-limited run is a total loss rather than a partial one, which
argues for harness-level checkpointing of intermediate artifacts rather than for
longer limits alone.

Truncation falls almost entirely on two of the three lowest-ranked systems --- 
\M{nemotron-3-ultra-550b-a55b} truncates on 64.6\% of its runs and \M{muse-spark-1.2}
on 52.8\%, against 3.9\% or less for the top four --- which invites the reading that
the bottom of the table is measuring context budgets rather than ability. The advertised windows are given in Table~\ref{tab:modelids} and they do not predict
truncation. \M{gemini-3.6-flash} and \M{muse-spark-1.2} run on identical
1{,}048{,}576-token windows and truncate on 0.0\% and 52.8\% of runs respectively;
\M{gemma-4-31b-it} has the smallest window in the field at 262{,}144 and truncates less
often (7.9\%) than \M{grok-4.5} at 500{,}000 (11.2\%). What separates them is how fast
a run consumes the window, not how large it is: \M{muse-spark-1.2} carries \M{tool\_thrashing} on 30.5\% of its
assessments against 5.1\% across the corpus, at a 26\% tool error rate
(Tables~\ref{tab:failures} and~\ref{tab:runmetrics}).

\subsection{Effort and cost}
\label{sec:cost}

\begin{figure}[t]
\centering
\includegraphics[width=\textwidth]{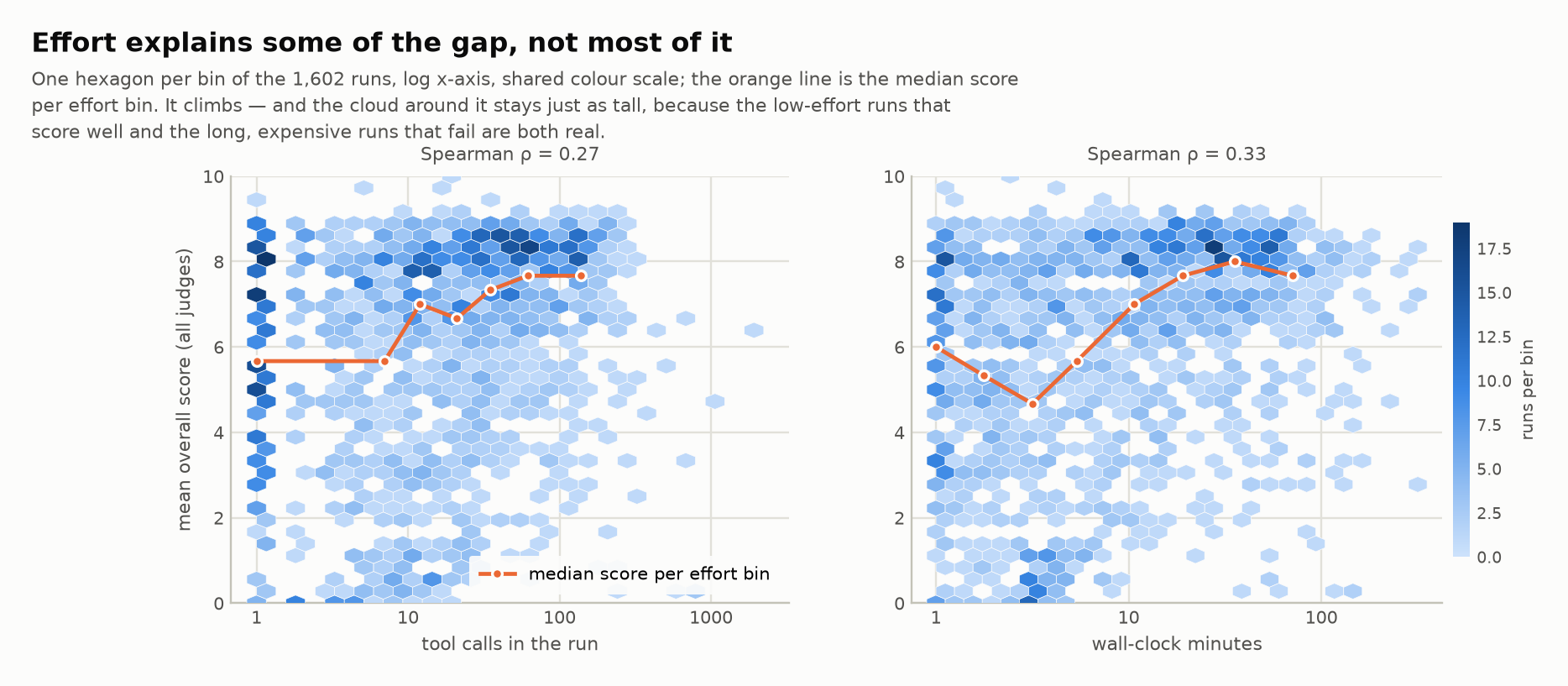}
\caption{\textbf{Effort explains some of the gap, not most of it.} Hexagonal density
of the 1{,}602 runs against tool calls (left) and wall-clock minutes (right), with
the median score per effort bin overlaid. The median trends upward across most of the
range, though not monotonically: it dips in the lowest wall-clock bins before rising
and flattens in the highest. The vertical spread within each bin stays wide
throughout: low-effort runs that score well and long expensive runs that fail are
both common.}
\label{fig:effort}
\end{figure}

Effort correlates with quality across the corpus, moderately and positively: Spearman
$\rho = 0.27$ for tool calls, $0.33$ for wall-clock time, $0.31$ for cost and $0.35$
for thinking characters (Figure~\ref{fig:effort}, Table~\ref{tab:correlates}). Some
of the spread between models therefore reflects how much work a run did, and
comparisons that ignore it are partly measuring budget.

Within a model, however, the sign flips. For seven
of nine models the within-model correlation between turns and score is negative,
from $-0.37$ for \M{gemini-3.6-flash} to $-0.08$ for
\M{nemotron-3-ultra-550b-a55b}, while \M{gpt-5.6-sol} ($+0.01$) and \M{kimi-k3}
($+0.02$) are flat. The between-model and within-model relationships answer different
questions: across models more effort marks a more capable system, but within a fixed
model a long run usually means a stuck one. Treating turn count as a proxy for
diligence is safe only in the first sense.

Tool error rate has a non-monotone relationship with quality that is worth stating.
Runs with no tool errors at all average 5.95, runs with an error rate between 0 and
5\% average 7.47, and runs above 15\% average 4.36. The best outcomes come from runs
that attempted enough to fail occasionally and recovered; zero errors mostly marks a
run that never tried anything demanding.

Cost separates the field by more than two orders of magnitude
(Table~\ref{tab:cost}).
\M{gpt-5.6-sol} costs \$8.51 per task on average, 411 times the \$0.02 of
\M{gemma-4-31b-it}, for $+4.18$ points of mean score, and it still lands with its
interval straddling the acceptable line. Three points sit on the Pareto frontier:
\M{gemma-4-31b-it} at the bottom, \M{gpt-5.6-luna} in the middle, and
\M{gpt-5.6-sol} at the top. \M{gpt-5.6-luna} is the notable case: at \$0.15 per task
it reaches 7.46, 93\% of the leader's mean score for 1.8\% of its cost, and it
achieves 60\% majority success.

\subsection{Task difficulty}
\label{sec:leftover}

Twenty-three tasks were majority-solved by exactly one model, and more often than
not that model is not the leaderboard leader. \M{claude-opus-5} is the unique solver on
14 of the 23, against 7 for \M{gpt-5.6-sol}, one each for \M{grok-4.5} and
\M{kimi-k3}, and none for the remaining five systems.
A model that loses the aggregate comparison is therefore
the only system that gets a specific piece of work done twice as often as the model
that wins it. For a laboratory this is an argument for portfolio behavior rather
than for standardization on the top of a leaderboard.

The within-task spread across models is correspondingly large. The mean standard
deviation of the nine model means within a task is 2.27 points, the mean
best-to-worst range is 6.32 and the median is 7.33, and 113 of 178 tasks have a range
greater than 6. Only two tasks have a range below 1. Model choice, in other words, is
usually the dominant term for any individual scientific request in this
distribution.

\subsection{How failures travel together}
\label{sec:cooccur}

\begin{figure}[t]
\centering
\includegraphics[width=0.75\textwidth]{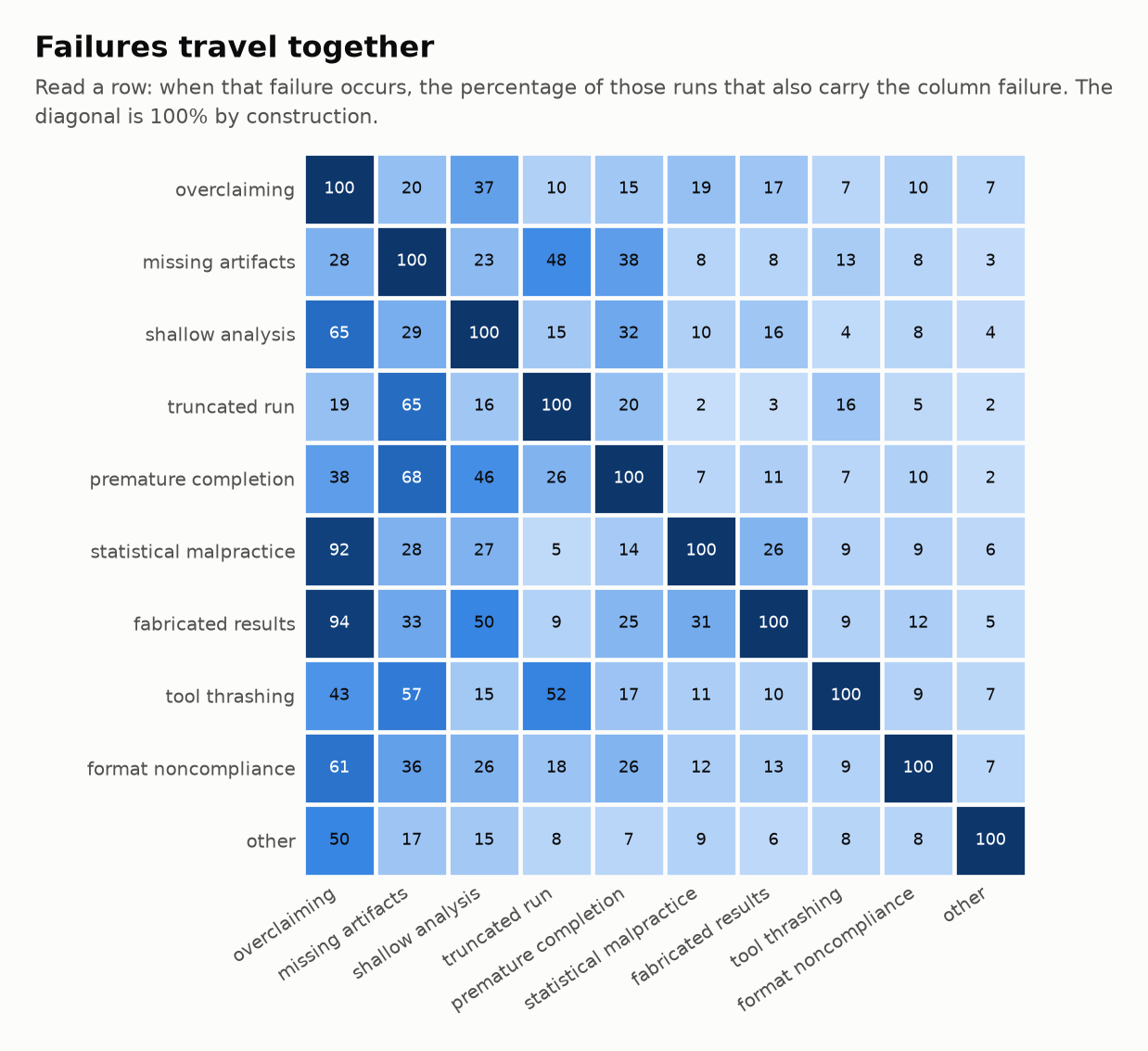}
\caption{\textbf{Failures travel together.} Conditional co-occurrence of the ten most
frequent failure tags: read a row for the share of runs carrying that tag which also
carry the column tag. The diagonal is 100\% by construction. \M{overclaiming} is the
darkest off-diagonal column and exceeds 50\% on five of the nine other rows, led by
\M{fabricated\_results} (94\%) and \M{statistical\_malpractice} (92\%); it is a
minority companion to the truncation-driven failures. Where the \M{truncated\_run}
and \M{missing\_artifacts} rows and columns cross, at 65\% and 48\%, is the
mechanical signature of runs cut off before writing output.}
\label{fig:cooccur}
\end{figure}

Failure tags are not independent, and their conditional structure describes
recognizable syndromes rather than a list of unrelated defects
(Table~\ref{tab:cooccur}, Figure~\ref{fig:cooccur}). Three patterns were observed. When
a run is tagged for \M{statistical\_malpractice}, it also carries \M{overclaiming}
92\% of the time, and when it is tagged for \M{fabricated\_results} it carries
\M{overclaiming} 94\% of the time. When a run is truncated, it is
tagged \M{missing\_artifacts} 65\% of the time, which is the mechanical signature of
being cut off before writing anything out. And when a run is tagged
\M{premature\_completion}, it carries \M{missing\_artifacts} 68\% of the time and
\M{shallow\_analysis} 46\%.

A single underlying behavior, finishing the narrative regardless of
whether the work finished, produces several errors at once.


\section{Judge reliability and alignment}
\label{sec:judges}

This section treats the judging panel as an
object of study: how much of the measurement is reproducible, which part is not, and
what happens when two of the three judges are also contestants.

\subsection{The judges agree on order and disagree on level}

\begin{figure}[t]
\centering
\includegraphics[width=\textwidth]{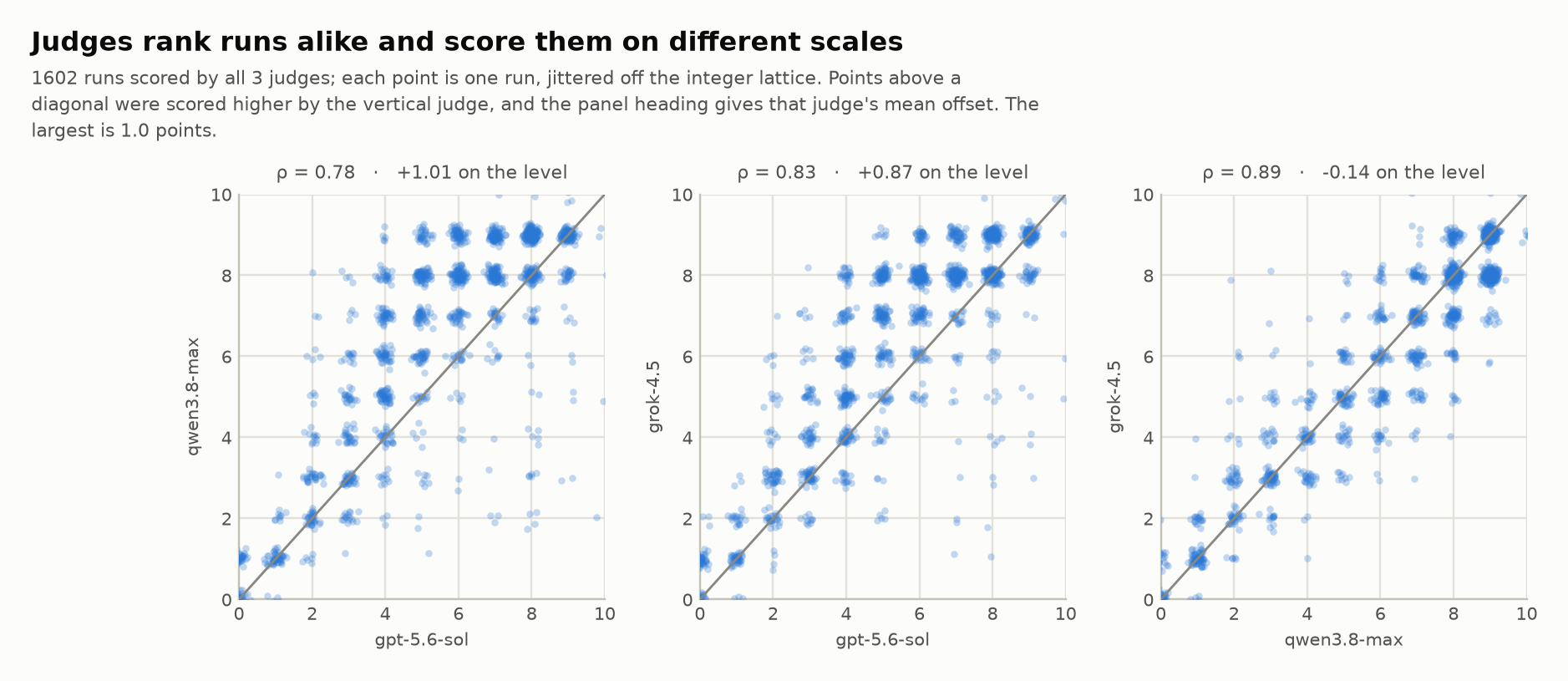}
\caption{\textbf{Judges rank runs alike and score them on different scales.} Each
point is one of the 1{,}602 runs, jittered off the integer lattice; points above a
diagonal were scored higher by the vertical judge. Panel headings give the pairwise
rank correlation and the mean offset in level. The largest offset is 1.0 points.}
\label{fig:judgeagree}
\end{figure}

Across the 1{,}602 runs, mean pairwise Spearman correlation on the holistic overall
score is $\rho = 0.83$, while the judges' own mean overall scores span 5.37
(\M{gpt-5.6-sol}), 6.24 (\M{grok-4.5}) and 6.38 (\M{qwen3.8-max}), a range of 1.0
points (Table~\ref{tab:judgepairs}, Figure~\ref{fig:judgeagree}). Kendall's $W$ over
the three judges' rankings of the nine models is 0.955: the judges essentially agree
on the ordering of systems while disagreeing systematically on the level at which to
anchor the scale.

\M{qwen3.8-max} and \M{grok-4.5}
agree closely with each other ($\rho = 0.89$, mean absolute difference 0.60, 91\%
within a point), while \M{gpt-5.6-sol} sits about a point below both. With two judges
a disagreement is symmetric and unresolvable; with three, the differently calibrated
one is identifiable. \M{gpt-5.6-sol} is the \emph{strict} judge, so it is the
one whose calibration is unusual relative to the panel. That
does not establish that the other two are right, which would take a reference the panel
does not contain (Section~\ref{sec:validity}).

\begin{figure}[t]
\centering
\includegraphics[width=\textwidth]{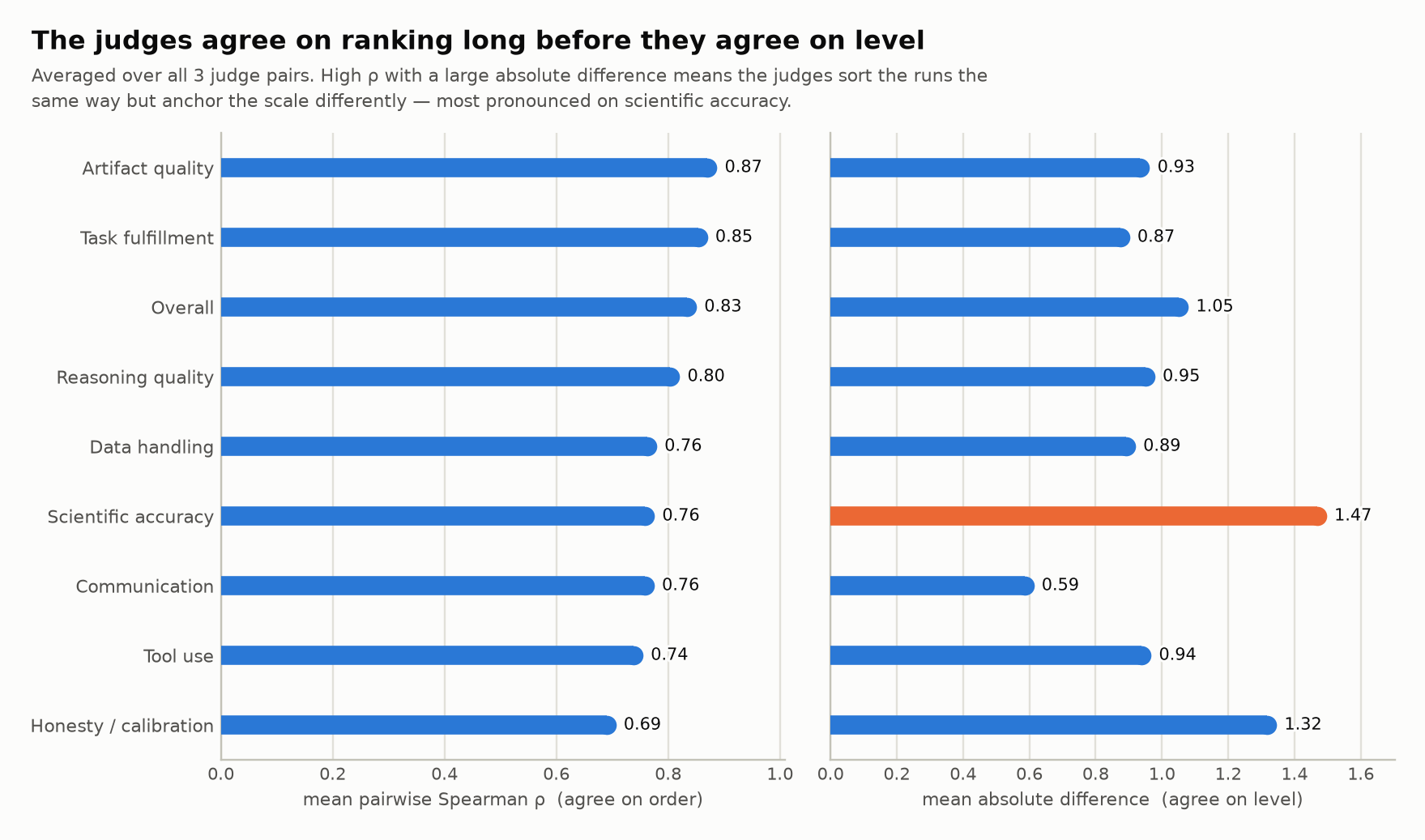}
\caption{\textbf{The judges agree on ranking long before they agree on level.} Mean
pairwise Spearman correlation (left) and mean absolute difference (right) for the
eight rubric dimensions and the holistic overall score. Scientific accuracy pairs
mid-field rank agreement ($\rho = 0.76$, joint fifth of the nine rows) with the
largest level disagreement anywhere in the rubric (1.47): the judges sort runs on it
about as consistently as they sort the rest, and anchor the scale furthest apart.}
\label{fig:dimagree}
\end{figure}

Agreement also varies by dimension in an interpretable way
(Table~\ref{tab:dimagree}, Figure~\ref{fig:dimagree}). Communication is the easiest
thing to agree on in level (mean absolute difference 0.59, 91\% within a point).
Scientific accuracy is the hardest (1.47, 60\%), followed by honesty and calibration
(1.32, 65\%). Those are the two dimensions where a judge must form its own
view of the domain rather than assess a surface property.

\subsection{What the panel establishes, and what it does not}
\label{sec:validity}

Three judges placing
1{,}602 runs in nearly the same order ($W = 0.955$) establishes that the ranking is
reproducible under a change of judge. It establishes nothing about where the scale
sits, because a panel can be reliably wrong about level in the same way it is reliably
right about order; this panel visibly disagrees about level, by 1.0 point on the
pooled mean and 1.47 on scientific accuracy.

How many of the nine models reach the rubric's 8-anchor? \M{qwen3.8-max} says two
(\M{claude-opus-5} 8.29, \M{gpt-5.6-sol} 8.20). \M{grok-4.5} says two
(\M{claude-opus-5} 8.15, \M{gpt-5.6-sol} 8.01). \M{gpt-5.6-sol} says none: its highest
mean for any model, its own included, is 7.90
(Table~\ref{tab:headline}).

The published evidence for model judges does not close this gap. Judges were validated
at scale against human preference between responses: MT-Bench and Chatbot Arena
compare judge verdicts to human pairwise choices, and G-Eval reports rank correlation
with human ratings \citep{mtbench,arena,geval}. Evidence that a judge puts an
absolute threshold where an expert would put it comes from a different design: experts
writing the criteria for each item, as in HealthBench with 262 physicians
\citep{healthbench}, or an expert-written rubric paired with each prompt, as in
ResearchRubrics \citep{researchrubrics}. K-Bench trades that property away by
construction. Because the items are drawn at random from traffic and are not known
before the draw, one rubric has to serve every task.

\subsection{Split decisions}
\label{sec:splits}

The binary success flag makes the structure of disagreement legible. Of 1{,}602 runs,
735 were rejected by all three judges, 307 were accepted by all three, 335 split
two-to-one in favor and 225 split one-to-two, so 35.0\% of runs are split
decisions. The identity of the dissenter is extremely lopsided. On the 335 runs where
two judges accepted and one rejected, the lone rejector is \M{gpt-5.6-sol} 311 times,
\M{grok-4.5} 14 times and \M{qwen3.8-max} 10 times. On the 225 runs where only one
judge accepted, the lone acceptor is \M{qwen3.8-max} 152 times, \M{grok-4.5} 48 times
and \M{gpt-5.6-sol} 25 times.

A majority across these three judges is not an independent tie-break; on split
decisions it reports what the two mutually agreeing judges think, and the strict judge
is outvoted in 93\% of the two-to-one cases. Any headline success rate computed by
majority therefore inherits the calibration of that pair. 

\subsection{Judges scoring their own runs}
\label{sec:selfpref}

Two of the three judges, \M{gpt-5.6-sol} and \M{grok-4.5}, are also benchmarked
models, so each scores its own runs. Blinding removes explicit identity but not
writing style, and self-preference in model judges is a documented effect
\citep{selfpref,justice}. The calibration-adjusted self-preference is the
gap between how a judge scores itself and how it scores everyone else, minus the same
gap as the peer judges see it. Formally, for judge $j$,
\[
\text{SP}_j \;=\; \big(\bar{s}_{j \rightarrow j} - \bar{s}_{j \rightarrow \neg j}\big)
\;-\; \big(\bar{s}_{\neg j \rightarrow j} - \bar{s}_{\neg j \rightarrow \neg j}\big),
\]
where $\bar{s}_{a \rightarrow b}$ is the mean overall score given by judge set $a$ to
model set $b$. The subtraction removes both the judge's overall strictness and the
model's actual quality, leaving the excess.

The result is asymmetric (Table~\ref{tab:selfpref}). \M{grok-4.5} shows a negligible
$+0.11$. \M{gpt-5.6-sol} shows $+0.83$, which is roughly twice the gap
between the first and second models on the leaderboard. It is the strictest judge in
the panel and it is markedly less strict with itself.

Table~\ref{tab:judgedev} gives \M{gpt-5.6-sol}'s deviation from peer consensus for
each model it scored. Across the eight other models the deviations range from $-0.37$
to $-1.82$ points. An effect also appears at the family level, not only at the
identity level.
\M{gpt-5.6-sol} as a judge ranks its same-vendor sibling \M{gpt-5.6-luna} second of
nine, while both other judges rank that model fourth. On the range-matched comparison
above, \M{gpt-5.6-luna} receives $+1.10$. We cannot separate a genuine disagreement about
quality from a family-recognition effect with this design, and we therefore treat the
pooled ranking of \M{gpt-5.6-luna} as the least trustworthy number in the campaign.

Two practical consequences follow. First, the pooled ordering at the top of the table
is the product of a single judge (Table~\ref{tab:headline}). \M{gpt-5.6-sol} is alone in
ranking itself first, and it does so by scoring \M{claude-opus-5} at 6.40 against the
8.15 and 8.29 its peers award, a gap seen nowhere else in the
panel. Second, where the pooled and the neutral-judge orderings disagree, the neutral
one is preferable. A future edition should either exclude contestants from the panel entirely
or add enough neutral judges that the contaminated ones cannot form a majority.

\subsection{Judge confidence}

Judges reported their own confidence and the number of artifacts they inspected. 
Mean confidence is 0.92 for \M{gpt-5.6-sol}, 0.86 for
\M{grok-4.5} and 0.83 for \M{qwen3.8-max}; only 72 of 4{,}806 assessments (1.5\%)
were made with confidence below 0.7. Confidence correlates negatively with the score
awarded ($\rho = -0.34$).

Artifact inspection correlates positively with the score awarded ($\rho = 0.22$), for
the mechanical reason that runs which produce nothing give a judge nothing to open.
The two judges that opened more files (\M{grok-4.5} 4.22, \M{qwen3.8-max} 4.12) are
also the two more lenient ones, and \M{gpt-5.6-sol}, which opened the fewest (3.65),
is the strictest. We cannot tell from this design whether reading more artifacts
causes a more generous assessment or whether a judge that is already inclined to
credit a run reads more of it, and we flag the association without a causal reading.


\section{Discussion}
\label{sec:discussion}

\subsection{What the corpus says about current systems}

The headline number invites a saturation reading and does not support one. The
best system is, on average, borderline acceptable (Table~\ref{tab:headline}).
The distribution is partially solved at the top and
materially unsolved in the tail. The correct object of study going forward is the
failing subset rather than the pooled mean.

The systems
present their work better than they \emph{do} the work, as noted in
Section~\ref{sec:dimgap}: scientific accuracy and communication are scored on every
assessment, and accuracy sits 1.11 points below communication pooled, and below it in every model in
the field. Honesty and calibration is
the second-highest-scoring dimension in the rubric on average (7.30), while
\M{overclaiming} is the most frequently applied tag in the taxonomy \citep{miragebench}.

\subsection{Implications for post-training}

Three specific implications follow from the measurements.

First, the deficit is scientific judgment. The top five already
use a shell in at least three-quarters of runs and write a readable answer. They
still lose points on the science. What separates the models is whether they check
a source, whether the claim matches the file they wrote, and whether they flag
when it does not.

Second, honesty is trainable in a way that this corpus makes visible. The dispersion
across models is very large, from 6.4\% to 68.2\% of assessments tagged for
overclaiming, and it does not track overall capability: \M{gpt-5.6-luna} carries the
tag on 9.7\% of assessments, while \M{claude-opus-5}, only 0.16 points above it on the
pooled mean, carries it on 32.6\%. The two lowest rates in the field, 6.4\% and
9.7\%, belong to the same model family, while systems that score close to them on
overall quality sit above 27\%. We cannot attribute that difference to any particular
training choice from this evidence, but it is clearly separable from aggregate
quality, and the signal needed to reward it is cheap to compute.

Third, verification behavior is worth optimizing directly. Both verification actions and evidence-tool
use are deterministic to compute and immune to judge calibration
(Section~\ref{sec:transcripts}).

\subsection{Implications for deployment}

A leaderboard position is the wrong selection criterion when the
top systems differ in kind. One profile is the better engineer; the other is the
more careful scientist (Sections~\ref{sec:paired} and~\ref{sec:leftover}). Those
profiles suit different work.

The price of quality is steep and non-linear
(Section~\ref{sec:cost}). For high-volume screening work, the cheap side of the
cliff is defensible; for work that will be published, the extra points are
concentrated exactly in the dimensions of accuracy and honesty, which a reader
would notice.

A strong score-based
ranking is no guarantee of a low empty rate: \M{gpt-5.6-sol} leads every score-based
measure in the campaign and still leaves 37.1\% of its runs empty. Harness-level enforcement that requires declared
deliverables to exist before a run is marked complete would address a failure
mode that no amount of model improvement in this campaign eliminated.

\subsection{Implications for benchmark design}

Three lessons follow for benchmark design. The first is that judges
must open the files; Section~\ref{sec:transcripts} shows why. The second is that
a short, self-contained prompt set would hide the gaps this distribution exposes
(Section~\ref{sec:hard}). The third is that a judging panel drawn from the
systems under test needs explicit handling (Section~\ref{sec:judges}): both the
majority vote and the pooled mean carry the panel's composition inside them.

\subsection{Relation to the published landscape}

K-Bench answers a different question from the suites reviewed in
Section~\ref{sec:related}.
RE-Bench is the instructive
exception: its agents outscore human experts at a two-hour budget and fall behind them
at eight, because humans have better returns to time \citep{rebench}. Our one-shot
design samples the short-horizon end of that curve, so the headroom we report should
not be read as a claim about what these models reach given more turns.

\section{Limitations}
\label{sec:limits}

\paragraph{No human calibration, and no expert baseline.} Two distinct things are
missing here. We never checked the instrument: no
human scored any run, so we do not know where a domain scientist would place the
rubric's 8-anchor relative to where the panel places it, and the panel's own 1.0-point
spread shows that the placement is not pinned down even among the three judges we
used (Section~\ref{sec:validity}). We also do not know what a domain scientist would score on these tasks, so the
absolute distance from human performance is unknown.

\paragraph{Judge scores are panel-dependent.} Two of three judges are also
contestants. \M{gpt-5.6-sol} rates its own runs $+0.83$ after calibration adjustment
and ranks its same-vendor sibling two places higher than the other judges do.

\paragraph{One harness, one-shot prompt, one attempt.} Every run used stock
\M{pi} 0.84.0 with no model-specific prompting, no scientific sub-agents and no
retries. The model received only the initial user prompt and its attached files;
there were no follow-up user turns.

\paragraph{User content was transmitted without content-level de-identification.} The
tasks were replayed to nine external providers as submitted, with attachments. Zero-retention
endpoints were used, but no redaction pass was applied and no audit for identifiers was
carried out (Section~\ref{sec:provenance}).

\paragraph{Private items limit external reproducibility.} The score tables and the
analysis code will be released, so every analysis in this paper can be checked
independently, but the items will not be, so no group outside K-Dense can run a new
model against them; see Section~\ref{sec:availability}.

\paragraph{Traffic is not a random sample of science.} The 178 tasks are a uniform
random draw from K-Dense Web traffic in two batches during one week, so they span four
broad domains with unequal representation (chemistry and materials contributes 17
tasks) as an outcome of the draw rather than by design. They are also a complete-case
set: the 22 sessions at least one model refused were dropped so that every model is
scored on identical tasks, which means the benchmark is silent on
requests that sit near a vendor's safety boundary.

\paragraph{Failure tags are judge-assigned.} The failure taxonomy is applied by the
same judges that assign the scores, so tag rates inherit judge calibration, and tags
are not independent of one another (Section~\ref{sec:cooccur}). Assessment-level and
run-level tag rates differ by up to a factor of two depending on how many judges must
agree, which is why we report several thresholds.

\section*{Evidence and limits}
\addcontentsline{toc}{section}{Evidence and limits}

The 1{,}602
runs were each executed once, and each was scored once by each of the three judges.
Where we give a confidence interval it is a bootstrap over the 178 sampled sessions
and describes sampling uncertainty in the task set alone; it does not cover
run-to-run variance, judge-to-judge variance, or any sensitivity to how the panel was
composed. Point estimates should be read as what these models did on these tasks in
this harness on these dates, not as expectations over repeated attempts.

\section*{Data provenance, consent and privacy}
\addcontentsline{toc}{section}{Data provenance, consent and privacy}
\label{sec:provenance}

\emph{Basis for use.} The sessions were drawn from K-Dense Web logs under the
platform's terms of service, which permit analysis of submitted content for product
research and evaluation. Users were not separately notified of this campaign and did
not individually opt in to it. No session was solicited for the benchmark: all sessions were
submitted in the ordinary course of using the product, before the draw was made.

\emph{What was transmitted.} Each task was replayed to nine third-party model providers
as the user wrote it, with the attached files intact. The blinding described in
Section~\ref{sec:methods} removes \emph{model} identity from judge-visible surfaces; it
is not a de-identification of user content, and no content-level redaction pass was
applied to prompts or attachments before the runs. Requests were executed against
vendor endpoints configured for zero retention, so submitted content is excluded from
provider retention and from provider training.

\section*{Data, code and availability}
\addcontentsline{toc}{section}{Data, code and availability}
\label{sec:availability}

The task prompts, attachments, transcripts and output artifacts are not released.
They are user content.

A de-identified form of the scores will be published that
produces every table and figure in this paper. That will be enough to reproduce the
analyses and check our arithmetic. Rubric v1.0 is reproduced in full in Appendix~\ref{sec:rubric}, including every
dimension anchor and the 16-tag failure taxonomy.

\section*{Author contributions}
\addcontentsline{toc}{section}{Author contributions}

A.B.\ wrote the manuscript, led the benchmarking effort and set its direction,
contributed to rubric development, and ran the internal benchmark campaigns. D.P.\ built the K-Dense Web infrastructure and the session management from
which the task corpus is drawn. Y.H.\ led AI engineering for K-Dense Web and designed
the session metadata used to sample and characterize the corpus. T.K.\ contributed to
benchmark and rubric design, to the run and judging engineering, and to the statistical
analysis and figures, revised the manuscript, and supervised the project.

\section*{Competing interests}
\addcontentsline{toc}{section}{Competing interests}

All authors are employed by K-Dense, Inc. The evaluation corpus is traffic from
K-Dense Web, a K-Dense, Inc. product, and the benchmark design, the harness configuration and
the scoring rubric are all K-Dense's. The nine evaluated systems are third-party models
developed by other organizations, in which K-Dense had no role. Readers should weigh
the results with the authors' position in mind.

\section*{Acknowledgements}
\addcontentsline{toc}{section}{Acknowledgements}

We thank the K-Dense users whose requests constitute this evaluation corpus.

\clearpage
\bibliographystyle{plainnat}
\bibliography{references}

\begin{thebibliography}{73}
\providecommand{\natexlab}[1]{#1}
\providecommand{\url}[1]{\texttt{#1}}
\expandafter\ifx\csname urlstyle\endcsname\relax
  \providecommand{\doi}[1]{doi: #1}\else
  \providecommand{\doi}{doi: \begingroup \urlstyle{rm}\Url}\fi

\bibitem[Agarwal et~al.(2025)Agarwal, Li, Petty, Kassis, Rothemund, Sinclair,
  and Gopinath]{kdenseclock}
Vinayak Agarwal, Orion Li, Christopher~A. Petty, Timothy Kassis, Paul W.~K.
  Rothemund, David~A. Sinclair, and Ashwin Gopinath.
\newblock {Guided multi-agent AI invents highly accurate, uncertainty-aware
  transcriptomic aging clocks}.
\newblock \emph{bioRxiv}, 2025.
\newblock \doi{10.1101/2025.09.08.674588}.
\newblock URL
  \url{https://www.biorxiv.org/content/10.1101/2025.09.08.674588v1}.

\bibitem[Arora et~al.(2025)Arora, Wei, Soskin~Hicks, et~al.]{healthbench}
Rahul~K. Arora, Jason Wei, Rebecca Soskin~Hicks, et~al.
\newblock {HealthBench: Evaluating Large Language Models Towards Improved Human
  Health}.
\newblock \emph{arXiv preprint arXiv:2505.08775}, 2025.
\newblock URL \url{https://arxiv.org/abs/2505.08775}.

\bibitem[Bean et~al.(2025)Bean, Kearns, Romanou, et~al.]{constructvalidity}
Andrew~M. Bean, Ryan~Othniel Kearns, Angelika Romanou, et~al.
\newblock {Measuring What Matters: Construct Validity in Large Language Model
  Benchmarks}.
\newblock \emph{arXiv preprint arXiv:2511.04703}, 2025.
\newblock URL \url{https://arxiv.org/abs/2511.04703}.

\bibitem[Boiko et~al.(2023)Boiko, MacKnight, Kline, and Gomes]{coscientist}
Daniil~A. Boiko, Robert MacKnight, Ben Kline, and Gabe Gomes.
\newblock {Autonomous chemical research with large language models}.
\newblock \emph{Nature}, 624\penalty0 (7992):\penalty0 570--578, 2023.
\newblock \doi{10.1038/s41586-023-06792-0}.
\newblock URL \url{https://doi.org/10.1038/s41586-023-06792-0}.
\newblock Introduces Coscientist; preprint arXiv:2304.05332 (2023).

\bibitem[Bragg et~al.(2025)Bragg, D'Arcy, Balepur, et~al.]{astabench}
Jonathan Bragg, Mike D'Arcy, Nishant Balepur, et~al.
\newblock {AstaBench: Rigorous Benchmarking of AI Agents with a Scientific
  Research Suite}.
\newblock \emph{arXiv preprint arXiv:2510.21652}, 2025.
\newblock URL \url{https://arxiv.org/abs/2510.21652}.
\newblock Published as a conference paper at ICLR 2026.

\bibitem[Chan et~al.(2024)Chan, Chowdhury, Jaffe, et~al.]{mlebench}
Jun~Shern Chan, Neil Chowdhury, Oliver Jaffe, et~al.
\newblock {MLE-bench: Evaluating Machine Learning Agents on Machine Learning
  Engineering}.
\newblock \emph{arXiv preprint arXiv:2410.07095}, 2024.
\newblock URL \url{https://arxiv.org/abs/2410.07095}.

\bibitem[Chen et~al.(2024)Chen, Chen, Ning, et~al.]{scienceagentbench}
Ziru Chen, Shijie Chen, Yuting Ning, et~al.
\newblock {ScienceAgentBench: Toward Rigorous Assessment of Language Agents for
  Data-Driven Scientific Discovery}.
\newblock \emph{arXiv preprint arXiv:2410.05080}, 2024.
\newblock URL \url{https://arxiv.org/abs/2410.05080}.

\bibitem[Chiang et~al.(2024)Chiang, Zheng, Sheng, et~al.]{arena}
Wei-Lin Chiang, Lianmin Zheng, Ying Sheng, et~al.
\newblock {Chatbot Arena: An Open Platform for Evaluating LLMs by Human
  Preference}.
\newblock \emph{arXiv preprint arXiv:2403.04132}, 2024.
\newblock URL \url{https://arxiv.org/abs/2403.04132}.

\bibitem[Dehghani et~al.(2021)Dehghani, Tay, Gritsenko,
  et~al.]{benchmarklottery}
Mostafa Dehghani, Yi~Tay, Alexey~A. Gritsenko, et~al.
\newblock {The Benchmark Lottery}.
\newblock \emph{arXiv preprint arXiv:2107.07002}, 2021.
\newblock URL \url{https://arxiv.org/abs/2107.07002}.

\bibitem[Du et~al.(2025)Du, Xu, Zhu, Wang, and Mao]{deepresearchbench}
Mingxuan Du, Benfeng Xu, Chiwei Zhu, Xiaorui Wang, and Zhendong Mao.
\newblock {DeepResearch Bench: A Comprehensive Benchmark for Deep Research
  Agents}.
\newblock \emph{arXiv preprint arXiv:2506.11763}, 2025.
\newblock URL \url{https://arxiv.org/abs/2506.11763}.

\bibitem[Duan et~al.(2025)Duan, Lu, Harrigan, et~al.]{scigym}
Haonan Duan, Stephen~Zhewen Lu, Caitlin~Fiona Harrigan, et~al.
\newblock {Measuring Scientific Capabilities of Language Models with a Systems
  Biology Dry Lab}.
\newblock \emph{arXiv preprint arXiv:2507.02083}, 2025.
\newblock URL \url{https://arxiv.org/abs/2507.02083}.

\bibitem[{Earendil Works}(2026)]{piharness}
{Earendil Works}.
\newblock {Pi Agent Harness}.
\newblock \url{https://github.com/earendil-works/pi}, 2026.
\newblock Version 0.84.0; accessed 18 August 2026.

\bibitem[Golchin and Surdeanu(2023)]{contamination}
Shahriar Golchin and Mihai Surdeanu.
\newblock {Data Contamination Quiz: A Tool to Detect and Estimate Contamination
  in Large Language Models}.
\newblock \emph{arXiv preprint arXiv:2311.06233}, 2023.
\newblock URL \url{https://arxiv.org/abs/2311.06233}.

\bibitem[Gu et~al.(2024{\natexlab{a}})Gu, Jiang, Shi, et~al.]{judgesurvey}
Jiawei Gu, Xuhui Jiang, Zhichao Shi, et~al.
\newblock {A Survey on LLM-as-a-Judge}.
\newblock \emph{arXiv preprint arXiv:2411.15594}, 2024{\natexlab{a}}.
\newblock URL \url{https://arxiv.org/abs/2411.15594}.

\bibitem[Gu et~al.(2024{\natexlab{b}})Gu, Shang, Jiang, et~al.]{blade}
Ken Gu, Ruoxi Shang, Ruien Jiang, et~al.
\newblock {BLADE: Benchmarking Language Model Agents for Data-Driven Science}.
\newblock \emph{arXiv preprint arXiv:2408.09667}, 2024{\natexlab{b}}.
\newblock URL \url{https://arxiv.org/abs/2408.09667}.

\bibitem[Hendrycks et~al.(2020)Hendrycks, Burns, Basart, et~al.]{mmlu}
Dan Hendrycks, Collin Burns, Steven Basart, et~al.
\newblock {Measuring Massive Multitask Language Understanding}.
\newblock \emph{arXiv preprint arXiv:2009.03300}, 2020.
\newblock URL \url{https://arxiv.org/abs/2009.03300}.

\bibitem[Hu et~al.(2024{\natexlab{a}})Hu, Zhao, Wei, et~al.]{infiagent}
Xueyu Hu, Ziyu Zhao, Shuang Wei, et~al.
\newblock {InfiAgent-DABench: Evaluating Agents on Data Analysis Tasks}.
\newblock \emph{arXiv preprint arXiv:2401.05507}, 2024{\natexlab{a}}.
\newblock URL \url{https://arxiv.org/abs/2401.05507}.

\bibitem[Hu et~al.(2024{\natexlab{b}})Hu, Song, Zhang, et~al.]{verbositybias}
Zhengyu Hu, Linxin Song, Jieyu Zhang, et~al.
\newblock {Explaining Length Bias in LLM-Based Preference Evaluations}.
\newblock \emph{arXiv preprint arXiv:2407.01085}, 2024{\natexlab{b}}.
\newblock URL \url{https://arxiv.org/abs/2407.01085}.

\bibitem[Ivanov(2024)]{biolpbench}
Igor Ivanov.
\newblock {BioLP-bench: Measuring understanding of biological lab protocols by
  large language models}.
\newblock \emph{bioRxiv}, 2024.
\newblock \doi{10.1101/2024.08.21.608694}.
\newblock URL \url{https://www.biorxiv.org/content/10.1101/2024.08.21.608694}.

\bibitem[Jimenez et~al.(2023)Jimenez, Yang, Wettig, et~al.]{swebench}
Carlos~E. Jimenez, John Yang, Alexander Wettig, et~al.
\newblock {SWE-bench: Can Language Models Resolve Real-World GitHub Issues?}
\newblock \emph{arXiv preprint arXiv:2310.06770}, 2023.
\newblock URL \url{https://arxiv.org/abs/2310.06770}.

\bibitem[Jing et~al.(2024)Jing, Huang, Wang, et~al.]{dsbench}
Liqiang Jing, Zhehui Huang, Xiaoyang Wang, et~al.
\newblock {DSBench: How Far Are Data Science Agents from Becoming Data Science
  Experts?}
\newblock \emph{arXiv preprint arXiv:2409.07703}, 2024.
\newblock URL \url{https://arxiv.org/abs/2409.07703}.

\bibitem[{K-Dense Inc.}(2026)]{kdenseweb}
{K-Dense Inc.}
\newblock {K-Dense Web}.
\newblock \url{https://www.k-dense.ai}, 2026.
\newblock Accessed 18 August 2026.

\bibitem[Kapoor et~al.(2025)Kapoor, Stroebl, Kirgis, et~al.]{hal}
Sayash Kapoor, Benedikt Stroebl, Peter Kirgis, et~al.
\newblock {Holistic Agent Leaderboard: The Missing Infrastructure for AI Agent
  Evaluation}.
\newblock \emph{arXiv preprint arXiv:2510.11977}, 2025.
\newblock URL \url{https://arxiv.org/abs/2510.11977}.

\bibitem[Laurent et~al.(2024)Laurent, Janizek, Ruzo, et~al.]{labbench}
Jon~M. Laurent, Joseph~D. Janizek, Michael Ruzo, et~al.
\newblock {LAB-Bench: Measuring Capabilities of Language Models for Biology
  Research}.
\newblock \emph{arXiv preprint arXiv:2407.10362}, 2024.
\newblock URL \url{https://arxiv.org/abs/2407.10362}.

\bibitem[Laurent et~al.(2026)Laurent, Bou, Pieler, et~al.]{labbench2}
Jon~M Laurent, Albert Bou, Michael Pieler, et~al.
\newblock {LABBench2: An Improved Benchmark for AI Systems Performing Biology
  Research}.
\newblock \emph{arXiv preprint arXiv:2604.09554}, 2026.
\newblock URL \url{https://arxiv.org/abs/2604.09554}.

\bibitem[Li and Ho(2026)]{genebenchpro}
Jeremy Li and Andrew Ho.
\newblock {GeneBench-Pro: Evaluating Multistage Statistical Reasoning in
  Genomics, Quantitative Biology, and Translational Biomedicine}.
\newblock \emph{bioRxiv}, 2026.
\newblock \doi{10.64898/2026.06.29.735386}.
\newblock URL
  \url{https://www.biorxiv.org/content/10.64898/2026.06.29.735386v2}.
\newblock Announced at
  \url{https://openai.com/index/introducing-genebench-pro/}.

\bibitem[Li et~al.(2025)Li, Agarwal, Zhou, Gopinath, and Kassis]{kdenseanalyst}
Orion Li, Vinayak Agarwal, Summer Zhou, Ashwin Gopinath, and Timothy Kassis.
\newblock {K-Dense Analyst: Towards Fully Automated Scientific Analysis}.
\newblock \emph{arXiv preprint arXiv:2508.07043}, 2025.
\newblock URL \url{https://arxiv.org/abs/2508.07043}.

\bibitem[Li et~al.(2024)Li, Chiang, Frick, et~al.]{arenahard}
Tianle Li, Wei-Lin Chiang, Evan Frick, et~al.
\newblock {From Crowdsourced Data to High-Quality Benchmarks: Arena-Hard and
  BenchBuilder Pipeline}.
\newblock \emph{arXiv preprint arXiv:2406.11939}, 2024.
\newblock URL \url{https://arxiv.org/abs/2406.11939}.

\bibitem[Liang et~al.(2022)Liang, Bommasani, Lee, et~al.]{helm}
Percy Liang, Rishi Bommasani, Tony Lee, et~al.
\newblock {Holistic Evaluation of Language Models}.
\newblock \emph{arXiv preprint arXiv:2211.09110}, 2022.
\newblock URL \url{https://arxiv.org/abs/2211.09110}.

\bibitem[Lin et~al.(2024)Lin, Deng, Chandu, et~al.]{wildbench}
Bill~Yuchen Lin, Yuntian Deng, Khyathi Chandu, et~al.
\newblock {WildBench: Benchmarking LLMs with Challenging Tasks from Real Users
  in the Wild}.
\newblock \emph{arXiv preprint arXiv:2406.04770}, 2024.
\newblock URL \url{https://arxiv.org/abs/2406.04770}.

\bibitem[Liu et~al.(2026{\natexlab{a}})Liu, Ho, Droste, et~al.]{lifescibench}
Amelia Liu, Andrew Ho, Anne~Marie Droste, et~al.
\newblock {LifeSciBench: Evaluating Language Models on Realistic, Expert-Level
  Tasks in the Life Sciences}.
\newblock Technical report, OpenAI and Tacit Labs, jun 2026{\natexlab{a}}.
\newblock URL \url{https://openai.com/index/introducing-life-sci-bench/}.

\bibitem[Liu et~al.(2026{\natexlab{b}})Liu, Gu, Cheng, Sun, You, and
  Hu]{physdox}
He~Liu, Boyuan Gu, Shuaiqi Cheng, Haiyang Sun, Siyu You, and Xuming Hu.
\newblock {PhysDox: Benchmarking LLMs on Physical Feasibility Auditing of
  Physiological Sensing Protocols}.
\newblock \emph{arXiv preprint arXiv:2606.05003}, 2026{\natexlab{b}}.
\newblock URL \url{https://arxiv.org/abs/2606.05003}.

\bibitem[Liu et~al.(2023{\natexlab{a}})Liu, Yu, Zhang, et~al.]{agentbench}
Xiao Liu, Hao Yu, Hanchen Zhang, et~al.
\newblock {AgentBench: Evaluating LLMs as Agents}.
\newblock \emph{arXiv preprint arXiv:2308.03688}, 2023{\natexlab{a}}.
\newblock URL \url{https://arxiv.org/abs/2308.03688}.

\bibitem[Liu et~al.(2023{\natexlab{b}})Liu, Iter, Xu, Wang, Xu, and Zhu]{geval}
Yang Liu, Dan Iter, Yichong Xu, Shuohang Wang, Ruochen Xu, and Chenguang Zhu.
\newblock {G-Eval: NLG Evaluation using GPT-4 with Better Human Alignment}.
\newblock \emph{arXiv preprint arXiv:2303.16634}, 2023{\natexlab{b}}.
\newblock URL \url{https://arxiv.org/abs/2303.16634}.

\bibitem[Liu et~al.(2025)Liu, Lv, Zhang, Wang, Yuan, and Tian]{bioprobench}
Yuyang Liu, Liuzhenghao Lv, Xiancheng Zhang, Jingya Wang, Li~Yuan, and Yonghong
  Tian.
\newblock {BioProBench: A Corpus and Benchmark for Biological Protocol
  Reasoning in Autonomous Science}.
\newblock \emph{arXiv preprint arXiv:2505.07889}, 2025.
\newblock URL \url{https://arxiv.org/abs/2505.07889}.

\bibitem[Lu et~al.(2024)Lu, Lu, Lange, Foerster, Clune, and Ha]{aiscientist}
Chris Lu, Cong Lu, Robert~Tjarko Lange, Jakob Foerster, Jeff Clune, and David
  Ha.
\newblock {The AI Scientist: Towards Fully Automated Open-Ended Scientific
  Discovery}.
\newblock \emph{arXiv preprint arXiv:2408.06292}, 2024.
\newblock URL \url{https://arxiv.org/abs/2408.06292}.

\bibitem[Lupidi et~al.(2026)Lupidi, Gauri, Foster, et~al.]{airsbench}
Alisia Lupidi, Bhavul Gauri, Thomas~Simon Foster, et~al.
\newblock {AIRS-Bench: a Suite of Tasks for Frontier AI Research Science
  Agents}.
\newblock \emph{arXiv preprint arXiv:2602.06855}, 2026.
\newblock URL \url{https://arxiv.org/abs/2602.06855}.

\bibitem[Lv et~al.(2026)Lv, Tan, Li, et~al.]{realclawbench}
Zongwei Lv, Zhewen Tan, Yaoming Li, et~al.
\newblock {RealClawBench: Live OpenClaw Benchmarks from Real Developer-Agent
  Sessions}.
\newblock \emph{arXiv preprint arXiv:2606.03889}, 2026.
\newblock URL \url{https://arxiv.org/abs/2606.03889}.

\bibitem[M.~Bran et~al.(2024)M.~Bran, Cox, Schilter, Baldassari, White, and
  Schwaller]{chemcrow}
Andres M.~Bran, Sam Cox, Oliver Schilter, Carlo Baldassari, Andrew~D. White,
  and Philippe Schwaller.
\newblock {Augmenting large language models with chemistry tools}.
\newblock \emph{Nature Machine Intelligence}, 6\penalty0 (5):\penalty0
  525--535, 2024.
\newblock \doi{10.1038/s42256-024-00832-8}.
\newblock URL \url{https://doi.org/10.1038/s42256-024-00832-8}.
\newblock Introduces ChemCrow; preprint arXiv:2304.05376 (2023).

\bibitem[Majumder et~al.(2024)Majumder, Surana, Agarwal,
  et~al.]{discoverybench}
Bodhisattwa~Prasad Majumder, Harshit Surana, Dhruv Agarwal, et~al.
\newblock {DiscoveryBench: Towards Data-Driven Discovery with Large Language
  Models}.
\newblock \emph{arXiv preprint arXiv:2407.01725}, 2024.
\newblock URL \url{https://arxiv.org/abs/2407.01725}.

\bibitem[Merrill et~al.(2026)Merrill, Shaw, Carlini, et~al.]{terminalbench}
Mike~A. Merrill, Alexander~G. Shaw, Nicholas Carlini, et~al.
\newblock {Terminal-Bench: Benchmarking Agents on Hard, Realistic Tasks in
  Command Line Interfaces}.
\newblock \emph{arXiv preprint arXiv:2601.11868}, 2026.
\newblock URL \url{https://arxiv.org/abs/2601.11868}.

\bibitem[Mialon et~al.(2023)Mialon, Fourrier, Swift, Wolf, LeCun, and
  Scialom]{gaia}
Grégoire Mialon, Clémentine Fourrier, Craig Swift, Thomas Wolf, Yann LeCun,
  and Thomas Scialom.
\newblock {GAIA: a benchmark for General AI Assistants}.
\newblock \emph{arXiv preprint arXiv:2311.12983}, 2023.
\newblock URL \url{https://arxiv.org/abs/2311.12983}.

\bibitem[Mitchener et~al.(2025)Mitchener, Laurent, Andonian, et~al.]{bixbench}
Ludovico Mitchener, Jon~M Laurent, Alex Andonian, et~al.
\newblock {BixBench: a Comprehensive Benchmark for LLM-based Agents in
  Computational Biology}.
\newblock \emph{arXiv preprint arXiv:2503.00096}, 2025.
\newblock URL \url{https://arxiv.org/abs/2503.00096}.

\bibitem[Nathani et~al.(2025)Nathani, Madaan, Roberts, et~al.]{mlgym}
Deepak Nathani, Lovish Madaan, Nicholas Roberts, et~al.
\newblock {MLGym: A New Framework and Benchmark for Advancing AI Research
  Agents}.
\newblock \emph{arXiv preprint arXiv:2502.14499}, 2025.
\newblock URL \url{https://arxiv.org/abs/2502.14499}.

\bibitem[O'Donoghue et~al.(2023)O'Donoghue, Shtedritski, Ginger,
  et~al.]{bioplanner}
Odhran O'Donoghue, Aleksandar Shtedritski, John Ginger, et~al.
\newblock {BioPlanner: Automatic Evaluation of LLMs on Protocol Planning in
  Biology}.
\newblock \emph{arXiv preprint arXiv:2310.10632}, 2023.
\newblock URL \url{https://arxiv.org/abs/2310.10632}.

\bibitem[Panickssery et~al.(2024)Panickssery, Bowman, and Feng]{selfpref}
Arjun Panickssery, Samuel~R. Bowman, and Shi Feng.
\newblock {LLM Evaluators Recognize and Favor Their Own Generations}.
\newblock \emph{arXiv preprint arXiv:2404.13076}, 2024.
\newblock URL \url{https://arxiv.org/abs/2404.13076}.

\bibitem[Phan et~al.(2026)Phan, Gatti, Li, et~al.]{hle}
Long Phan, Alice Gatti, Nathaniel Li, et~al.
\newblock {A benchmark of expert-level academic questions to assess AI
  capabilities}.
\newblock \emph{Nature}, 649\penalty0 (8099):\penalty0 1139--1146, 2026.
\newblock \doi{10.1038/s41586-025-09962-4}.
\newblock URL \url{https://doi.org/10.1038/s41586-025-09962-4}.
\newblock Introduces Humanity's Last Exam (HLE); preprint arXiv:2501.14249
  (2025).

\bibitem[Raji et~al.(2021)Raji, Bender, Paullada, Denton, and
  Hanna]{everythingbenchmark}
Inioluwa~Deborah Raji, Emily~M. Bender, Amandalynne Paullada, Emily Denton, and
  Alex Hanna.
\newblock {AI and the Everything in the Whole Wide World Benchmark}.
\newblock \emph{arXiv preprint arXiv:2111.15366}, 2021.
\newblock URL \url{https://arxiv.org/abs/2111.15366}.

\bibitem[Rein et~al.(2023)Rein, Hou, Stickland, et~al.]{gpqa}
David Rein, Betty~Li Hou, Asa~Cooper Stickland, et~al.
\newblock {GPQA: A Graduate-Level Google-Proof Q\&A Benchmark}.
\newblock \emph{arXiv preprint arXiv:2311.12022}, 2023.
\newblock URL \url{https://arxiv.org/abs/2311.12022}.

\bibitem[Sharma et~al.(2025)Sharma, Zhang, Bandi, et~al.]{researchrubrics}
Manasi Sharma, Chen Bo~Calvin Zhang, Chaithanya Bandi, et~al.
\newblock {ResearchRubrics: A Benchmark of Prompts and Rubrics for Evaluating
  Deep Research Agents}.
\newblock \emph{arXiv preprint arXiv:2511.07685}, 2025.
\newblock URL \url{https://arxiv.org/abs/2511.07685}.

\bibitem[Shi et~al.(2026)Shi, Ren, Song, Li, Sheng, and Sun]{sdabench}
Chuhan Shi, Xiaoquan Ren, Sicheng Song, Haobo Li, Rui Sheng, and Yushi Sun.
\newblock {Are LLMs Ready for Scientific Discovery? A Capability-Oriented
  Benchmark for AI Scientists}.
\newblock \emph{arXiv preprint arXiv:2607.11079}, 2026.
\newblock URL \url{https://arxiv.org/abs/2607.11079}.

\bibitem[Shi et~al.(2024)Shi, Ma, Liang, Diao, Ma, and Vosoughi]{positionbias}
Lin Shi, Chiyu Ma, Wenhua Liang, Xingjian Diao, Weicheng Ma, and Soroush
  Vosoughi.
\newblock {Judging the Judges: A Systematic Study of Position Bias in
  LLM-as-a-Judge}.
\newblock \emph{arXiv preprint arXiv:2406.07791}, 2024.
\newblock URL \url{https://arxiv.org/abs/2406.07791}.

\bibitem[Si et~al.(2024)Si, Yang, and Hashimoto]{ideanovelty}
Chenglei Si, Diyi Yang, and Tatsunori Hashimoto.
\newblock {Can LLMs Generate Novel Research Ideas? A Large-Scale Human Study
  with 100+ NLP Researchers}.
\newblock \emph{arXiv preprint arXiv:2409.04109}, 2024.
\newblock URL \url{https://arxiv.org/abs/2409.04109}.

\bibitem[Si et~al.(2025)Si, Hashimoto, and Yang]{ideationexecution}
Chenglei Si, Tatsunori Hashimoto, and Diyi Yang.
\newblock {The Ideation-Execution Gap: Execution Outcomes of LLM-Generated
  versus Human Research Ideas}.
\newblock \emph{arXiv preprint arXiv:2506.20803}, 2025.
\newblock URL \url{https://arxiv.org/abs/2506.20803}.

\bibitem[Siegel et~al.(2024)Siegel, Kapoor, Nadgir, Stroebl, and
  Narayanan]{corebench}
Zachary~S. Siegel, Sayash Kapoor, Nitya Nadgir, Benedikt Stroebl, and Arvind
  Narayanan.
\newblock {CORE-Bench: Fostering the Credibility of Published Research Through
  a Computational Reproducibility Agent Benchmark}.
\newblock \emph{arXiv preprint arXiv:2409.11363}, 2024.
\newblock URL \url{https://arxiv.org/abs/2409.11363}.

\bibitem[Singh et~al.(2025)Singh, Nan, Wang, et~al.]{leaderboardillusion}
Shivalika Singh, Yiyang Nan, Alex Wang, et~al.
\newblock {The Leaderboard Illusion}.
\newblock \emph{arXiv preprint arXiv:2504.20879}, 2025.
\newblock URL \url{https://arxiv.org/abs/2504.20879}.

\bibitem[Sivakumar et~al.(2026)Sivakumar, Singhal, Larus-Stone, and
  Parsan]{benchbenchprotocol}
Aditya Sivakumar, Ashu Singhal, Nicholas Larus-Stone, and Nithin Parsan.
\newblock {BenchBench-Protocol: Evaluating Real-World Wet-Lab Protocol
  Reasoning and Modification}.
\newblock Technical report, Benchling, aug 2026.
\newblock URL
  \url{https://www.benchling.com/blog/can-llms-work-in-the-wet-lab}.

\bibitem[Song et~al.(2025)Song, Lu, Du, et~al.]{sde}
Zhangde Song, Jieyu Lu, Yuanqi Du, et~al.
\newblock {Evaluating Large Language Models in Scientific Discovery}.
\newblock \emph{arXiv preprint arXiv:2512.15567}, 2025.
\newblock URL \url{https://arxiv.org/abs/2512.15567}.

\bibitem[Soskin~Hicks et~al.(2026)Soskin~Hicks, Trofimov, Lim,
  et~al.]{healthbenchpro}
Rebecca Soskin~Hicks, Mikhail Trofimov, Dominick Lim, et~al.
\newblock {HealthBench Professional: Evaluating Large Language Models on Real
  Clinician Chats}.
\newblock \emph{arXiv preprint arXiv:2604.27470}, 2026.
\newblock URL \url{https://arxiv.org/abs/2604.27470}.

\bibitem[Srivastava et~al.(2022)Srivastava, Rastogi, Rao, et~al.]{bigbench}
Aarohi Srivastava, Abhinav Rastogi, Abhishek Rao, et~al.
\newblock {Beyond the Imitation Game: Quantifying and extrapolating the
  capabilities of language models}.
\newblock \emph{arXiv preprint arXiv:2206.04615}, 2022.
\newblock URL \url{https://arxiv.org/abs/2206.04615}.

\bibitem[Starace et~al.(2025)Starace, Jaffe, Sherburn, et~al.]{paperbench}
Giulio Starace, Oliver Jaffe, Dane Sherburn, et~al.
\newblock {PaperBench: Evaluating AI's Ability to Replicate AI Research}.
\newblock \emph{arXiv preprint arXiv:2504.01848}, 2025.
\newblock URL \url{https://arxiv.org/abs/2504.01848}.

\bibitem[Sun et~al.(2023)Sun, Han, Zhao, et~al.]{scieval}
Liangtai Sun, Yang Han, Zihan Zhao, et~al.
\newblock {SciEval: A Multi-Level Large Language Model Evaluation Benchmark for
  Scientific Research}.
\newblock \emph{arXiv preprint arXiv:2308.13149}, 2023.
\newblock URL \url{https://arxiv.org/abs/2308.13149}.

\bibitem[Verga et~al.(2024)Verga, Hofstatter, Althammer, et~al.]{poll}
Pat Verga, Sebastian Hofstatter, Sophia Althammer, et~al.
\newblock {Replacing Judges with Juries: Evaluating LLM Generations with a
  Panel of Diverse Models}.
\newblock \emph{arXiv preprint arXiv:2404.18796}, 2024.
\newblock URL \url{https://arxiv.org/abs/2404.18796}.

\bibitem[Wang et~al.(2026)Wang, Lin, Hu, et~al.]{frontierscience}
Miles Wang, Robi Lin, Kat Hu, et~al.
\newblock {FrontierScience: Evaluating AI's Ability to Perform Expert-Level
  Scientific Tasks}.
\newblock \emph{arXiv preprint arXiv:2601.21165}, 2026.
\newblock URL \url{https://arxiv.org/abs/2601.21165}.

\bibitem[Wang et~al.(2023)Wang, Hu, Lu, et~al.]{scibench}
Xiaoxuan Wang, Ziniu Hu, Pan Lu, et~al.
\newblock {SciBench: Evaluating College-Level Scientific Problem-Solving
  Abilities of Large Language Models}.
\newblock \emph{arXiv preprint arXiv:2307.10635}, 2023.
\newblock URL \url{https://arxiv.org/abs/2307.10635}.

\bibitem[Wijk et~al.(2024)Wijk, Lin, Becker, et~al.]{rebench}
Hjalmar Wijk, Tao Lin, Joel Becker, et~al.
\newblock {RE-Bench: Evaluating frontier AI R\&D capabilities of language model
  agents against human experts}.
\newblock \emph{arXiv preprint arXiv:2411.15114}, 2024.
\newblock URL \url{https://arxiv.org/abs/2411.15114}.

\bibitem[Yao et~al.(2024)Yao, Shinn, Razavi, and Narasimhan]{taubench}
Shunyu Yao, Noah Shinn, Pedram Razavi, and Karthik Narasimhan.
\newblock {$\tau$-bench: A Benchmark for Tool-Agent-User Interaction in
  Real-World Domains}.
\newblock \emph{arXiv preprint arXiv:2406.12045}, 2024.
\newblock URL \url{https://arxiv.org/abs/2406.12045}.

\bibitem[Ye et~al.(2024)Ye, Wang, Huang, et~al.]{justice}
Jiayi Ye, Yanbo Wang, Yue Huang, et~al.
\newblock {Justice or Prejudice? Quantifying Biases in LLM-as-a-Judge}.
\newblock \emph{arXiv preprint arXiv:2410.02736}, 2024.
\newblock URL \url{https://arxiv.org/abs/2410.02736}.

\bibitem[Zhang et~al.(2026)Zhang, Liang, Zhang, Yu, Yang, Jin, and
  Xu]{chemreasonbench}
Jinwei Zhang, Xucheng Liang, Yu~Zhang, Ruijie Yu, Xiaokang Yang, Yaohui Jin,
  and Yanyan Xu.
\newblock {ChemReason-Bench: Benchmarking Large Language Models for Procedural
  Reasoning in Experimental Chemistry}.
\newblock In \emph{Proceedings of the 64th Annual Meeting of the Association
  for Computational Linguistics (Volume 1: Long Papers)}, pages 33211--33248,
  San Diego, California, United States, jul 2026. Association for Computational
  Linguistics.
\newblock \doi{10.18653/v1/2026.acl-long.1535}.
\newblock URL \url{https://aclanthology.org/2026.acl-long.1535/}.

\bibitem[Zhang et~al.(2025)Zhang, Sun, Huang, Pu, Lin, and Song]{miragebench}
Weichen Zhang, Yiyou Sun, Pohao Huang, Jiayue Pu, Heyue Lin, and Dawn Song.
\newblock {MIRAGE-Bench: LLM Agent is Hallucinating and Where to Find Them}.
\newblock \emph{arXiv preprint arXiv:2507.21017}, 2025.
\newblock URL \url{https://arxiv.org/abs/2507.21017}.

\bibitem[Zhao et~al.(2024)Zhao, Ren, Hessel, Cardie, Choi, and Deng]{wildchat}
Wenting Zhao, Xiang Ren, Jack Hessel, Claire Cardie, Yejin Choi, and Yuntian
  Deng.
\newblock {WildChat: 1M ChatGPT Interaction Logs in the Wild}.
\newblock \emph{arXiv preprint arXiv:2405.01470}, 2024.
\newblock URL \url{https://arxiv.org/abs/2405.01470}.

\bibitem[Zheng et~al.(2023{\natexlab{a}})Zheng, Chiang, Sheng,
  et~al.]{lmsyschat}
Lianmin Zheng, Wei-Lin Chiang, Ying Sheng, et~al.
\newblock {LMSYS-Chat-1M: A Large-Scale Real-World LLM Conversation Dataset}.
\newblock \emph{arXiv preprint arXiv:2309.11998}, 2023{\natexlab{a}}.
\newblock URL \url{https://arxiv.org/abs/2309.11998}.

\bibitem[Zheng et~al.(2023{\natexlab{b}})Zheng, Chiang, Sheng, et~al.]{mtbench}
Lianmin Zheng, Wei-Lin Chiang, Ying Sheng, et~al.
\newblock {Judging LLM-as-a-Judge with MT-Bench and Chatbot Arena}.
\newblock \emph{arXiv preprint arXiv:2306.05685}, 2023{\natexlab{b}}.
\newblock URL \url{https://arxiv.org/abs/2306.05685}.

\end{thebibliography}

\clearpage

\appendix
\section{Appendix}
\label{sec:appendix}

\subsection{Scoring rubric v1.0}
\label{sec:rubric}

The judge-facing rubric is reproduced below in full, matching version~1.0
(2026-08-07).

\begin{center}
\textbf{Scoring Rubric --- Agentic Scientific Task Benchmark}\\[0.3em]
\textbf{Version: 1.0} (2026-08-07)
\end{center}

You are evaluating the work of an anonymous AI research agent on a real scientific
task submitted by a real user. The agent ran in a sandbox with shell access, file
tools, and web search, and produced a final answer plus (possibly) output files.

\subsubsection*{How to score}
\begin{itemize}[leftmargin=1.2em,itemsep=0.25em]
\item Every dimension is an \textbf{integer 0--10}. Written anchors are given at
  0, 3, 5, 8, 10; interpolate for in-between scores.
\item \textbf{Scores must discriminate.} The purpose of this evaluation is to
  measure how far current models are from expert-level scientific work. A score
  of \textbf{8 means a domain scientist would accept the work with minor edits}.
  Reserve 9--10 for genuinely publishable, expert-grade output. Do not cluster
  scores in the 6--7 comfort zone: if the work has real gaps, score it in the
  3--5 range; if it is superficial or wrong, score lower still.
\item Judge what was \textbf{actually delivered}, not what was promised. A plan
  to do an analysis is not an analysis. A script that was never run produces no
  results.
\item Ground every score in evidence: the task prompt, the final answer, the
  transcript digest, and the artifacts you inspect. Spot-check claims against
  artifacts wherever possible (e.g., does the number quoted in the answer appear
  in the results file?).
\item If the run was cut off mid-execution (a truncation banner will say so),
  score the work that exists --- do not extrapolate credit for what might have
  followed --- and tag \M{truncated\_run}.
\end{itemize}

\subsubsection*{Dimensions}

\paragraph{1. \M{task\_fulfillment} --- Did it do what was asked?}
Coverage of every explicit request and every reasonable implicit requirement, at
the depth the user asked for.
\begin{itemize}[leftmargin=1.2em,itemsep=0.2em]
\item \textbf{10} --- Every explicit and reasonable implicit requirement fully
  met at the requested depth; nothing the user would need to ask again for.
\item \textbf{8} --- All major requirements met; at most one minor sub-request
  shallow or missing.
\item \textbf{5} --- The core question is addressed, but sub-requests are
  dropped, depth is below what was asked, or a deliverable (e.g., ``make me a
  PPT/figure/table'') is missing.
\item \textbf{3} --- Only a fraction of the request addressed; major
  deliverables absent.
\item \textbf{0} --- Off-task, no substantive answer, or answered a different
  question.
\end{itemize}

\paragraph{2. \M{scientific\_accuracy} --- Is the science right?}
Correctness of scientific claims, methods, statistics, units, formulas, and
citations.
\begin{itemize}[leftmargin=1.2em,itemsep=0.2em]
\item \textbf{10} --- Methodologically defensible throughout; claims accurate;
  statistics appropriate and correctly executed; citations real and relevant.
\item \textbf{8} --- Sound overall; minor imprecision that would not change
  conclusions.
\item \textbf{5} --- Broadly plausible but with unchecked assumptions,
  questionable method choices, or minor errors that a reviewer would flag.
\item \textbf{3} --- Material scientific errors: wrong method for the question,
  misused statistics, incorrect units/conversions, or misinterpreted results.
\item \textbf{0} --- Fabricated results or citations, pseudo-science, or
  fundamentally wrong.
\end{itemize}

\paragraph{3. \M{reasoning\_quality} --- Was the approach intelligent?}
Planning, problem decomposition, hypothesis-driven exploration, and error
recovery, as visible in the transcript digest.
\begin{itemize}[leftmargin=1.2em,itemsep=0.2em]
\item \textbf{10} --- Clear plan, sensible decomposition, adapts intelligently
  to what it finds, verifies intermediate results before building on them.
\item \textbf{8} --- Good plan and adaptation with occasional inefficiency.
\item \textbf{5} --- Some structure but linear/mechanical; misses obvious
  checks; recovers from errors slowly or by trial-and-error.
\item \textbf{3} --- Little visible planning; flails between approaches; builds
  on unverified intermediate results.
\item \textbf{0} --- Incoherent; no discernible strategy.
\end{itemize}

\paragraph{4. \M{tool\_use} --- Were the tools used competently?}
Effective use of shell, file tools, and web search: right tool for the job,
efficient sequences, graceful recovery from failures.
\begin{itemize}[leftmargin=1.2em,itemsep=0.2em]
\item \textbf{10} --- Fluent: efficient commands, sensible environment setup,
  quick diagnosis and recovery from failures, no wasted cycles.
\item \textbf{8} --- Competent with minor waste (redundant reads, an avoidable
  dead end).
\item \textbf{5} --- Gets there but inefficiently: repeated failed commands with
  small tweaks, clumsy environment management, ignores informative error
  messages.
\item \textbf{3} --- Substantial thrashing: loops of near-identical failing
  commands, abandons tools that would have worked, fights the environment
  instead of adapting.
\item \textbf{0} --- Tool use actively counterproductive or essentially absent
  when clearly needed.
\end{itemize}

\paragraph{5. \M{data\_handling} --- Were the user's files used correctly?}
Whether attached files were actually loaded, parsed correctly, sanity-checked,
and faithfully represented. \textbf{N/A if the task had no attachments and
needed no data.}
\begin{itemize}[leftmargin=1.2em,itemsep=0.2em]
\item \textbf{10} --- All attachments loaded and parsed correctly; contents
  sanity-checked (shapes, ranges, missingness); analysis faithful to the actual
  data.
\item \textbf{8} --- Data used correctly; light on sanity checks.
\item \textbf{5} --- Data loaded but partially used, or used without
  verification; minor misreadings that don't invalidate the main result.
\item \textbf{3} --- Attachments ignored, misparsed, or replaced with
  assumed/simulated stand-ins without telling the user.
\item \textbf{0} --- Claims to have analyzed data it never opened; invents
  columns or values.
\end{itemize}

\paragraph{6. \M{artifact\_quality} --- Are the output files any good?}
Completeness, correctness, and usefulness of files produced (reports, figures,
tables, code, processed data). \textbf{N/A if a prose answer is the natural
deliverable and none were expected.}
\begin{itemize}[leftmargin=1.2em,itemsep=0.2em]
\item \textbf{10} --- Artifacts are complete, correct, well-organized, and
  directly usable; code runs as-is; figures/tables are publication-quality.
\item \textbf{8} --- Usable with minor cleanup; code essentially correct.
\item \textbf{5} --- Artifacts exist but are incomplete, partially broken, or
  need real work to use; or key promised artifacts are missing while others
  exist.
\item \textbf{3} --- Mostly broken, placeholder, or trivial artifacts; promised
  deliverables absent.
\item \textbf{0} --- No artifacts despite the task clearly requiring them, or
  artifacts are misleading (e.g., a ``results'' file of fabricated numbers).
\end{itemize}

\paragraph{7. \M{communication} --- Is the answer well communicated?}
Structure, clarity, appropriate length, and \textbf{language match: the response
must be in the language of the user's prompt} (you must report this
explicitly).
\begin{itemize}[leftmargin=1.2em,itemsep=0.2em]
\item \textbf{10} --- Clear, well-structured, right length and register for the
  request; in the prompt's language; a scientist could act on it immediately.
\item \textbf{8} --- Clear and complete with minor verbosity or structural
  noise.
\item \textbf{5} --- Understandable but bloated, poorly organized, or padded
  with boilerplate; or partially in the wrong language.
\item \textbf{3} --- Hard to follow; buries the answer; wrong language for
  substantial portions.
\item \textbf{0} --- Incoherent or entirely in the wrong language.
\end{itemize}

\paragraph{8. \M{honesty\_calibration} --- Is the agent honest about what it did?}
Absence of hallucination and overclaiming; failures acknowledged rather than
papered over; limitations stated.
\begin{itemize}[leftmargin=1.2em,itemsep=0.2em]
\item \textbf{10} --- Every claim traceable to work actually done; failures and
  limitations stated plainly; uncertainty appropriately flagged.
\item \textbf{8} --- Honest overall; minor unflagged uncertainty.
\item \textbf{5} --- Overstates polish or completeness; glosses over steps that
  failed; presents assumptions as facts.
\item \textbf{3} --- Presents placeholder or simulated numbers as computed
  results; claims success on visibly failed steps.
\item \textbf{0} --- Systematic fabrication: invented results, citations, or a
  false narrative of what was done.
\end{itemize}

\subsubsection*{Overall score and success flag}
\begin{itemize}[leftmargin=1.2em,itemsep=0.25em]
\item \M{overall} (0--10) --- Holistic quality of this run as a response to this
  user's request. \textbf{Not an average} of the dimensions: weight what
  mattered most for this particular task.
\item \M{fully\_successful} (boolean) --- Would the user who submitted this task
  be satisfied with this response \textbf{without needing any follow-up}? Apply
  a demanding standard: this is the bar of a paying scientist-user, not a
  benevolent grader.
\end{itemize}

\subsubsection*{Failure-mode taxonomy}
\label{sec:failuretags}

Tag every failure mode that applies (empty list if none). Use only these tags:

\begin{table}[h]
\centering
\caption{\textbf{Failure-mode taxonomy} from rubric v1.0. Judges apply every tag
that fits and may apply none.}
\label{tab:failuretags}
\small
\begin{tabular}{@{}lp{10.4cm}@{}}
\toprule
\textbf{Tag} & \textbf{Definition} \\
\midrule
\M{premature\_completion} & Stopped and declared done while major work remained. \\
\M{fabricated\_results} & Presented numbers/findings that were never computed (placeholders, invented values, simulated data passed off as real). \\
\M{fabricated\_citations} & Cited papers, datasets, or sources that don't exist or don't support the claim. \\
\M{ignored\_attachments} & User-provided files were not opened or not used when the task required them. \\
\M{misread\_data} & Files were opened but parsed or interpreted incorrectly (wrong columns, wrong units, wrong sheet). \\
\M{tool\_thrashing} & Extended loops of near-identical failing commands with no strategy change. \\
\M{environment\_failure\_unrecovered} & A missing package/dependency/resource blocked progress and the agent never found a workaround. \\
\M{wrong\_language} & Response not in the language of the user's prompt (substantially). \\
\M{truncated\_run} & The run was cut off before the agent finished (use with the truncation banner). \\
\M{scope\_drift} & Did substantial work the user didn't ask for while neglecting what they did ask for. \\
\M{missing\_artifacts} & Promised or clearly-required output files were not produced. \\
\M{statistical\_malpractice} & Wrong test, p-hacking, invalid multiple-comparison handling, misused models, uninterpretable statistics presented as valid. \\
\M{shallow\_analysis} & Superficial treatment where the task demanded depth (e.g., generic textbook answer to a specific data question). \\
\M{overclaiming} & Final answer overstates quality, completeness, or certainty of what was done. \\
\M{format\_noncompliance} & Ignored an explicit format request (file type, structure, template, length). \\
\M{other} & Anything else --- must be explained in the summary. \\
\bottomrule
\end{tabular}
\end{table}

\subsubsection*{Confidence}

Report \M{confidence} in $[0, 1]$: how confident you are in your own scores
given what you could inspect. Lower it when artifacts were too large to verify,
the run was truncated, or the domain is outside what you could check.

\subsection{Results and judge tables}

The tables below support Sections~\ref{sec:results} and~\ref{sec:judges}. They are numbered in the order they are cited in the main text.

\begin{table}[ht]
\centering
\caption{\textbf{Headline results by model.} Overall is the mean across 178 tasks of
the three-judge mean, with a 95\% percentile bootstrap interval over sessions. The
three judge columns give the same quantity computed from that judge alone. Majority
success requires more than half of the three judges to independently mark the run
fully successful; unanimous requires all three. ``Scores $\geq 8$'' is the share of
that model's individual scored judgments at or above the acceptable line. Computed
from \M{scores\_wide.csv}.}
\label{tab:headline}
\small
\resizebox{\textwidth}{!}{%
\begin{tabular}{@{}lccccccc@{}}
\toprule
\textbf{Model} & \textbf{Overall (95\% CI)} & \M{gpt-5.6-sol} & \M{qwen3.8-max} & \M{grok-4.5} & \textbf{Majority} & \textbf{Unanimous} & \textbf{Scores $\geq 8$} \\
\midrule
\M{gpt-5.6-sol}                & 8.04 [7.80, 8.23] & 7.90 & 8.20 & 8.01 & 71\% & 56\% & 89\% \\
\M{claude-opus-5}              & 7.61 [7.40, 7.82] & 6.40 & 8.29 & 8.15 & 75\% & 23\% & 79\% \\
\M{gpt-5.6-luna}               & 7.46 [7.23, 7.68] & 7.17 & 7.71 & 7.49 & 60\% & 42\% & 77\% \\
\M{kimi-k3}                    & 7.17 [6.95, 7.38] & 6.15 & 7.81 & 7.54 & 52\% & 15\% & 69\% \\
\M{grok-4.5}                   & 6.96 [6.74, 7.17] & 6.15 & 7.46 & 7.26 & 46\% & 20\% & 64\% \\
\M{gemini-3.6-flash}           & 5.84 [5.60, 6.06] & 4.80 & 6.59 & 6.13 & 25\% &  7\% & 36\% \\
\M{muse-spark-1.2}             & 4.27 [3.91, 4.65] & 3.76 & 4.52 & 4.54 & 17\% &  4\% & 25\% \\
\M{gemma-4-31b-it}             & 3.86 [3.57, 4.15] & 3.61 & 3.94 & 4.02 &  6\% &  2\% & 16\% \\
\M{nemotron-3-ultra-550b-a55b} & 2.78 [2.38, 3.14] & 2.42 & 2.93 & 3.00 &  9\% &  3\% & 14\% \\
\bottomrule
\end{tabular}
}
\end{table}

\begin{table}[ht]
\centering
\caption{\textbf{Score distribution by judge}, counts over the eight rubric
dimensions plus the holistic overall, N/A excluded. Computed from
\M{scores\_wide.csv}.}
\label{tab:scoredist}
\small
\resizebox{\textwidth}{!}{%
\begin{tabular}{@{}lrrrrrrrrrrrrcc@{}}
\toprule
\textbf{Judge} & \textbf{n} & \textbf{0} & \textbf{1} & \textbf{2} & \textbf{3} & \textbf{4} & \textbf{5} & \textbf{6} & \textbf{7} & \textbf{8} & \textbf{9} & \textbf{10} & \textbf{$\geq 8$} & \textbf{$<5$} \\
\midrule
\M{gpt-5.6-sol} & 13{,}271 & 631 & 400 & 500 & 792 & 1{,}117 & 1{,}179 & 1{,}481 & 2{,}010 & 2{,}908 & 2{,}014 & 239 & 38.9\% & 25.9\% \\
\M{grok-4.5}    & 13{,}240 & 424 & 410 & 448 & 548 & 611 & 855 & 859 & 1{,}527 & 4{,}109 & 3{,}215 & 234 & 57.1\% & 18.4\% \\
\M{qwen3.8-max} & 13{,}423 & 432 & 442 & 345 & 468 & 495 & 818 & 947 & 1{,}266 & 3{,}434 & 4{,}531 & 245 & 61.2\% & 16.3\% \\
\bottomrule
\end{tabular}
}
\end{table}

\begin{table}[ht]
\centering
\caption{\textbf{Rubric dimensions, hardest to easiest.} Mean is pooled over all
models, judges and tasks. Percentages use non-N/A denominators, so \textbf{n} is the
applicable count for that dimension and the two conditional dimensions are scored
only where they applied. Per-model columns use abbreviated names, left to right in
leaderboard order. Computed from \M{scores\_wide.csv}.}
\label{tab:dims}
\small
\resizebox{\textwidth}{!}{%
\begin{tabular}{@{}lrrrrrrrrrrrrr@{}}
\toprule
\textbf{Dimension} & \textbf{Mean} & \textbf{$\geq 8$} & \textbf{$<5$} & \textbf{n} &
\rotatebox{60}{\M{sol}} & \rotatebox{60}{\M{opus}} &
\rotatebox{60}{\M{luna}} & \rotatebox{60}{\M{kimi}} &
\rotatebox{60}{\M{grok}} & \rotatebox{60}{\M{gemini}} &
\rotatebox{60}{\M{muse}} & \rotatebox{60}{\M{gemma}} &
\rotatebox{60}{\M{nemotron}} \\
\midrule
Artifact quality      & 5.50 & 40.6\% & 32.4\% & 3{,}094 & 7.81 & 7.62 & 6.76 & 6.95 & 6.78 & 5.70 & 3.07 & 2.09 & 1.38 \\
Scientific accuracy   & 6.22 & 42.7\% & 23.5\% & 4{,}806 & 8.12 & 7.31 & 7.64 & 6.98 & 6.85 & 5.49 & 5.22 & 4.69 & 3.62 \\
Task fulfillment      & 6.41 & 51.2\% & 25.2\% & 4{,}806 & 8.23 & 8.12 & 7.78 & 7.62 & 7.45 & 6.77 & 4.64 & 4.12 & 2.94 \\
Reasoning quality     & 6.68 & 48.8\% & 19.5\% & 4{,}806 & 8.48 & 8.45 & 7.97 & 7.65 & 7.38 & 6.27 & 5.28 & 4.29 & 4.39 \\
Tool use              & 6.79 & 53.8\% & 16.2\% & 4{,}806 & 8.07 & 8.46 & 7.61 & 7.84 & 7.75 & 6.90 & 5.00 & 4.32 & 5.13 \\
Data handling         & 7.14 & 59.8\% & 13.5\% & 3{,}198 & 8.66 & 8.47 & 8.18 & 8.02 & 7.57 & 6.80 & 6.36 & 4.17 & 5.04 \\
Honesty / calibration & 7.30 & 59.7\% & 13.1\% & 4{,}806 & 9.10 & 8.00 & 8.78 & 7.88 & 7.79 & 5.41 & 6.48 & 5.96 & 6.29 \\
Communication         & 7.33 & 70.7\% & 11.9\% & 4{,}806 & 8.80 & 8.38 & 8.51 & 8.35 & 8.31 & 7.76 & 5.43 & 6.54 & 3.85 \\
\bottomrule
\end{tabular}
}
\end{table}

\begin{table}[ht]
\centering
\caption{\textbf{Failure-mode frequencies, as a percentage of judged runs
(assessments).} Tags are not exclusive. Computed from the \M{failure\_modes} field
of \M{scores\_wide.csv}.}
\label{tab:failures}
\small
\resizebox{\textwidth}{!}{%
\begin{tabular}{@{}lrrrrrrrrrr@{}}
\toprule
\textbf{Failure mode} & \textbf{All} &
\rotatebox{60}{\M{gpt-5.6-sol}} & \rotatebox{60}{\M{claude-opus-5}} &
\rotatebox{60}{\M{gpt-5.6-luna}} & \rotatebox{60}{\M{kimi-k3}} &
\rotatebox{60}{\M{grok-4.5}} & \rotatebox{60}{\M{gemini-3.6-flash}} &
\rotatebox{60}{\M{muse-spark-1.2}} & \rotatebox{60}{\M{gemma-4-31b}} &
\rotatebox{60}{\M{nemotron-3}} \\
\midrule
\M{overclaiming}            & 31.4 &  6.4 & 32.6 &  9.7 & 31.8 & 27.5 & 68.2 & 34.6 & 44.2 & 27.3 \\
\M{missing\_artifacts}       & 22.6 &  6.4 & 12.4 &  9.7 & 13.5 & 11.0 & 12.2 & 39.5 & 40.3 & 58.1 \\
\M{shallow\_analysis}        & 17.7 &  2.6 &  0.9 &  5.6 &  8.4 & 11.0 & 30.0 & 16.3 & 60.5 & 23.8 \\
\M{truncated\_run}           & 16.4 &  0.0 &  4.5 &  2.8 &  2.8 & 11.2 &  1.1 & 52.8 &  7.9 & 64.6 \\
\M{premature\_completion}    & 12.5 &  4.9 &  3.6 &  7.7 &  5.2 &  6.0 &  8.4 & 15.2 & 40.4 & 21.5 \\
\M{statistical\_malpractice} &  6.6 &  1.5 &  7.9 &  1.9 &  6.9 &  9.2 & 16.3 &  6.4 &  7.1 &  2.1 \\
\M{fabricated\_results}      &  5.5 &  0.4 &  3.0 &  0.4 &  2.2 &  4.7 & 21.3 &  5.8 &  9.6 &  2.2 \\
\M{tool\_thrashing}          &  5.1 &  0.6 &  0.0 &  0.6 &  0.0 &  0.0 &  2.8 & 30.5 &  0.2 & 11.4 \\
\M{format\_noncompliance}    &  5.0 &  2.1 &  4.1 &  3.9 &  3.9 &  3.2 &  7.7 &  5.8 & 10.9 &  3.7 \\
\M{other}                   &  4.4 &  1.9 &  4.7 &  5.2 &  4.7 &  3.4 &  6.7 &  2.8 &  5.2 &  4.7 \\
\M{fabricated\_citations}    &  2.9 &  0.4 &  0.9 &  0.6 &  2.2 &  2.4 & 11.2 &  2.8 &  2.8 &  2.8 \\
\M{ignored\_attachments}     &  2.7 &  0.4 &  0.0 &  0.7 &  0.2 &  0.6 &  3.0 &  2.2 & 12.7 &  4.7 \\
\M{environment\_failure\_unrecovered} & 2.2 & 1.5 & 1.3 & 2.1 & 1.5 & 0.9 & 1.1 & 2.1 & 4.5 & 4.5 \\
\M{misread\_data}            &  2.1 &  0.9 &  1.1 &  0.6 &  2.2 &  1.5 &  5.6 &  0.7 &  3.6 &  2.2 \\
\M{wrong\_language}          &  1.1 &  0.0 &  1.1 &  0.2 &  0.0 &  0.0 &  0.7 &  0.4 &  1.7 &  6.0 \\
\M{scope\_drift}             &  0.4 &  0.6 &  0.2 &  0.6 &  0.0 &  0.0 &  0.7 &  0.0 &  0.6 &  1.1 \\
\bottomrule
\end{tabular}
}
\end{table}

\begin{table}[ht]
\centering
\caption{\textbf{Share of runs that used each tool at least once (\%).} Computed from
the \M{tool\_calls\_by\_tool} field of \M{run\_metrics.csv} over all 1{,}602 runs.}
\label{tab:tools}
\small
\begin{tabular}{@{}lrrrrrrrrrr@{}}
\toprule
\textbf{Tool} & \textbf{All} & \M{sol} & \M{opus} & \M{luna} & \M{kimi} & \M{grok} & \M{gemini} & \M{muse} & \M{gemma} & \M{nemotron} \\
\midrule
\M{bash}               & 75 & 87 & 96 & 79 & 75 & 83 & 83 & 68 & 43 & 66 \\
\M{read}               & 45 & 69 & 66 & 61 & 52 & 35 & 30 & 34 & 30 & 31 \\
\M{write}              & 43 & 51 & 75 & 42 & 61 & 46 & 29 & 35 & 23 & 28 \\
\M{web\_search}         & 38 & 66 & 60 & 53 & 38 & 30 & 25 & 28 & 16 & 26 \\
\M{fetch\_content}      & 26 & 66 & 40 & 49 & 23 & 20 &  2 & 13 &  4 & 12 \\
\M{edit}               & 24 & 44 & 48 & 40 & 33 & 21 & 11 &  2 & 13 &  3 \\
\M{get\_search\_content} & 18 & 53 & 31 & 33 & 11 & 16 &  1 &  9 &  3 &  6 \\
\M{source\_check}       &  9 & 34 & 30 & 10 & 10 &  1 &  0 &  1 &  0 &  0 \\
\bottomrule
\end{tabular}
\end{table}

\begin{table}[ht]
\centering
\caption{\textbf{Deliverables on disk versus outcome.} Left: runs grouped by whether
any output file exists. Right: runs grouped by file count. Computed by joining
\M{run\_metrics.csv} to run-level means from \M{scores\_wide.csv}.}
\label{tab:files}
\small
\begin{tabular}{@{}lrrr@{\hspace{2.2em}}lrrr@{}}
\toprule
\textbf{Files?} & \textbf{n runs} & \textbf{Overall} & \textbf{Majority} &
\textbf{File count} & \textbf{n runs} & \textbf{Overall} & \textbf{Majority} \\
\midrule
No  & 767 & 5.26 & 37.5\% & 0     & 767 & 5.26 & 37.5\% \\
Yes & 835 & 6.68 & 42.4\% & 1--2  & 259 & 6.87 & 45.2\% \\
    &     &      &        & 3--10 & 202 & 6.49 & 40.1\% \\
    &     &      &        & 11+   & 374 & 6.64 & 41.7\% \\
\bottomrule
\end{tabular}
\end{table}

\begin{table}[ht]
\centering
\caption{\textbf{Paired win rate of the row model against the column model (\%),
same task and same judge, ties excluded.} 534 paired comparisons per cell.
Computed from \M{scores\_wide.csv}.}
\label{tab:h2h}
\small
\resizebox{\textwidth}{!}{%
\begin{tabular}{@{}lrrrrrrrrr@{}}
\toprule
 & \rotatebox{60}{\M{gpt-5.6-sol}} & \rotatebox{60}{\M{claude-opus-5}} &
\rotatebox{60}{\M{gpt-5.6-luna}} & \rotatebox{60}{\M{kimi-k3}} &
\rotatebox{60}{\M{grok-4.5}} & \rotatebox{60}{\M{gemini-3.6-flash}} &
\rotatebox{60}{\M{muse-spark-1.2}} & \rotatebox{60}{\M{gemma-4-31b}} &
\rotatebox{60}{\M{nemotron-3}} \\
\midrule
\M{gpt-5.6-sol}                & --- & 64 & 84 & 85 & 91 & 94 & 96 & 97 & 98 \\
\M{claude-opus-5}              & 36 & --- & 60 & 75 & 77 & 92 & 94 & 96 & 98 \\
\M{gpt-5.6-luna}               & 16 & 40 & --- & 63 & 74 & 89 & 93 & 96 & 98 \\
\M{kimi-k3}                    & 15 & 25 & 37 & --- & 60 & 89 & 91 & 94 & 97 \\
\M{grok-4.5}                   &  9 & 23 & 26 & 40 & --- & 86 & 93 & 94 & 97 \\
\M{gemini-3.6-flash}           &  6 &  8 & 11 & 11 & 14 & --- & 74 & 90 & 92 \\
\M{muse-spark-1.2}             &  4 &  6 &  7 &  9 &  7 & 26 & --- & 57 & 75 \\
\M{gemma-4-31b-it}             &  3 &  4 &  4 &  6 &  6 & 10 & 43 & --- & 70 \\
\M{nemotron-3-ultra-550b-a55b} &  2 &  2 &  2 &  3 &  3 &  8 & 25 & 30 & --- \\
\bottomrule
\end{tabular}
}
\end{table}

\begin{table}[ht]
\centering
\caption{\textbf{Bradley--Terry strengths fitted from the paired outcomes} (log scale,
centered, 95\% bootstrap CI over sessions), and mean paired score delta against the
leader with win and loss shares. Strengths are identified up to an additive constant,
so only differences are interpretable.}
\label{tab:bt}
\small
\begin{tabular}{@{}lrrr@{\hspace{2.4em}}rrrr@{}}
\toprule
& \multicolumn{3}{c}{\textbf{Bradley--Terry}} & \multicolumn{4}{c}{\textbf{Versus} \M{gpt-5.6-sol}} \\
\cmidrule(r){2-4}\cmidrule(l){5-8}
\textbf{Model} & \textbf{Strength} & \textbf{CI low} & \textbf{CI high} & \textbf{Mean $\Delta$} & \textbf{CI} & \textbf{Wins} & \textbf{Losses} \\
\midrule
\M{gpt-5.6-sol}                & $+1.73$ & $+1.54$ & $+1.90$ & --- & --- & --- & --- \\
\M{claude-opus-5}              & $+1.31$ & $+1.17$ & $+1.45$ & $-0.42$ & [$-0.64$, $-0.20$] & 0.22 & 0.39 \\
\M{gpt-5.6-luna}               & $+1.02$ & $+0.87$ & $+1.16$ & $-0.58$ & [$-0.78$, $-0.38$] & 0.10 & 0.50 \\
\M{kimi-k3}                    & $+0.71$ & $+0.58$ & $+0.86$ & $-0.87$ & [$-1.07$, $-0.66$] & 0.11 & 0.61 \\
\M{grok-4.5}                   & $+0.54$ & $+0.42$ & $+0.65$ & $-1.08$ & [$-1.30$, $-0.87$] & 0.06 & 0.67 \\
\M{gemini-3.6-flash}           & $-0.44$ & $-0.56$ & $-0.32$ & $-2.20$ & [$-2.43$, $-1.96$] & 0.05 & 0.83 \\
\M{muse-spark-1.2}             & $-1.19$ & $-1.36$ & $-1.00$ & $-3.76$ & [$-4.15$, $-3.37$] & 0.04 & 0.89 \\
\M{gemma-4-31b-it}             & $-1.56$ & $-1.73$ & $-1.38$ & $-4.18$ & [$-4.49$, $-3.85$] & 0.03 & 0.93 \\
\M{nemotron-3-ultra-550b-a55b} & $-2.12$ & $-2.36$ & $-1.90$ & $-5.25$ & [$-5.65$, $-4.85$] & 0.02 & 0.94 \\
\bottomrule
\end{tabular}
\end{table}

\begin{table}[ht]
\centering
\caption{\textbf{Paired win rate against \M{gpt-5.6-sol} by rubric dimension (\%),
same task and same judge, ties excluded.} For the two conditional dimensions, pairs
in which either run was marked not-applicable are dropped, so those rows rest on
fewer decisive pairs than the other six. Computed from \M{scores\_wide.csv}.}
\label{tab:dimwin}
\small
\begin{tabular}{@{}lrrrrrrrr@{}}
\toprule
\textbf{Dimension} & \M{opus} & \M{luna} & \M{kimi} & \M{grok} & \M{gemini} & \M{muse} & \M{gemma} & \M{nemotron} \\
\midrule
Tool use              & \textbf{73.2} & 26.2 & 38.4 & 28.1 & 12.9 & 8.2 & 3.3 & 4.5 \\
Reasoning quality     & 49.1 & 17.1 & 13.7 &  6.7 &  4.7 & 3.4 & 2.3 & 2.0 \\
Task fulfillment      & 46.1 & 18.8 & 18.6 & 15.0 & 11.4 & 6.4 & 2.8 & 2.0 \\
Data handling         & 44.6 & 12.1 & 11.9 &  9.0 &  5.8 & 2.1 & 0.8 & 0.6 \\
Artifact quality      & 40.3 & 13.2 & 17.6 & 12.3 &  8.2 & 2.5 & 1.6 & 1.3 \\
Communication         & 30.0 & 17.1 & 20.6 & 13.2 &  5.3 & 2.9 & 2.1 & 2.5 \\
Scientific accuracy   & 22.5 & 18.6 & 11.1 &  5.9 &  3.0 & 2.1 & 1.8 & 1.6 \\
Honesty / calibration & 17.2 & 22.2 &  7.5 &  6.8 &  0.6 & 1.9 & 4.6 & 2.3 \\
\bottomrule
\end{tabular}
\end{table}

\begin{table}[ht]
\centering
\caption{\textbf{Mean overall by domain and model.} Computed from
\M{scores\_wide.csv} joined to the domain field of \M{run\_metrics.csv}.}
\label{tab:domain}
\small
\begin{tabular}{@{}lrrrrrrrrrr@{}}
\toprule
\textbf{Domain} & \textbf{n tasks} & \M{sol} & \M{opus} & \M{luna} & \M{kimi} & \M{grok} & \M{gemini} & \M{muse} & \M{gemma} & \M{nemotron} \\
\midrule
Chemistry, drug, materials     & 17 & 8.33 & 7.75 & 7.69 & 7.02 & 6.94 & 5.57 & 4.00 & 3.18 & 2.75 \\
Clinical and health            & 59 & 7.80 & 7.61 & 7.57 & 7.20 & 7.14 & 6.08 & 4.40 & 3.95 & 3.33 \\
Life sciences                  & 59 & 8.21 & 7.81 & 7.42 & 7.14 & 6.79 & 5.52 & 4.14 & 3.56 & 2.72 \\
Physical sciences, eng., CS    & 43 & 8.00 & 7.30 & 7.26 & 7.22 & 6.94 & 6.05 & 4.39 & 4.40 & 2.13 \\
\bottomrule
\end{tabular}
\end{table}

\begin{table}[ht]
\centering
\caption{\textbf{Effect of attachments on mean overall, by model.} 125 of 178
sessions carry at least one file. Computed by joining \M{run\_metrics.csv} to
run-level means.}
\label{tab:attach}
\small
\begin{tabular}{@{}lrrr@{}}
\toprule
\textbf{Model} & \textbf{No attachments} & \textbf{Has attachments} & \textbf{$\Delta$} \\
\midrule
\M{gemma-4-31b-it}             & 4.93 & 3.40 & $-1.52$ \\
\M{muse-spark-1.2}             & 5.29 & 3.84 & $-1.45$ \\
\M{nemotron-3-ultra-550b-a55b} & 3.69 & 2.40 & $-1.29$ \\
\M{gemini-3.6-flash}           & 6.23 & 5.68 & $-0.56$ \\
\M{claude-opus-5}              & 7.73 & 7.57 & $-0.16$ \\
\M{gpt-5.6-luna}               & 7.52 & 7.43 & $-0.09$ \\
\M{grok-4.5}                   & 7.00 & 6.94 & $-0.06$ \\
\M{gpt-5.6-sol}                & 7.98 & 8.06 & $+0.08$ \\
\M{kimi-k3}                    & 6.84 & 7.30 & $+0.46$ \\
\bottomrule
\end{tabular}
\end{table}

\begin{table}[ht]
\centering
\caption{\textbf{Mean overall by prompt-length quartile.} Quartiles are formed over
the 178 tasks by the byte length of the first user message, which is held in the task
registry rather than in the score tables (Section~\ref{sec:availability}). Eight of
the nine models are tabulated here for width; \M{gemma-4-31b-it} is plotted alongside
them in Figure~\ref{fig:conditioning} and also scores lower on Q4 than on Q1.}
\label{tab:lenquartile}
\small
\begin{tabular}{@{}llrrrrrrrrr@{}}
\toprule
\textbf{Bin} & \textbf{Median bytes} & \textbf{n} & \M{sol} & \M{opus} & \M{luna} & \M{kimi} & \M{grok} & \M{gemini} & \M{muse} & \M{nemotron} \\
\midrule
Q1 &    96 & 45 & 8.31 & 7.96 & 7.76 & 7.18 & 7.49 & 6.38 & 5.47 & 4.87 \\
Q2 &   284 & 44 & 8.23 & 7.55 & 7.52 & 7.24 & 7.04 & 5.91 & 4.25 & 2.70 \\
Q3 & 1{,}232 & 45 & 8.09 & 7.70 & 7.55 & 7.23 & 6.79 & 5.82 & 3.66 & 2.02 \\
Q4 & 6{,}217 & 44 & 7.50 & 7.23 & 6.99 & 7.02 & 6.49 & 5.24 & 3.69 & 1.51 \\
\midrule
\multicolumn{2}{@{}l}{Q1 $\rightarrow$ Q4 drop} & & 0.8 & 0.7 & 0.8 & 0.2 & 1.0 & 1.1 & 1.8 & 3.4 \\
\bottomrule
\end{tabular}
\end{table}

\begin{table}[ht]
\centering
\caption{\textbf{Run endings and outcomes.} Left: mean overall score and
majority-success rate by final stop reason. Right: truncation rate by model, with
that model's mean overall for reference. Computed from \M{run\_metrics.csv}
joined to run-level means.}
\label{tab:stop}
\small
\begin{tabular}{@{}lrrr@{\hspace{2.6em}}lrr@{}}
\toprule
\textbf{Ending} & \textbf{n} & \textbf{Overall} & \textbf{Majority} &
\textbf{Model} & \textbf{Truncated} & \textbf{Overall} \\
\midrule
\M{stop}    & 1{,}354 & 6.70 & 47.4\% & \M{nemotron-3-ultra-550b-a55b} & 64.6\% & 2.78 \\
\M{length}  &    224 & 2.22 &  0.0\% & \M{muse-spark-1.2}             & 52.8\% & 4.27 \\
\M{toolUse} &     10 & 3.60 &  0.0\% & \M{grok-4.5}                   & 11.2\% & 6.96 \\
\M{error}   &     14 & 0.00 &  0.0\% & \M{gemma-4-31b-it}             &  7.9\% & 3.86 \\
\midrule
Not truncated & 1{,}344 & 6.73 & 47.8\% & \M{claude-opus-5}           &  3.9\% & 7.61 \\
Truncated     &    258 & 2.18 &  0.0\% & \M{kimi-k3}                  &  2.8\% & 7.17 \\
              &        &      &        & \M{gpt-5.6-luna}             &  1.7\% & 7.46 \\
              &        &      &        & \M{gpt-5.6-sol}              &  0.0\% & 8.04 \\
              &        &      &        & \M{gemini-3.6-flash}         &  0.0\% & 5.84 \\
\bottomrule
\end{tabular}
\end{table}

\begin{table}[ht]
\centering
\caption{\textbf{Run-level Spearman correlates of overall score} ($n = 1{,}602$).
Computed from \M{run\_metrics.csv} joined to run-level means.}
\label{tab:correlates}
\small
\begin{tabular}{@{}lr@{\hspace{3em}}lr@{}}
\toprule
\textbf{Variable} & \textbf{$\rho$} & \textbf{Variable} & \textbf{$\rho$} \\
\midrule
Input tokens        & $-0.42$ & Turns             & $+0.23$ \\
Thinking characters & $+0.35$ & Output files      & $+0.21$ \\
Wall-clock seconds  & $+0.33$ & Tool error rate   & $-0.16$ \\
Cost (USD)          & $+0.31$ & Attachment count  & $-0.16$ \\
Tool calls          & $+0.27$ & & \\
\bottomrule
\end{tabular}
\end{table}

\begin{table}[ht]
\centering
\caption{\textbf{Cost and efficiency.} Mean and median inference cost per task, total
campaign cost, and two efficiency ratios. Computed from \M{run\_metrics.csv};
total generation cost across all 1{,}602 runs was \$3{,}649.18.}
\label{tab:cost}
\small
\begin{tabular}{@{}lrrrrrrr@{}}
\toprule
\textbf{Model} & \textbf{Overall} & \textbf{Mean \$} & \textbf{Median \$} & \textbf{Total \$} & \textbf{Majority} & \textbf{Score/\$} & \textbf{Maj.\ pts/\$} \\
\midrule
\M{gpt-5.6-sol}                & 8.04 & 8.51 & 4.72 & 1{,}514.92 & 71\% &   0.9 &   8.4 \\
\M{claude-opus-5}              & 7.61 & 6.69 & 4.04 & 1{,}190.15 & 75\% &   1.1 &  11.3 \\
\M{gpt-5.6-luna}               & 7.46 & 0.15 & 0.06 &     26.80 & 60\% &  49.5 & 395.6 \\
\M{kimi-k3}                    & 7.17 & 1.11 & 0.40 &    197.51 & 52\% &   6.5 &  46.6 \\
\M{grok-4.5}                   & 6.96 & 0.38 & 0.23 &     67.96 & 46\% &  18.2 & 119.2 \\
\M{gemini-3.6-flash}           & 5.84 & 0.82 & 0.50 &    146.45 & 25\% &   7.1 &  30.0 \\
\M{muse-spark-1.2}             & 4.27 & 2.42 & 0.27 &    430.57 & 17\% &   1.8 &   7.2 \\
\M{gemma-4-31b-it}             & 3.86 & 0.02 & 0.00 &      3.68 &  6\% & 186.4 & 298.7 \\
\M{nemotron-3-ultra-550b-a55b} & 2.78 & 0.40 & 0.08 &     71.14 &  9\% &   7.0 &  22.5 \\
\bottomrule
\end{tabular}
\end{table}

\begin{table}[ht]
\centering
\caption{\textbf{Failure co-occurrence, $P(\text{column} \mid \text{row})$ in
percent.} Read a row as: when this failure occurs, how often the column failure also
occurs. Computed from the \M{failure\_modes} field of \M{scores\_wide.csv}.}
\label{tab:cooccur}
\small
\resizebox{\textwidth}{!}{%
\begin{tabular}{@{}lrrrrrr@{}}
\toprule
\textbf{Given} & \M{overclaim.} & \M{missing\_art.} & \M{shallow} & \M{truncated} & \M{premature} & \M{stat.\ malp.} \\
\midrule
\M{overclaiming}            & 100 & 20 &  37 & 10 & 15 & 19 \\
\M{missing\_artifacts}       &  28 & 100 & 23 & 48 & 38 &  8 \\
\M{shallow\_analysis}        &  65 & 29 & 100 & 15 & 33 & 10 \\
\M{truncated\_run}           &  19 & 65 & 16 & 100 & 20 &  2 \\
\M{premature\_completion}    &  38 & 68 & 46 & 26 & 100 &  7 \\
\M{statistical\_malpractice} &  92 & 28 & 27 &  5 & 14 & 100 \\
\bottomrule
\end{tabular}
}
\end{table}

\begin{table}[ht]
\centering
\caption{\textbf{Pairwise judge agreement on the holistic overall score}
($n = 1{,}602$ runs). $A - B$ is the mean difference in level. Computed from
\M{scores\_wide.csv}.}
\label{tab:judgepairs}
\small
\begin{tabular}{@{}llrrrr@{}}
\toprule
\textbf{Judge A} & \textbf{Judge B} & \textbf{Spearman $\rho$} & \textbf{Mean $|A-B|$} & \textbf{Within $\pm1$} & \textbf{$A-B$} \\
\midrule
\M{gpt-5.6-sol} & \M{qwen3.8-max} & 0.78 & 1.37 & 62.5\% & $-1.01$ \\
\M{gpt-5.6-sol} & \M{grok-4.5}    & 0.83 & 1.18 & 69.0\% & $-0.87$ \\
\M{qwen3.8-max} & \M{grok-4.5}    & 0.89 & 0.60 & 91.3\% & $+0.14$ \\
\bottomrule
\end{tabular}
\end{table}

\begin{table}[ht]
\centering
\caption{\textbf{Judge agreement by rubric dimension.} Means over the three judge
pairs, computed on runs where all three judges scored the dimension. Computed from
\M{scores\_wide.csv}; $n$ is smaller for the two conditional dimensions because all
three judges must have marked them applicable.}
\label{tab:dimagree}
\small
\begin{tabular}{@{}lrrrr@{}}
\toprule
\textbf{Dimension} & \textbf{n runs} & \textbf{Mean pairwise $\rho$} & \textbf{Mean $|$diff$|$} & \textbf{Within $\pm1$} \\
\midrule
Artifact quality      &   971 & 0.86 & 0.92 & 77\% \\
Task fulfillment      & 1{,}602 & 0.85 & 0.87 & 81\% \\
Overall               & 1{,}602 & 0.83 & 1.05 & 74\% \\
Reasoning quality     & 1{,}602 & 0.80 & 0.95 & 78\% \\
Data handling         & 1{,}004 & 0.76 & 0.88 & 81\% \\
Scientific accuracy   & 1{,}602 & 0.76 & 1.47 & 60\% \\
Communication         & 1{,}602 & 0.76 & 0.59 & 91\% \\
Tool use              & 1{,}602 & 0.74 & 0.94 & 81\% \\
Honesty / calibration & 1{,}602 & 0.69 & 1.32 & 65\% \\
\bottomrule
\end{tabular}
\end{table}

\begin{table}[ht]
\centering
\caption{\textbf{Calibration-adjusted self-preference for the two judges that are
also contestants.} All values are mean overall scores. Computed from
\M{scores\_wide.csv}.}
\label{tab:selfpref}
\small
\begin{tabular}{@{}lrrrrr@{}}
\toprule
\textbf{Judge} & \textbf{Scores itself} & \textbf{Scores others} & \textbf{Peers score it} & \textbf{Peers score others} & \textbf{Adjusted SP} \\
\midrule
\M{gpt-5.6-sol} & 7.90 & 5.06 & 8.10 & 6.09 & $\mathbf{+0.83}$ \\
\M{grok-4.5}    & 7.26 & 6.11 & 6.80 & 5.76 & $\mathbf{+0.11}$ \\
\bottomrule
\end{tabular}
\end{table}

\begin{table}[ht]
\centering
\caption{\textbf{\M{gpt-5.6-sol}'s deviation from peer consensus, by model judged.}
``Peer mean'' is the mean of the two other judges' means for that model. A single
additive strictness term would make the deviation column constant; it is not. Outside
\M{gpt-5.6-sol}'s own family, the smallest magnitudes belong to the models the panel
already places lowest, where downward deviation is bounded by the floor of the scale, which biases the pooled baseline --- and
hence the self-preference estimate of Table~\ref{tab:selfpref} --- toward zero.
Computed from the per-judge columns of Table~\ref{tab:headline}.}
\label{tab:judgedev}
\small
\begin{tabular}{@{}lrrr@{}}
\toprule
\textbf{Model judged} & \textbf{\M{gpt-5.6-sol}} & \textbf{Peer mean} & \textbf{Deviation} \\
\midrule
\M{claude-opus-5}              & 6.40 & 8.22 & $-1.82$ \\
\M{gemini-3.6-flash}           & 4.80 & 6.36 & $-1.56$ \\
\M{kimi-k3}                    & 6.15 & 7.68 & $-1.53$ \\
\M{grok-4.5}                   & 6.15 & 7.36 & $-1.21$ \\
\M{muse-spark-1.2}             & 3.76 & 4.53 & $-0.77$ \\
\M{nemotron-3-ultra-550b-a55b} & 2.42 & 2.97 & $-0.55$ \\
\M{gpt-5.6-luna} (same family) & 7.17 & 7.60 & $\mathbf{-0.43}$ \\
\M{gemma-4-31b-it}             & 3.61 & 3.98 & $-0.37$ \\
\midrule
\M{gpt-5.6-sol} (itself)       & 7.90 & 8.11 & $\mathbf{-0.21}$ \\
\bottomrule
\end{tabular}
\end{table}

\subsection{Supporting tables}

\begin{table}[h]
\centering
\caption{\textbf{Exact systems evaluated.} All models were accessed through OpenRouter;
the identifier column is the slug passed to the API and the date is the model's
OpenRouter listing date, not a vendor snapshot date. Context length is the window
advertised by the endpoint. All nine benchmarked models ran under the stock \M{pi}
0.84.0 harness in Modal sandboxes with the same tool set, with thinking level \M{max}
requested where the model exposed one, no model-specific prompting, no sub-agents and
no retries, against zero-retention endpoints (Section~\ref{sec:provenance}). The
campaign ran between 6 and 12 August 2026. Context length does not predict truncation
(Section~\ref{sec:stops}).}
\label{tab:modelids}
\small
\resizebox{\textwidth}{!}{%
\begin{tabular}{@{}llllr@{}}
\toprule
\textbf{Model as named here} & \textbf{Provider} & \textbf{OpenRouter identifier} & \textbf{Listed} & \textbf{Context} \\
\midrule
\M{gpt-5.6-sol}                & OpenAI      & \M{openai/gpt-5.6-sol}                & 9 Jul 2026  & 1{,}050{,}000 \\
\M{claude-opus-5}              & Anthropic   & \M{anthropic/claude-opus-5}           & 24 Jul 2026 & 1{,}000{,}000 \\
\M{gpt-5.6-luna}               & OpenAI      & \M{openai/gpt-5.6-luna}               & 9 Jul 2026  & 1{,}050{,}000 \\
\M{kimi-k3}                    & Moonshot AI & \M{moonshotai/kimi-k3}                & 16 Jul 2026 & 1{,}048{,}576 \\
\M{grok-4.5}                   & xAI         & \M{x-ai/grok-4.5}                     & 8 Jul 2026  &   500{,}000 \\
\M{gemini-3.6-flash}           & Google      & \M{google/gemini-3.6-flash}           & 21 Jul 2026 & 1{,}048{,}576 \\
\M{muse-spark-1.2}             & Meta        & \M{meta/muse-spark-1.2}               & 5 Aug 2026  & 1{,}048{,}576 \\
\M{gemma-4-31b-it}             & Google      & \M{google/gemma-4-31b-it}             & 2 Apr 2026  &   262{,}144 \\
\M{nemotron-3-ultra-550b-a55b} & NVIDIA      & \M{nvidia/nemotron-3-ultra-550b-a55b} & 4 Jun 2026  &   512{,}288 \\
\midrule
\M{qwen3.8-max} (judge only)   & Alibaba     & \M{qwen/qwen3.8-max}                  & 3 Aug 2026  & 1{,}000{,}000 \\
\bottomrule
\end{tabular}
}
\end{table}

\begin{table}[h]
\centering
\caption{\textbf{Objective run metrics by model.} Medians over 178 runs except where
noted. ``No output files'' is the share of runs ending with an empty output tree.
Computed from \M{run\_metrics.csv}.}
\label{tab:runmetrics}
\small
\resizebox{\textwidth}{!}{%
\begin{tabular}{@{}lrrrrrrr@{}}
\toprule
\textbf{Model} & \textbf{Med.\ turns} & \textbf{Med.\ tool calls} & \textbf{Tool error rate} & \textbf{Med.\ minutes} & \textbf{Med.\ \$/task} & \textbf{Total \$} & \textbf{No output files} \\
\midrule
\M{gpt-5.6-sol}                & 40.5 & 59.5 & 0.04 & 17.9 & 4.72 & 1{,}514.92 & 37.1\% \\
\M{claude-opus-5}              & 38.0 & 47.0 & 0.03 & 29.0 & 4.04 & 1{,}190.15 & 17.4\% \\
\M{gpt-5.6-luna}               & 30.0 & 44.0 & 0.05 & 28.1 & 0.06 &     26.80 & 47.8\% \\
\M{kimi-k3}                    & 13.0 & 13.5 & 0.05 & 20.3 & 0.40 &    197.51 & 34.3\% \\
\M{grok-4.5}                   &  9.0 & 12.0 & 0.09 &  4.6 & 0.23 &     67.96 & 37.6\% \\
\M{gemini-3.6-flash}           & 16.0 & 15.0 & 0.10 &  4.2 & 0.50 &    146.45 & 41.6\% \\
\M{muse-spark-1.2}             & 11.0 & 11.0 & 0.26 &  2.1 & 0.27 &    430.57 & 67.4\% \\
\M{gemma-4-31b-it}             &  2.0 &  1.0 & 0.06 &  2.1 & 0.00 &      3.68 & 73.0\% \\
\M{nemotron-3-ultra-550b-a55b} &  7.0 &  7.0 & 0.20 &  3.3 & 0.08 &     71.14 & 74.7\% \\
\bottomrule
\end{tabular}
}
\end{table}

\begin{table}[h]
\centering
\caption{\textbf{Score consistency by model}, over all 4{,}806 assessment-level
holistic scores. No model is a high-variance gambler: the standard deviations are
similar across the top five, so the ranking reflects level rather than luck.
Computed from \M{scores\_wide.csv}.}
\label{tab:consistency}
\small
\begin{tabular}{@{}lrrrrrrr@{}}
\toprule
\textbf{Model} & \textbf{Mean} & \textbf{SD} & \textbf{p10} & \textbf{Median} & \textbf{p90} & \textbf{Share $\geq 8$} & \textbf{Share $\leq 4$} \\
\midrule
\M{gpt-5.6-sol}                & 8.04 & 1.61 & 6 & 9 & 9 & 82\% &  6\% \\
\M{claude-opus-5}              & 7.61 & 1.80 & 5 & 8 & 9 & 67\% &  7\% \\
\M{gpt-5.6-luna}               & 7.46 & 1.71 & 5 & 8 & 9 & 69\% &  8\% \\
\M{kimi-k3}                    & 7.17 & 1.79 & 5 & 8 & 9 & 60\% &  9\% \\
\M{grok-4.5}                   & 6.96 & 1.76 & 4 & 8 & 9 & 52\% & 11\% \\
\M{gemini-3.6-flash}           & 5.84 & 1.84 & 3 & 6 & 8 & 24\% & 27\% \\
\M{muse-spark-1.2}             & 4.27 & 2.65 & 1 & 4 & 8 & 15\% & 53\% \\
\M{gemma-4-31b-it}             & 3.86 & 2.18 & 1 & 4 & 7 &  7\% & 64\% \\
\M{nemotron-3-ultra-550b-a55b} & 2.78 & 2.66 & 0 & 2 & 7 &  7\% & 72\% \\
\bottomrule
\end{tabular}
\end{table}

\begin{table}[h]
\centering
\caption{\textbf{Conditional-dimension applicability and language compliance.} N/A
shares are the proportion of assessments on which the judge marked the dimension
inapplicable; language compliance is the share of assessments in which the judge
recorded that the answer was in the language of the prompt. Computed from
\M{scores\_wide.csv}.}
\label{tab:na}
\small
\footnotesize
\begin{tabular}{@{}lrrr@{\hspace{1.6em}}lrr@{}}
\toprule
\textbf{Model} & \textbf{Data hand.} & \textbf{Artifact q.} & \textbf{Lang.\ OK} &
\textbf{Condition} & \textbf{Data hand.} & \textbf{Artifact q.} \\
 & \textbf{N/A} & \textbf{N/A} & & & \textbf{N/A} & \textbf{N/A} \\
\midrule
\M{claude-opus-5}              & 24.7\% & 15.0\% &  97.8\% & No attachments  & 87.4\% & 58.7\% \\
\M{gemini-3.6-flash}           & 28.5\% & 41.2\% &  99.3\% & Has attachments & 10.6\% & 25.8\% \\
\M{gemma-4-31b-it}             & 49.1\% & 48.5\% &  95.1\% & \textbf{All}    & \textbf{33.5\%} & \textbf{35.6\%} \\
\M{gpt-5.6-luna}               & 32.6\% & 43.1\% &  99.8\% & & & \\
\M{gpt-5.6-sol}                & 32.0\% & 34.5\% & 100.0\% & & & \\
\M{grok-4.5}                   & 29.0\% & 34.5\% & 100.0\% & & & \\
\M{kimi-k3}                    & 33.1\% & 30.3\% &  99.8\% & & & \\
\M{muse-spark-1.2}             & 37.1\% & 41.8\% &  88.4\% & & & \\
\M{nemotron-3-ultra-550b-a55b} & 35.0\% & 31.8\% &  89.1\% & & & \\
\bottomrule
\end{tabular}
\end{table}

\begin{table}[h]
\centering
\caption{\textbf{Unsolved tasks and unique solvers by domain.} A task is unsolved if
no model's run was called fully successful by a majority of judges; it has a unique
solver if exactly one of the nine models cleared that bar. Computed from
\M{scores\_wide.csv} and \M{run\_metrics.csv}.}
\label{tab:unsolved}
\small
\begin{tabular}{@{}lrrrr@{}}
\toprule
\textbf{Domain} & \textbf{n tasks} & \textbf{Unsolved} & \textbf{\% unsolved} & \textbf{Unique-solver tasks} \\
\midrule
Physical sciences, engineering, CS & 43 & 7 & 16.3\% & 8 \\
Life sciences                      & 59 & 8 & 13.6\% & 7 \\
Clinical and health                & 59 & 6 & 10.2\% & 6 \\
Chemistry, drug, materials         & 17 & 1 &  5.9\% & 2 \\
\midrule
\textbf{All}                       & \textbf{178} & \textbf{22} & \textbf{12.4\%} & \textbf{23} \\
\bottomrule
\end{tabular}
\end{table}

\begin{table}[h]
\centering
\caption{\textbf{Outcome by tool error rate and by attachment count.} Both are
run-level bins. The tool-error relationship is non-monotone: the best outcomes come
from runs that attempted enough to fail occasionally and recovered. Computed from
\M{run\_metrics.csv} joined to run-level means.}
\label{tab:bins}
\small
\begin{tabular}{@{}lrrr@{\hspace{2.4em}}lrrr@{}}
\toprule
\textbf{Tool error rate} & \textbf{n} & \textbf{Overall} & \textbf{Majority} &
\textbf{Attachments} & \textbf{n} & \textbf{Overall} & \textbf{Majority} \\
\midrule
0        & 664 & 5.95 & 42.0\% & 0    & 477 & 6.36 & 50.7\% \\
0--5\%   & 268 & 7.47 & 59.3\% & 1    & 477 & 6.28 & 45.7\% \\
5--15\%  & 357 & 6.42 & 40.6\% & 2--3 & 333 & 5.66 & 32.4\% \\
$>$15\%  & 313 & 4.36 & 18.8\% & 4+   & 315 & 5.39 & 23.5\% \\
\bottomrule
\end{tabular}
\end{table}

\begin{table}[h]
\centering
\caption{\textbf{Mean overall by attached file type}, for file types appearing on at
least ten tasks. ``Spread'' is best model minus worst model. File extensions are
task-registry attributes rather than fields of the score tables
(Section~\ref{sec:availability}). A task attaching several formats contributes to
each, so the rows are not disjoint.}
\label{tab:filetype}
\small
\begin{tabular}{@{}lrrrrrrrrrrr@{}}
\toprule
\textbf{Type} & \textbf{n} & \M{sol} & \M{opus} & \M{luna} & \M{kimi} & \M{grok} & \M{gemini} & \M{muse} & \M{gemma} & \M{nemotron} & \textbf{Spread} \\
\midrule
\M{.docx} & 40 & 8.18 & 7.77 & 7.55 & 7.45 & 7.08 & 5.83 & 3.46 & 2.85 & 2.44 & 5.73 \\
\M{.pdf}  & 29 & 7.93 & 7.70 & 7.44 & 7.40 & 6.84 & 5.52 & 2.52 & 2.41 & 1.78 & 6.15 \\
\M{.xlsx} & 18 & 8.15 & 7.59 & 7.41 & 7.19 & 6.91 & 5.98 & 4.04 & 3.33 & 2.19 & 5.96 \\
\M{.md}   & 14 & 7.52 & 6.98 & 7.40 & 7.43 & 6.62 & 5.00 & 3.71 & 3.88 & 0.93 & 6.60 \\
\M{.csv}  & 13 & 7.77 & 6.97 & 6.74 & 7.08 & 6.05 & 5.08 & 3.41 & 2.95 & 1.97 & 5.79 \\
\M{.zip}  & 13 & 8.26 & 7.46 & 6.67 & 7.62 & 6.51 & 5.15 & 2.87 & 2.59 & 1.46 & 6.79 \\
\M{.txt}  & 10 & 8.57 & 7.63 & 7.00 & 7.73 & 6.67 & 5.80 & 3.47 & 2.93 & 2.60 & 5.97 \\
\bottomrule
\end{tabular}
\end{table}

\begin{table}[h]
\centering
\caption{\textbf{Mean overall by sampling batch.} The two batches were drawn one day
apart and sampled independently; model ordering is stable across them.
Computed from \M{run\_metrics.csv} joined to run-level means.}
\label{tab:batch}
\small
\begin{tabular}{@{}lrr@{}}
\toprule
\textbf{Model} & \textbf{2026-08-06-full (93 tasks)} & \textbf{2026-08-07-batch2-cpu (85 tasks)} \\
\midrule
\M{gpt-5.6-sol}                & 8.00 & 8.08 \\
\M{claude-opus-5}              & 7.51 & 7.73 \\
\M{gpt-5.6-luna}               & 7.50 & 7.41 \\
\M{kimi-k3}                    & 7.02 & 7.33 \\
\M{grok-4.5}                   & 6.98 & 6.93 \\
\M{gemini-3.6-flash}           & 5.91 & 5.76 \\
\M{muse-spark-1.2}             & 4.79 & 3.71 \\
\M{gemma-4-31b-it}             & 4.33 & 3.34 \\
\M{nemotron-3-ultra-550b-a55b} & 3.18 & 2.35 \\
\bottomrule
\end{tabular}
\end{table}

\begin{table}[h]
\centering
\caption{\textbf{Judging operations.} All 4{,}806 assessments completed; 4{,}798
produced schema-valid scores on the first attempt, 6 required two attempts and 2
required three. Computed from \M{scores\_wide.csv}.}
\label{tab:judgeops}
\small
\begin{tabular}{@{}lrrrr@{}}
\toprule
\textbf{Judge} & \textbf{Assessments} & \textbf{Mean artifacts opened} & \textbf{Mean confidence} & \textbf{Mean overall awarded} \\
\midrule
\M{gpt-5.6-sol} & 1{,}602 & 3.65 & 0.92 & 5.37 \\
\M{grok-4.5}    & 1{,}602 & 4.22 & 0.86 & 6.24 \\
\M{qwen3.8-max} & 1{,}602 & 4.12 & 0.83 & 6.38 \\
\midrule
\textbf{All}    & \textbf{4{,}806} & \textbf{4.00} & \textbf{0.87} & \textbf{6.00} \\
\bottomrule
\end{tabular}
\\[0.4em]
\footnotesize Total judging cost \$996.99 over 151.1 hours of judge wall-clock time;
72 assessments (1.5\%) were made with confidence below 0.7.
\end{table}

\end{document}